\PassOptionsToPackage{table}{xcolor}
\documentclass{article}
\usepackage{iclr2027_conference,times}
\usepackage{amsmath,amsfonts,amssymb,amsthm}
\usepackage{graphicx}
\graphicspath{{./}{figures/}}
\usepackage[breaklinks=true]{hyperref}
\usepackage{url}
\usepackage{booktabs}
\usepackage{multirow}
\usepackage{array}
\usepackage{tabularx}
\usepackage{enumitem}
\usepackage{nicefrac}
\usepackage{microtype}
\usepackage{wrapfig}
\usepackage{placeins}
\usepackage{needspace}
\usepackage[most]{tcolorbox}
\hypersetup{
colorlinks=true,
citecolor=blue!55!black,
linkcolor=blue!55!black,
urlcolor=blue!55!black,
}

\definecolor{darkblue}{rgb}{0.0,0.0,0.45}
\definecolor{darkred}{rgb}{0.55,0.0,0.0}

\newcommand{\ms}[2]{\ensuremath{#1\pm#2}}
\newcommand{\bms}[2]{\ensuremath{\mathbf{#1}\pm#2}}
\definecolor{oursblue}{RGB}{226,238,249}
\definecolor{improvegray}{RGB}{242,242,242}
\definecolor{improvegreen}{RGB}{0,120,72}
\newcommand{\ourscell}[1]{\cellcolor{oursblue}{#1}}
\newcommand{\improvecell}[1]{\cellcolor{improvegray}{#1}}
\newcommand{\improveval}[1]{\improvecell{\textcolor{improvegreen}{\ensuremath{\mathbf{#1\%\uparrow}}}}}

\newtcolorbox{keybox}{
colback=gray!8,
colframe=black,
boxrule=0.5pt,
arc=2pt,
left=6pt,
right=6pt,
top=6pt,
bottom=6pt,
}

\newtcolorbox{resultbox}{
colback=gray!7,
colframe=black,
boxrule=0.5pt,
arc=2pt,
left=6pt,
right=6pt,
top=6pt,
bottom=6pt,
}

\makeatletter
\newcommand{\StartAppendixWithContents}{\newpage
  \appendix
  \onecolumn
  \begingroup
  \hypersetup{linkcolor=darkred}
  \phantomsection
  \pdfbookmark[1]{Appendix}{appendix}
  \begin{center}
    {\LARGE\bfseries Appendix\par}
    \vspace{1em}
  \end{center}
  \setcounter{tocdepth}{2}
  \small
  \setlength{\parskip}{0pt}
  \@starttoc{apc}
  \endgroup
  \newpage
}
\newcommand{\appsection}[1]{\Needspace{8\baselineskip}\section{#1}\addcontentsline{apc}{section}{\protect\numberline{\thesection}#1}}
\newcommand{\appsubsection}[1]{\Needspace{6\baselineskip}\subsection{#1}\addcontentsline{apc}{subsection}{\protect\numberline{\thesubsection}#1}}
\makeatother

\theoremstyle{plain}
\newtheorem{theorem}{Theorem}
\newtheorem{proposition}[theorem]{Proposition}
\newtheorem{lemma}[theorem]{Lemma}
\newtheorem{corollary}[theorem]{Corollary}
\theoremstyle{remark}
\newtheorem{remark}[theorem]{Remark}

\title{RAP: Radial-Angular Random Projection for Hyperbolic Representations}
\author{Jiahong Liu\textsuperscript{1}\quad
Wenhao Yu\textsuperscript{1}\quad
Yifan Li\textsuperscript{1}\quad
Yanyu Chen\textsuperscript{1}\\
\textbf{Rex Ying\textsuperscript{2}\quad
Menglin Yang\textsuperscript{3}\quad
Irwin King\textsuperscript{1}}\\[0.45em]
{\normalfont\small\textsuperscript{1}The Chinese University of Hong Kong
\quad\textsuperscript{2}Yale University}\\
{\normalfont\small\textsuperscript{3}The Hong Kong University of Science and Technology (Guangzhou)}\\[0.25em]
{\normalfont\small Email: \texttt{jiahong.liu21@gmail.com}}}
\hypersetup{
  pdftitle={RAP: Radial-Angular Random Projection for Hyperbolic Representations},
  pdfauthor={Jiahong Liu, Wenhao Yu, Yifan Li, Yanyu Chen, Rex Ying, Menglin Yang, Irwin King}
}

\begin{document}
\maketitle
\lhead{Preprint}

\begin{abstract}
Random projection is a basic compression primitive for Euclidean
representation learning, with the Johnson--Lindenstrauss lemma giving a
target dimension $k=O(\varepsilon^{-2}\log n)$ independent of the ambient
dimension. As hyperbolic representations gain increasing attention in modern
Transformers, vision-language models, and large language models, the same
need for oblivious dimensionality reduction arises, but no analogous
primitive yet exists for hyperbolic representations. We study this gap.
Two natural attempts fail in dual ways: a tangent-space scheme gives a
clean local guarantee but is chart-bound and pays a curvature penalty,
while a fully hyperbolic projection in the hyperboloid model is geometrically native
but mixes radial and angular error before feeding them into the
conditioning of $\operatorname{arcosh}$. We propose \emph{radial-angular
random projection} (RAP), which preserves the radial coordinate exactly
and applies a Euclidean random projection only to the angular direction,
lifting the result back to the target hyperboloid. RAP is
hyperbolic-to-hyperbolic by construction, costs one Gaussian matrix-vector
product, and reduces hyperbolic compression to a normalized angular JL
question on $\mathbb{S}^{d-1}$. We prove a finite-set RAP guarantee with
explicit $\sinh(\sqrt{c}\,r_x)\sinh(\sqrt{c}\,r_y)$ conditioning,
complemented by a local tangent-space JL theorem and an
arcosh-conditioning lemma for the fully hyperbolic comparator. Across an
origin-centered broad-radius point cloud, a balanced tree, a clustered
Gaussian, and the WordNet noun hierarchy, RAP attains the lowest mean
relative distortion in every $(n,k)$ cell, reducing distortion
over the strongest hyperbolic baseline by approximately $48\%$--$49\%$ on
broad-radius data. RAP also scales to million-point datasets and improves
retrieval on vision--language representations over hyperbolic projection baselines.
Code is available in the
\href{https://github.com/HUBERILT/RAP}{project repository}.
\end{abstract}

\section{Introduction}

The Johnson--Lindenstrauss (JL) lemma states that any $n$-point set in
Euclidean space admits a linear embedding into dimension
$k = O(\varepsilon^{-2}\log n)$ with all pairwise distances preserved up to
$1\pm\varepsilon$ \citep{johnson1984extensions,dasgupta2003elementary,larsen2017optimal}.
The resulting random-projection primitive underlies image and text retrieval
\citep{bingham2001random}, approximate nearest-neighbor search
\citep{ailon2009fast}, randomized numerical linear algebra
\citep{halko2011finding}, the random-feature view of scalable attention
\citep{choromanski2021rethinking}, and more recent low-rank decomposition
and projection schemes for compressing the weights and key-value caches of
large language models \citep{saha2024compressing,chang2025palu}.

Hyperbolic representations have become a standard tool wherever the data
carries latent hierarchy \citep{yang2026survey}.
Poincar\'e and Lorentz embeddings achieve low
distortion on taxonomies and trees \citep{nickel2017poincare,sala2018representation},
hyperbolic neural networks lift the standard differentiable toolkit onto
the manifold \citep{ganea2018hyperbolic,chami2019hyperbolic}, and recent
work extends this geometry to Transformers, vision-language models, and
hyperbolic LLMs \citep{yang2024hypformer,he2025helm}. As these models grow,
their hyperbolic embeddings become high-dimensional. JL gives Euclidean
representations a generic answer to this regime, oblivious to the input
set and independent of the ambient $d$; no analogous primitive yet exists
for hyperbolic representations. The algorithmic question becomes
unavoidable: \textit{can high-dimensional hyperbolic embeddings be randomly
compressed while preserving the distances that encode hierarchy?}

\textbf{Challenges.} There are two challenges to developing a hyperbolic
analogue of JL. \textbf{First}, hyperbolic space has no global linear chart,
where a chart is a coordinate map from a manifold neighborhood to a Euclidean
vector space;
Euclidean random projections act on a single ambient vector space, but on a
curved manifold a linear map only makes sense locally, inside a tangent
space whose validity decays with radius. \textbf{Second}, the hyperbolic
distance passes through $\operatorname{arcosh}$, whose derivative behaves
like $(z-1)^{-1/2}$ near $z=1$; small perturbations of Lorentz inner
products are amplified at small distances and compressed at large ones,
which defeats uniform multiplicative control. The two natural responses
inherit these challenges in dual forms. A tangent-space scheme
$\exp\circ\,\mathbf{P}\circ\log$ yields a clean local theorem but is
chart-bound and pays an explicit $cR^2$ curvature penalty. A fully
hyperbolic scheme that sketches spatial coordinates and
renormalizes is geometrically native but mixes radial and angular error
before feeding them into the ill-conditioned $\operatorname{arcosh}$.

\textbf{Our approach.} To address these challenges, we propose
\emph{radial-angular random projection} (RAP), an oblivious
hyperbolic-to-hyperbolic random projection that decomposes each point in
polar form, preserves the radial coordinate exactly, and applies a Euclidean
random projection only to the angular direction. Concretely, around a chosen
origin $o\in\mathbb{H}^d_c$, every point factors as $x=(r_x,\mathbf{u}_x)$
with $r_x=d_c(o,x)\ge 0$ and $\mathbf{u}_x\in\mathbb{S}^{d-1}$, and the
hyperbolic distance reads off through the law of cosines as a function of
$(r_x,r_y,\langle\mathbf{u}_x,\mathbf{u}_y\rangle)$. Given a Gaussian matrix
$\mathbf{P}\in\mathbb{R}^{k\times d}$, RAP keeps $r_x$ unchanged, sketches
$\mathbf{u}_x$ as $\mathbf{P}\mathbf{u}_x$, renormalizes the result on
$\mathbb{S}^{k-1}$, and lifts the polar pair back to $\mathbb{H}^k_c$. The
design is geometrically motivated: in a hyperbolic embedding of a
hierarchy, $r_x$ carries hierarchical depth and $\mathbf{u}_x$ carries
within-level similarity, so confining compression to the angular factor
preserves hierarchy by construction. Computationally, RAP requires one
Gaussian matrix-vector product followed by a normalization, and it reduces
the hyperbolic compression problem to a classical Euclidean JL question on
unit vectors in $\mathbb{S}^{d-1}$. The hyperbolic law of cosines supplies
the starting identity; the contribution is the oblivious normalized-angular
map together with a simultaneous finite-set guarantee.

\paragraph{Contributions.} The contributions of this paper are summarized
as follows.
\begin{itemize}[leftmargin=*,itemsep=2pt,topsep=2pt]
  \item We introduce RAP, an oblivious
  hyperbolic-to-hyperbolic random projection that preserves the radial
  coordinate exactly and reduces hyperbolic compression to a normalized
  Euclidean JL question on unit angular vectors in $\mathbb{S}^{d-1}$. The
  map requires a single Gaussian matrix-vector product followed by a
  normalization, and lands on the target hyperboloid by construction.
  \item We prove a finite-set RAP guarantee under bounded
  radii and pairwise separation, including hyperboloid validity, exact
  radial preservation, and an explicit
  $\sinh(\sqrt{c}\,r_x)\sinh(\sqrt{c}\,r_y)$ conditioning factor that
  governs how angular error transfers to hyperbolic distance. As
  complementary calibration, we establish a local tangent-space JL theorem
  with the explicit envelope $1\pm\varepsilon\pm O(cR^2)$, and an
  arcosh-conditioning lemma for nearby pairs in the fully hyperbolic
  comparator.
  \item On four datasets (an origin-centered
  broad-radius point cloud, a balanced tree, a clustered Gaussian, and a
  WordNet noun subtree) and across $n\in\{128,256,512\}$ and
  $k\in\{16,32,64,128\}$, RAP attains the lowest mean relative distortion in
  every $(n,k)$ cell. RAP also reduces worst-pair distortion and scales to
  million-point datasets. On real-world vision--language representations,
  it improves macro-averaged retrieval scores by $5.5\%$--$20.1\%$ relative
  to the stronger of the two hyperbolic baselines while maintaining competitive
  classification accuracy.
\end{itemize}

\section{Related Work}

\paragraph{Hyperbolic representation learning.}
Poincar\'e embeddings first established hyperbolic geometry as a
low-distortion host for taxonomies and tree-structured data
\citep{nickel2017poincare,sala2018representation}. The differentiable
toolkit on top of this foundation, including log/exp maps, M\"obius
operations, gyrovector arithmetic, and Riemannian optimization, was
developed by hyperbolic neural networks
\citep{ganea2018hyperbolic,shimizu2021hnnpp},
extended to graph convolution \citep{chami2019hyperbolic}, contrastive graph
learning \citep{zhang2025collapse}, image
classification \citep{khrulkov2020hyperbolic}, and word and sequence
embeddings \citep{tifrea2019poincare}. Probabilistic and optimization
infrastructure has grown alongside it: wrapped and Lorentzian distributions
support uncertainty modelling and metric learning
\citep{nagano2019wrapped,law2019lorentzian}, and Fr\'echet-mean
differentiation closes the gap for objectives that need a hyperbolic
centroid \citep{lou2020differentiating}. A complementary, increasingly
dominant line operates natively in the Lorentz model rather than passing
through repeated tangent charts: fully hyperbolic neural networks and CNNs
\citep{chen2021fully,bdeir2023fully}, hyperbolic Transformers
\citep{yang2024hypformer}, vision-language and multimodal hyperbolic
representations \citep{desai2023meru,pal2024compositional,peng2025hyperet,liu2026hmcl}, hyperbolic
LLMs and adaptation \citep{he2025helm,yang2025hyplora}, and recent
hyperbolic vision Transformers and Busemann-style efficient layers
\citep{alyoussef2026hexformer,vanderklis2026fast,chen2026busemann}.
These methods design \emph{trainable} layers and end-to-end architectures; our
contribution is orthogonal: an oblivious random-projection primitive that
compresses learned hyperbolic embeddings without retraining.
Appendix~\ref{app:extended-related-work} discusses broader graph-learning,
personalization, and foundation-model applications.

\paragraph{Hyperbolic dimensionality reduction.}
Existing methods for reducing the dimension of hyperbolic representations
are deterministic and data-dependent. Principal geodesic analysis
\citep{fletcher2004principal} is the classical Riemannian generalization of
PCA: it tangent-linearizes the data at a base point and applies Euclidean
PCA in that chart. Nested hyperbolic spaces \citep{fan2022nested}
parameterize a family of low-dimensional hyperbolic submanifolds and learn a
projection onto the closest one, treating dimension reduction as a
trainable component of the model. HoroPCA \citep{chami2021horopca} formulates
hyperbolic principal directions through horospherical projections in the
Poincar\'e ball, producing a faithful hyperbolic-to-hyperbolic reduction
that respects the manifold's boundary structure. Representation-tradeoff
analyses for hyperbolic embeddings of trees and other discrete structures
\citep{sala2018representation} clarify the achievable tradeoff between
target dimension and worst-case distortion in the curved setting. None of these constructions is
data-oblivious or carries a JL-style finite-set distance preservation
guarantee, so they answer a different question than the one we study: they
fit a low-dimensional hyperbolic representation to a particular dataset,
whereas we ask for a single random map that preserves pairwise distances
on any sufficiently regular finite point set. We therefore exclude these
methods from the empirical comparison in
Section~\ref{sec:experiments};
Appendix~\ref{app:learned-dr-comparison} details the distinction in
problem statement and tabulates how HoroPCA, nested hyperbolic spaces,
and RAP relate. {Half-space
\citep{benjamini2009dimension,deng2026locality} is included in
Tables~\ref{tab:summary} and~\ref{tab:wordnet};
Appendix~\ref{app:learned-dr-comparison}, below Table~\ref{tab:learned-dr-comparison},
details the geometric comparison.}
Hyperbolic compression also includes HiHPQ, which learns hierarchical
product-quantization codes for image retrieval \citep{qiu2024hihpq}; its
discrete coding objective differs from RAP's continuous distance-preserving
projection.

\paragraph{Classical JL and modern random sketching.}
Since \citet{johnson1984extensions}, Euclidean random projections have
accumulated a mature toolbox: streamlined probabilistic proofs
\citep{dasgupta2003elementary}, database-friendly and sparse variants
\citep{achlioptas2003database,matousek2008variants}, fast linear transforms
\citep{ailon2009fast}, feature-hashing applications
\citep{weinberger2009feature}, and tight $k=\Theta(\varepsilon^{-2}\log n)$
lower bounds \citep{larsen2017optimal}, with a broader metric-embedding
context tracing back to \citet{bourgain1985lipschitz}. Random projection
underlies image and text retrieval \citep{bingham2001random} and randomized
numerical linear algebra \citep{halko2011finding}; the random-feature view
of attention provides scalable Transformer approximations
\citep{choromanski2021rethinking}, and recent work compresses the weights
and key-value caches of large language models with low-rank/low-precision
decompositions and low-rank projection schemes
\citep{saha2024compressing,chang2025palu}. Despite this
breadth, every line in this literature is built on a global linear chart
and Euclidean inner products, so its derivations do not transfer verbatim
to a curved metric space; obtaining an oblivious random-projection
primitive for hyperbolic representations therefore requires a construction
tailored to the geometry, which is the gap the present paper closes.

\section{Preliminaries}

Let $\mathbb{H}^d_c$ denote $d$-dimensional hyperbolic space of constant
curvature $-c<0$, with hyperbolic distance $d_c(\cdot,\cdot)$, exponential
map $\exp_a:T_a\mathbb{H}^d_c\to\mathbb{H}^d_c$, and logarithmic map
$\log_a$ at $a\in\mathbb{H}^d_c$. The geometry is intrinsic: we instantiate
$\mathbb{H}^d_c$ either as the Poincar\'e ball or the Lorentz hyperboloid
when an explicit coordinate computation is needed, and the choice between
the two does not affect distances or the maps $\exp_a,\log_a$
(Appendix~\ref{app:hyperbolic-background}). We will use the polar form, the
local Euclideanization through normal coordinates, and a standard
distributional JL guarantee; the three together feed both the tangent-space
comparator and RAP. A complete notation table is given in
Appendix~\ref{app:notation}.

\begin{figure}[!t]
\centering
\includegraphics[width=0.9999\textwidth]{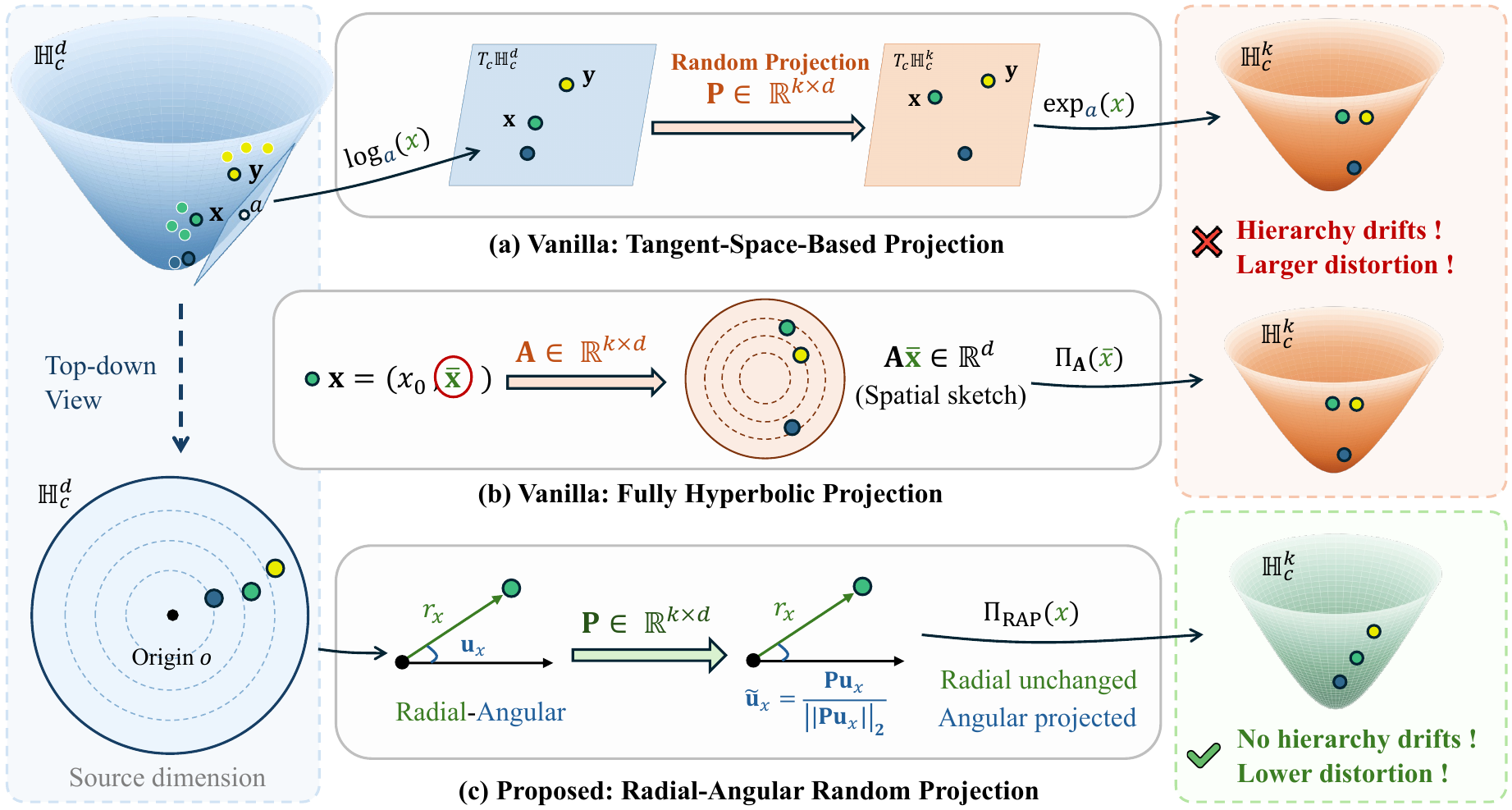}
\caption{Unified schematic of the three hyperbolic random projection
paradigms studied in this paper. Panel (a) shows the tangent-space baseline:
points are sent to a Euclidean tangent chart by $\log_a$, sketched by a
Gaussian JL matrix, and mapped back by an exponential map. Panel (b) shows
the fully hyperbolic baseline: spatial coordinates are projected in the
hyperboloid model and
the time coordinate is recomputed, which keeps the output on the hyperboloid
but couples radial and angular perturbations. Panel (c) shows RAP, our
proposed radial-angular projection: the radius $r_x$ is preserved exactly
while only the angular direction $\mathbf{u}_x$ is projected and
renormalized. This separation keeps hierarchy-carrying radial depth
unchanged and confines compression error to angular information.}
\label{fig:method-overview}
\end{figure}

\paragraph{Polar coordinates.}
Around an origin or anchor $o\in\mathbb{H}^d_c$, every point admits the polar
representation $x=(r_x,\mathbf{u}_x)$ with $r_x=d_c(o,x)\ge 0$ and
$\mathbf{u}_x\in\mathbb{S}^{d-1}$. The hyperbolic law of cosines reads
\begin{equation}
\cosh(\sqrt{c}\,d_c(x,y))
=
\cosh(\sqrt{c}\,r_x)\cosh(\sqrt{c}\,r_y)
-\sinh(\sqrt{c}\,r_x)\sinh(\sqrt{c}\,r_y)\langle \mathbf{u}_x,\mathbf{u}_y\rangle.
\label{eq:law-of-cosines}
\end{equation}
Here $r_x$ encodes hierarchical depth and
$\langle\mathbf{u}_x,\mathbf{u}_y\rangle$ encodes within-level similarity.

\paragraph{Local Euclideanization.}
In geodesic normal coordinates centered at $a$, the metric admits
\begin{equation}
g_{\mathbf{u}} = I + O(c\|\mathbf{u}\|^2),
\label{eq:metric-expansion}
\end{equation}
uniformly on small balls. Concretely, for any
$\mathbf{u},\mathbf{v}\in T_a\mathbb{H}^d_c$ with
$\|\mathbf{u}\|,\|\mathbf{v}\|\le R$ and $R\sqrt{c}$ small,
\begin{equation}
d_c(\exp_a\mathbf{u},\exp_a\mathbf{v})
=
\|\mathbf{u}-\mathbf{v}\|_2\,\bigl(1+O(cR^2)\bigr),
\end{equation}
so on a sufficiently small ball the hyperbolic distance agrees with the
Euclidean tangent distance up to a curvature correction.
Equation~\eqref{eq:metric-expansion} is what makes the tangent-space scheme analyzable, and it is also what
confines that scheme to small balls.

\paragraph{Euclidean JL matrix.}
A standard JL matrix is $\mathbf{P}\in\mathbb{R}^{k\times d}$ with i.i.d.\
$\mathcal{N}(0,1/k)$ entries. For any finite vector set
$\{\mathbf{u}_i\}_{i=1}^n$, with $k=O(\varepsilon^{-2}\log(n/\delta))$,
\begin{equation}
(1-\varepsilon)\|\mathbf{u}_i-\mathbf{u}_j\|
\le \|\mathbf{P}\mathbf{u}_i-\mathbf{P}\mathbf{u}_j\|
\le (1+\varepsilon)\|\mathbf{u}_i-\mathbf{u}_j\|
\quad \forall i,j
\label{eq:euclidean-jl}
\end{equation}
holds with probability at least $1-\delta$. Tangent-chart projections in
\S\ref{sec:tangent} are stated through orthonormal coordinate maps
$U_a:T_a\mathbb{H}^d_c\to\mathbb{R}^d$ and
$U_{\tilde a}:T_{\tilde a}\mathbb{H}^k_c\to\mathbb{R}^k$; the resulting maps
are basis-independent.

\section{Methodology}

\subsection{Problem Definition}
\label{sec:problem-definition}

Let $X=\{x_1,\ldots,x_n\}\subset\mathbb{H}^d_c$ and let $k<d$ be a target
dimension. We seek a randomized map
$\Phi:\mathbb{H}^d_c\to\mathbb{H}^k_c$ such that
\begin{equation}
\Pr_{\Phi}\!\left[
\max_{i<j}
\left|\frac{d_c(\Phi(x_i),\Phi(x_j))}{d_c(x_i,x_j)}-1\right|\le\varepsilon
\right]\ge 1-\delta,
\label{eq:hyp-jl-objective}
\end{equation}
with $k$ growing polylogarithmically in $n$. We denote the worst-pair
distortion of a realized map by
\begin{equation}
\Delta_X(\Phi)=\max_{i<j}
\left|\frac{d_c(\Phi(x_i),\Phi(x_j))}{d_c(x_i,x_j)}-1\right|.
\end{equation}
Equation~\eqref{eq:hyp-jl-objective} is intentionally stronger than what is
provable globally on a curved manifold, and the three schemes we study relax
it in different ways. The tangent-space scheme delivers the bound only inside
a small geodesic ball, paying a $cR^2$ curvature term outside the JL term
itself. The fully hyperbolic scheme is global on the manifold but loses a clean
multiplicative bound near coincident pairs because of the
$\operatorname{arcosh}$ distance map. RAP keeps the radial coordinate exact,
so Equation~\eqref{eq:hyp-jl-objective} reduces to a finite-set angular JL bound on
$\mathbb{S}^{d-1}$ multiplied by an explicit hyperbolic conditioning factor;
this is what we eventually prove in Appendix~\ref{app:rap-advantage}.
Figure~\ref{fig:method-overview} gives a unified schematic of these three
design choices before we define them formally.

\subsection{Vanilla Method I: Tangent-Space-Based Projection}
\label{sec:tangent}

Given anchors $a\in\mathbb{H}^d_c$ and $\tilde a\in\mathbb{H}^k_c$ and a
Gaussian JL matrix $\mathbf{P}\in\mathbb{R}^{k\times d}$, the tangent-space scheme
is
\begin{equation}
\Pi_{a,\mathbf{P}}(x)
=
\exp_{\tilde a}\!\left(U_{\tilde a}^{-1}\mathbf{P} U_a\log_a(x)\right).
\label{eq:tangent-scheme}
\end{equation}
The construction is local-by-design: $\log_a$ supplies a Euclidean chart, JL
acts there, and $\exp_{\tilde a}$ returns to the manifold. Its guarantee
inherits both this locality and a curvature penalty from Equation~\eqref{eq:metric-expansion}.
We state it as a calibrated reference point: it shows that, inside a small
enough ball, the standard JL distortion bound transfers to hyperbolic
distances up to an explicit $O(cR^2)$ correction.

\begin{resultbox}
\begin{theorem}[Local tangent-space JL]
\label{prop:local-jl}
Let $X=\{x_1,\dots,x_n\}\subset B(a,R)\subset\mathbb{H}^d_c$ with
$R\le R_0/\sqrt{c}$. Let $\mathbf{P}$ be a Gaussian JL matrix with
$k\ge C\varepsilon^{-2}\log(n/\delta)$, and assume the projected tangent
cloud lies in a ball where the same comparison is valid. Then with probability
at least $1-\delta$, for all $i,j$,
\begin{equation}
(1-\varepsilon-C_1 cR^2)\,d_c(x_i,x_j)
\le d_c(\Pi_{a,\mathbf{P}}x_i,\Pi_{a,\mathbf{P}}x_j)
\le (1+\varepsilon+C_1 cR^2)\,d_c(x_i,x_j).
\end{equation}
\end{theorem}
\end{resultbox}

The two error terms decouple. The $\varepsilon$ term is the standard JL
contribution and shrinks with the target dimension $k$; the $C_1 cR^2$ term
is the curvature contribution and does not vanish with $k$. The constants
$R_0$ and $C_1$ are universal: $R_0$ sets the locality range of the source
ball (we take $R\sqrt{c}\le 1/2$), and $C_1=5C_0$ absorbs the source-side,
target-side, and cross-term curvature errors composed in
Appendix~\ref{app:full-proof}, where the full proof is given. The locality
of the bound is structural rather than an artifact of the proof: a uniform
multiplicative bound on an unbounded geodesic ball would have to absorb the
$cR^2$ term as $R\to\infty$, which the metric expansion
in Equation~\eqref{eq:metric-expansion} rules out. The local form is therefore the
natural statement for any tangent-chart projection in negatively curved
space.

\subsection{Vanilla Method II: Fully Hyperbolic Projection}
\label{sec:lorentz}

In the hyperboloid model a point is $x=(x_0,\bar{\mathbf{x}})\in\mathbb{R}^{d+1}$
with time coordinate $x_0$ and spatial part $\bar{\mathbf{x}}\in\mathbb{R}^d$
satisfying $-x_0^2+\|\bar{\mathbf{x}}\|_2^2=-1/c$. A natural attempt at a
hyperbolic-to-hyperbolic map is to sketch only the $d$ spatial coordinates by
$\mathbf{A}\in\mathbb{R}^{k\times d}$ and recompute the time coordinate from
the hyperboloid constraint:
\begin{equation}
\Pi_{\mathbf{A}}(x)
=
\Bigl(\sqrt{1/c+\|\mathbf{A}\bar{\mathbf{x}}\|_2^2},\;\mathbf{A}\bar{\mathbf{x}}\Bigr).
\label{eq:lorentz-scheme}
\end{equation}
The renormalization is forced: any $\mathbf{A}\bar{\mathbf{x}}$ alone leaves
the hyperboloid, so recomputing the time coordinate returns the output to
$\mathbb{H}^k_c$. The resulting map is intrinsic and chart-free, but distance
is recovered through
\begin{equation}
d_c(x,y)=\tfrac{1}{\sqrt{c}}\operatorname{arcosh}\bigl(-c\langle x,y\rangle_L\bigr),
\qquad
\langle x,y\rangle_L=-x_0y_0+\langle\bar{\mathbf{x}},\bar{\mathbf{y}}\rangle,
\end{equation}
and $\operatorname{arcosh}'(z)\sim(z-1)^{-1/2}$ as $z\to 1^+$. Two effects
follow. First, the renormalized time coordinate depends on the perturbed
spatial norm $\|\mathbf{A}\bar{\mathbf{x}}\|_2$, so a single Euclidean sketch
perturbs both radial magnitude and angular correlation simultaneously
(Lemma~\ref{lem:lorentz-mixing}, Appendix~\ref{app:lorentz-conditioning}).
Second, the arcosh derivative amplifies inner-product perturbations on
nearby pairs and damps them on distant ones, so a uniform multiplicative
distance bound is unattainable across all distance scales.

\subsection{Proposed Method: Radial-Angular Random Projection (RAP)}
\label{sec:rap}

The hyperbolic law of cosines (Equation~\ref{eq:law-of-cosines}) writes
every pairwise distance as a function of the two radii and the angular
inner product. Preserving the radii exactly therefore reduces hyperbolic
JL to controlling the angular inner product on $\mathbb{S}^{d-1}$, where
standard Euclidean JL applies.

Fix an anchor $o\in\mathbb{H}^d_c$ and write $x=(r_x,\mathbf{u}_x)$
with $r_x=d_c(o,x)$ and $\mathbf{u}_x\in\mathbb{S}^{d-1}$. For a Gaussian
$\mathbf{P}\in\mathbb{R}^{k\times d}$,
\begin{equation}
\widetilde{\mathbf{u}}_x=\frac{\mathbf{P}\mathbf{u}_x}{\|\mathbf{P}\mathbf{u}_x\|_2},
\qquad
\Pi_{\mathrm{RAP}}(x)=\iota_k(r_x,\widetilde{\mathbf{u}}_x),
\label{eq:rap-def}
\end{equation}
where $\iota_k$ is the polar parameterization of $\mathbb{H}^k_c$ around its
canonical origin $\tilde o=(1/\sqrt{c},\mathbf{0})$:
\begin{equation}
\iota_k(r,\mathbf{v})
=
\frac{1}{\sqrt{c}}\bigl(\cosh(\sqrt{c}\,r),\,\sinh(\sqrt{c}\,r)\,\mathbf{v}\bigr),
\qquad \mathbf{v}\in\mathbb{S}^{k-1}.
\end{equation}
Computationally, RAP is one Gaussian matrix-vector product plus a
renormalization on $\mathbb{S}^{k-1}$; no exponential map is needed at the
target side.

Substituting Equation~\eqref{eq:rap-def} into
Equation~\eqref{eq:law-of-cosines}, the angular-inner-product error
transfers to the distance with conditioning weight
$\sinh(\sqrt{c}r_x)\sinh(\sqrt{c}r_y)$.

\begin{resultbox}
\begin{proposition}[RAP validity and angular reduction]
\label{prop:rap-advantage}
Assume $\|\mathbf{P}\mathbf{u}_x\|_2>0$ for the non-anchor points considered. Then
$\Pi_{\mathrm{RAP}}(x)\in\mathbb{H}^k_c$ and
$d_c(o,\Pi_{\mathrm{RAP}}(x))=r_x$. If $\mathbf{u}_x=\mathbf{u}_y$ then
$d_c(\Pi_{\mathrm{RAP}}(x),\Pi_{\mathrm{RAP}}(y))=|r_x-r_y|=d_c(x,y)$. For
any pair $x,y$, if
$|\langle\widetilde{\mathbf{u}}_x,\widetilde{\mathbf{u}}_y\rangle
-\langle\mathbf{u}_x,\mathbf{u}_y\rangle|\le\eta$,
then the corresponding perturbation of the law-of-cosines argument
in Equation~\eqref{eq:law-of-cosines} is at most
$\eta\,\sinh(\sqrt{c}r_x)\sinh(\sqrt{c}r_y)$.
\end{proposition}
\end{resultbox}

The RAP problem thus reduces to a normalized angular JL question on
$\mathbb{S}^{d-1}$: under what conditions does
$|\langle\widetilde{\mathbf{u}}_x,\widetilde{\mathbf{u}}_y\rangle
-\langle\mathbf{u}_x,\mathbf{u}_y\rangle|\le\eta$ hold uniformly over a
finite set of unit vectors? A standard JL norm-preservation event for the
angular vectors and their pairwise differences is enough
(Lemma~\ref{lem:normalized-angular-jl}). Combining this with the
$\sinh(\sqrt{c}r_x)\sinh(\sqrt{c}r_y)$ factor and the
$\operatorname{arcosh}$ inversion yields the finite-set RAP guarantee in
Appendix~\ref{app:rap-advantage}, conditioned on a bounded radius
$r_i\le R$ and a pairwise separation $d_c(x_i,x_j)\ge\tau>0$. The same
appendix records the proof of Proposition~\ref{prop:rap-advantage}.

Compared with the tangent scheme, RAP needs no small-ball restriction;
compared with the fully hyperbolic scheme, RAP eliminates the radial-mixing
term and only inherits the $\operatorname{arcosh}$ conditioning.
Appendix~\ref{app:theoretical-comparison} tabulates the three error
envelopes and records the structural source of each term.

\section{Experiments}
\label{sec:experiments}

\begin{table}[!p]
\centering
\caption{Mean relative distortion across synthetic benchmarks, reported as
mean$\pm$std. Lower is better; bold marks the best mean in each dataset--$n$
block. The Euclidean JL baseline applies Gaussian JL in origin-log
coordinates and compares Euclidean output to the original hyperbolic
distances.}
\label{tab:summary}
\small
\setlength{\tabcolsep}{2pt}
\renewcommand{\arraystretch}{0.86}
\begin{tabularx}{\textwidth}{@{}>{\raggedright\arraybackslash}p{0.58cm}>{\raggedright\arraybackslash}p{2.9cm}*{4}{>{\centering\arraybackslash}X}@{}}
\toprule[1.2pt]
$n$ & Scheme & $k=16$ & $k=32$ & $k=64$ & $k=128$ \\
\midrule
\multicolumn{6}{@{}l}{\textbf{Broad-radius benchmark}} \\
\addlinespace[1pt]
\multirow{6}{*}{$128$} & Euclidean JL & \ms{0.149}{0.008} & \ms{0.111}{0.007} & \ms{0.084}{0.005} & \ms{0.066}{0.004} \\
& Local tangent-space & \ms{0.142}{0.008} & \ms{0.100}{0.007} & \ms{0.070}{0.004} & \ms{0.049}{0.003} \\
& Fully hyperbolic & \ms{0.110}{0.006} & \ms{0.077}{0.005} & \ms{0.054}{0.002} & \ms{0.038}{0.002} \\
& {half-space} & {\ms{0.113}{0.006}} & {\ms{0.078}{0.005}} & {\ms{0.054}{0.002}} & {\ms{0.038}{0.002}} \\
& \ourscell{RAP (Ours)} & \ourscell{\bms{0.057}{0.002}} & \ourscell{\bms{0.040}{0.002}} & \ourscell{\bms{0.028}{0.001}} & \ourscell{\bms{0.020}{0.001}} \\
& \improvecell{$\Delta$ impr.} & \improveval{61.8} & \improveval{64.1} & \improveval{66.7} & \improveval{70.2} \\
\addlinespace[2pt]
\multirow{6}{*}{$256$} & Euclidean JL & \ms{0.149}{0.005} & \ms{0.111}{0.004} & \ms{0.084}{0.004} & \ms{0.067}{0.003} \\
& Local tangent-space & \ms{0.142}{0.006} & \ms{0.100}{0.004} & \ms{0.071}{0.003} & \ms{0.049}{0.002} \\
& Fully hyperbolic & \ms{0.110}{0.003} & \ms{0.078}{0.003} & \ms{0.054}{0.002} & \ms{0.038}{0.001} \\
& {half-space} & {\ms{0.112}{0.003}} & {\ms{0.078}{0.003}} & {\ms{0.054}{0.002}} & {\ms{0.038}{0.001}} \\
& \ourscell{RAP (Ours)} & \ourscell{\bms{0.057}{0.002}} & \ourscell{\bms{0.040}{0.001}} & \ourscell{\bms{0.028}{0.001}} & \ourscell{\bms{0.020}{0.001}} \\
& \improvecell{$\Delta$ impr.} & \improveval{61.8} & \improveval{64.3} & \improveval{67.0} & \improveval{70.6} \\
\addlinespace[2pt]
\multirow{6}{*}{$512$} & Euclidean JL & \ms{0.148}{0.003} & \ms{0.109}{0.003} & \ms{0.083}{0.002} & \ms{0.066}{0.002} \\
& Local tangent-space & \ms{0.140}{0.004} & \ms{0.098}{0.003} & \ms{0.069}{0.003} & \ms{0.049}{0.002} \\
& Fully hyperbolic & \ms{0.110}{0.003} & \ms{0.077}{0.002} & \ms{0.054}{0.002} & \ms{0.038}{0.001} \\
& {half-space} & {\ms{0.111}{0.003}} & {\ms{0.077}{0.002}} & {\ms{0.054}{0.002}} & {\ms{0.038}{0.001}} \\
& \ourscell{RAP (Ours)} & \ourscell{\bms{0.057}{0.001}} & \ourscell{\bms{0.040}{0.001}} & \ourscell{\bms{0.028}{0.001}} & \ourscell{\bms{0.020}{0.000}} \\
& \improvecell{$\Delta$ impr.} & \improveval{61.3} & \improveval{63.4} & \improveval{66.3} & \improveval{70.1} \\
\midrule
\multicolumn{6}{@{}l}{\textbf{Balanced tree}} \\
\addlinespace[1pt]
\multirow{6}{*}{$128$} & Euclidean JL & \ms{0.138}{0.007} & \ms{0.100}{0.007} & \ms{0.075}{0.007} & \ms{0.056}{0.006} \\
& Local tangent-space & \ms{0.137}{0.008} & \ms{0.097}{0.007} & \ms{0.069}{0.005} & \ms{0.050}{0.003} \\
& Fully hyperbolic & \ms{0.128}{0.007} & \ms{0.090}{0.006} & \ms{0.065}{0.004} & \ms{0.047}{0.003} \\
& {half-space} & {\ms{0.130}{0.009}} & {\ms{0.090}{0.006}} & {\ms{0.064}{0.005}} & {\ms{0.046}{0.003}} \\
& \ourscell{RAP (Ours)} & \ourscell{\bms{0.101}{0.004}} & \ourscell{\bms{0.071}{0.003}} & \ourscell{\bms{0.050}{0.003}} & \ourscell{\bms{0.036}{0.002}} \\
& \improvecell{$\Delta$ impr.} & \improveval{26.8} & \improveval{28.9} & \improveval{33.4} & \improveval{35.2} \\
\addlinespace[2pt]
\multirow{6}{*}{$256$} & Euclidean JL & \ms{0.139}{0.009} & \ms{0.100}{0.007} & \ms{0.075}{0.006} & \ms{0.056}{0.005} \\
& Local tangent-space & \ms{0.138}{0.009} & \ms{0.097}{0.006} & \ms{0.069}{0.005} & \ms{0.050}{0.002} \\
& Fully hyperbolic & \ms{0.129}{0.008} & \ms{0.090}{0.005} & \ms{0.064}{0.004} & \ms{0.046}{0.002} \\
& {half-space} & {\ms{0.131}{0.009}} & {\ms{0.091}{0.005}} & {\ms{0.064}{0.004}} & {\ms{0.046}{0.003}} \\
& \ourscell{RAP (Ours)} & \ourscell{\bms{0.102}{0.005}} & \ourscell{\bms{0.071}{0.004}} & \ourscell{\bms{0.050}{0.003}} & \ourscell{\bms{0.036}{0.002}} \\
& \improvecell{$\Delta$ impr.} & \improveval{26.7} & \improveval{28.9} & \improveval{33.6} & \improveval{36.2} \\
\addlinespace[2pt]
\multirow{6}{*}{$512$} & Euclidean JL & \ms{0.139}{0.008} & \ms{0.100}{0.007} & \ms{0.075}{0.007} & \ms{0.057}{0.005} \\
& Local tangent-space & \ms{0.138}{0.009} & \ms{0.097}{0.006} & \ms{0.069}{0.004} & \ms{0.050}{0.002} \\
& Fully hyperbolic & \ms{0.129}{0.007} & \ms{0.091}{0.005} & \ms{0.065}{0.004} & \ms{0.047}{0.002} \\
& {half-space} & {\ms{0.131}{0.009}} & {\ms{0.091}{0.005}} & {\ms{0.064}{0.004}} & {\ms{0.046}{0.002}} \\
& \ourscell{RAP (Ours)} & \ourscell{\bms{0.102}{0.004}} & \ourscell{\bms{0.071}{0.004}} & \ourscell{\bms{0.050}{0.002}} & \ourscell{\bms{0.036}{0.002}} \\
& \improvecell{$\Delta$ impr.} & \improveval{26.8} & \improveval{29.1} & \improveval{33.7} & \improveval{36.8} \\
\midrule
\multicolumn{6}{@{}l}{\textbf{Clustered Gaussian}} \\
\addlinespace[1pt]
\multirow{6}{*}{$128$} & Euclidean JL & \ms{0.153}{0.025} & \ms{0.113}{0.018} & \ms{0.086}{0.014} & \ms{0.074}{0.009} \\
& Local tangent-space & \ms{0.145}{0.029} & \ms{0.100}{0.016} & \ms{0.071}{0.014} & \ms{0.053}{0.011} \\
& Fully hyperbolic & \ms{0.123}{0.020} & \ms{0.085}{0.011} & \ms{0.059}{0.009} & \ms{0.043}{0.006} \\
& {half-space} & {\ms{0.126}{0.022}} & {\ms{0.086}{0.011}} & {\ms{0.059}{0.008}} & {\ms{0.043}{0.007}} \\
& \ourscell{RAP (Ours)} & \ourscell{\bms{0.093}{0.008}} & \ourscell{\bms{0.064}{0.004}} & \ourscell{\bms{0.043}{0.003}} & \ourscell{\bms{0.032}{0.003}} \\
& \improvecell{$\Delta$ impr.} & \improveval{39.2} & \improveval{42.8} & \improveval{49.8} & \improveval{57.1} \\
\addlinespace[2pt]
\multirow{6}{*}{$256$} & Euclidean JL & \ms{0.153}{0.024} & \ms{0.114}{0.019} & \ms{0.087}{0.014} & \ms{0.074}{0.009} \\
& Local tangent-space & \ms{0.145}{0.029} & \ms{0.101}{0.017} & \ms{0.071}{0.014} & \ms{0.052}{0.011} \\
& Fully hyperbolic & \ms{0.124}{0.020} & \ms{0.086}{0.011} & \ms{0.060}{0.008} & \ms{0.043}{0.007} \\
& {half-space} & {\ms{0.126}{0.021}} & {\ms{0.086}{0.011}} & {\ms{0.060}{0.008}} & {\ms{0.043}{0.007}} \\
& \ourscell{RAP (Ours)} & \ourscell{\bms{0.094}{0.007}} & \ourscell{\bms{0.065}{0.004}} & \ourscell{\bms{0.044}{0.003}} & \ourscell{\bms{0.032}{0.002}} \\
& \improvecell{$\Delta$ impr.} & \improveval{38.7} & \improveval{43.1} & \improveval{50.1} & \improveval{57.0} \\
\addlinespace[2pt]
\multirow{6}{*}{$512$} & Euclidean JL & \ms{0.153}{0.024} & \ms{0.114}{0.018} & \ms{0.087}{0.014} & \ms{0.075}{0.009} \\
& Local tangent-space & \ms{0.145}{0.028} & \ms{0.101}{0.016} & \ms{0.071}{0.013} & \ms{0.053}{0.011} \\
& Fully hyperbolic & \ms{0.124}{0.019} & \ms{0.085}{0.011} & \ms{0.060}{0.008} & \ms{0.043}{0.006} \\
& {half-space} & {\ms{0.126}{0.020}} & {\ms{0.086}{0.010}} & {\ms{0.060}{0.008}} & {\ms{0.043}{0.007}} \\
& \ourscell{RAP (Ours)} & \ourscell{\bms{0.094}{0.007}} & \ourscell{\bms{0.065}{0.004}} & \ourscell{\bms{0.044}{0.003}} & \ourscell{\bms{0.032}{0.002}} \\
& \improvecell{$\Delta$ impr.} & \improveval{38.7} & \improveval{43.1} & \improveval{50.1} & \improveval{57.4} \\
\bottomrule[1.2pt]
\end{tabularx}
\vspace{1pt}
\begin{minipage}{\textwidth}
\footnotesize
\emph{Note.} Mean and sample standard deviation over $20$ {paired} runs.
Broad-radius data use $d=128$ and average radial scales $\{0.5,1,2\}$ within
each run; tree and clustered data use $d=50$.
$\Delta$ impr. is RAP's reduction relative to Euclidean JL,
$100(1-D_{\mathrm{RAP}}/D_{\mathrm{EJL}})$; larger is better.
\end{minipage}
\end{table}

\paragraph{Setup.}
\label{sec:setup}
We compare {five} projection rules: the tangent-space vanilla method
(Section~\ref{sec:tangent}), the fully hyperbolic vanilla method
(Section~\ref{sec:lorentz}), RAP (Section~\ref{sec:rap}),
{half-space \citep{benjamini2009dimension,deng2026locality}}, and a Euclidean
JL diagnostic that sketches origin-log coordinates and measures Euclidean
output distance against the original hyperbolic distance
(Appendix~\ref{app:euclidean-baseline}); the diagnostic is intentionally
cross-geometry and is included only as a sanity check that a flattened
Euclidean sketch is not a substitute for a hyperbolic-to-hyperbolic map.
The two vanilla methods are not adversarially chosen: they instantiate the
two design patterns that dominate trainable
hyperbolic architectures (Sections~\ref{sec:tangent} and~\ref{sec:lorentz}).
We evaluate on four datasets: an origin-centered \emph{broad-radius}
hyperbolic point cloud in ambient dimension $128$, a balanced $4$-ary
depth-$5$ \emph{tree} in $\mathbb{H}^{50}$, a \emph{clustered Gaussian}
dataset of eight hyperbolic components in $\mathbb{H}^{50}$, and the
\emph{WordNet} noun subtree rooted at \texttt{animal.n.01}
\citep{miller1995wordnet}; WordNet is the
canonical hyperbolic-embedding benchmark introduced by
\citet{nickel2017poincare} and is used here as a standard real-data probe
in line with established practice in hyperbolic representation learning.
The synthetic datasets sweep $n\in\{128,256,512\}$ and
$k\in\{16,32,64,128\}$ over $20$ random seeds, with curvature fixed to
$c=1$. We report the mean relative distortion over nonzero-distance pairs;
lower values indicate output distances closer to the original $d_c(x,y)$.
Detailed dataset construction, the paired-comparison protocol, and the
choice of mean over worst-pair distortion are detailed in
Appendix~\ref{app:experimental-protocol}.

\begin{table}[!ht]
\centering
\caption{Mean relative distortion on the WordNet subtree rooted at
\texttt{animal.n.01} as subtree depth and target dimension vary, reported as
mean$\pm$std over $20$ random seeds. Lower is better; bold marks the best mean
in each combination of depth and target dimension. Euclidean JL uses the same origin-log
baseline as Table~\ref{tab:summary}.}
\label{tab:wordnet}
\small
\setlength{\tabcolsep}{2pt}
\renewcommand{\arraystretch}{0.95}
\begin{tabularx}{\textwidth}{@{}>{\raggedright\arraybackslash}p{0.8cm}>{\raggedright\arraybackslash}p{0.85cm}>{\raggedright\arraybackslash}p{2.8cm}*{4}{>{\centering\arraybackslash}X}@{}}
\toprule[1.2pt]
Depth & $n$ & Scheme & $k=16$ & $k=32$ & $k=64$ & $k=128$ \\
\midrule
\multirow{6}{*}{$3$} & \multirow{6}{*}{$273$} & Euclidean JL & \ms{0.139}{0.008} & \ms{0.099}{0.005} & \ms{0.070}{0.005} & \ms{0.050}{0.004} \\
& & Local tangent-space & \ms{0.140}{0.009} & \ms{0.099}{0.006} & \ms{0.069}{0.005} & \ms{0.050}{0.003} \\
& & Fully hyperbolic & \ms{0.135}{0.008} & \ms{0.095}{0.006} & \ms{0.067}{0.005} & \ms{0.048}{0.003} \\
& & {half-space} & {\ms{0.138}{0.008}} & {\ms{0.096}{0.006}} & {\ms{0.067}{0.005}} & {\ms{0.048}{0.003}} \\
& & \ourscell{RAP (Ours)} & \ourscell{\bms{0.099}{0.004}} & \ourscell{\bms{0.068}{0.003}} & \ourscell{\bms{0.048}{0.003}} & \ourscell{\bms{0.034}{0.002}} \\
& & \improvecell{$\Delta$ impr.} & \improveval{29.3} & \improveval{30.8} & \improveval{31.4} & \improveval{32.3} \\
\addlinespace[2pt]
\multirow{6}{*}{$4$} & \multirow{6}{*}{$512$} & Euclidean JL & \ms{0.140}{0.009} & \ms{0.101}{0.008} & \ms{0.071}{0.005} & \ms{0.052}{0.005} \\
& & Local tangent-space & \ms{0.140}{0.010} & \ms{0.100}{0.009} & \ms{0.070}{0.005} & \ms{0.050}{0.003} \\
& & Fully hyperbolic & \ms{0.133}{0.008} & \ms{0.095}{0.007} & \ms{0.066}{0.004} & \ms{0.047}{0.003} \\
& & {half-space} & {\ms{0.136}{0.009}} & {\ms{0.096}{0.008}} & {\ms{0.066}{0.005}} & {\ms{0.047}{0.003}} \\
& & \ourscell{RAP (Ours)} & \ourscell{\bms{0.100}{0.005}} & \ourscell{\bms{0.069}{0.004}} & \ourscell{\bms{0.048}{0.003}} & \ourscell{\bms{0.034}{0.002}} \\
& & \improvecell{$\Delta$ impr.} & \improveval{28.7} & \improveval{31.2} & \improveval{32.3} & \improveval{33.9} \\
\addlinespace[2pt]
\multirow{6}{*}{$5$} & \multirow{6}{*}{$512$} & Euclidean JL & \ms{0.139}{0.010} & \ms{0.102}{0.009} & \ms{0.073}{0.007} & \ms{0.053}{0.006} \\
& & Local tangent-space & \ms{0.138}{0.009} & \ms{0.099}{0.009} & \ms{0.070}{0.004} & \ms{0.049}{0.004} \\
& & Fully hyperbolic & \ms{0.129}{0.008} & \ms{0.093}{0.008} & \ms{0.065}{0.004} & \ms{0.046}{0.003} \\
& & {half-space} & {\ms{0.133}{0.008}} & {\ms{0.094}{0.008}} & {\ms{0.066}{0.004}} & {\ms{0.046}{0.004}} \\
& & \ourscell{RAP (Ours)} & \ourscell{\bms{0.101}{0.007}} & \ourscell{\bms{0.071}{0.005}} & \ourscell{\bms{0.049}{0.004}} & \ourscell{\bms{0.034}{0.003}} \\
& & \improvecell{$\Delta$ impr.} & \improveval{27.7} & \improveval{30.5} & \improveval{33.2} & \improveval{35.7} \\
\addlinespace[2pt]
\multirow{6}{*}{$6$} & \multirow{6}{*}{$512$} & Euclidean JL & \ms{0.139}{0.011} & \ms{0.103}{0.011} & \ms{0.076}{0.009} & \ms{0.056}{0.007} \\
& & Local tangent-space & \ms{0.137}{0.010} & \ms{0.099}{0.011} & \ms{0.070}{0.005} & \ms{0.050}{0.004} \\
& & Fully hyperbolic & \ms{0.128}{0.009} & \ms{0.092}{0.009} & \ms{0.065}{0.005} & \ms{0.046}{0.003} \\
& & {half-space} & {\ms{0.130}{0.008}} & {\ms{0.093}{0.009}} & {\ms{0.065}{0.005}} & {\ms{0.046}{0.003}} \\
& & \ourscell{RAP (Ours)} & \ourscell{\bms{0.103}{0.008}} & \ourscell{\bms{0.072}{0.007}} & \ourscell{\bms{0.051}{0.005}} & \ourscell{\bms{0.035}{0.003}} \\
& & \improvecell{$\Delta$ impr.} & \improveval{26.2} & \improveval{29.7} & \improveval{33.1} & \improveval{37.0} \\
\bottomrule[1.2pt]
\end{tabularx}
\vspace{1pt}
\begin{minipage}{\textwidth}
\footnotesize
\emph{Note.} We use the WordNet noun hierarchy rooted at \texttt{animal.n.01}.
Depth is the maximum hyponym depth retained before deterministic subsampling;
$n$ is the number of evaluated nodes after capping each subtree at $512$ nodes.
{All methods use the same paired runs and sorted WordNet traversal;
$k=64,128$ are expansion controls for $d=50$.}
$\Delta$ impr. denotes the relative distortion reduction of RAP over
Euclidean JL.
\end{minipage}
\end{table}

\raggedbottom
\subsection{Main Results}
\label{sec:main-results}

Tables~\ref{tab:summary} and~\ref{tab:wordnet} report mean relative
distortion across the four datasets and the full $(n,k)$ sweep. RAP attains
the lowest mean distortion in every $(n,k)$ cell.
{On broad-radius data at $n{=}512$, $k{=}32$, RAP
approximately halves the mean distortion: $0.040$, compared with $0.077$
for both fully hyperbolic and half-space at the reported precision.
The gain persists on structured datasets: on WordNet at depth $5$ and
$k{=}32$, RAP gives $0.071$, compared with $0.093$ for fully hyperbolic
and $0.094$ for half-space. Across all $29$ compression settings,
RAP reduces mean distortion relative to half-space by $20.9\%$--$49.4\%$.}

{The non-RAP rows show a consistent pattern: half-space and
fully hyperbolic projection perform similarly, and both outperform the
tangent-space and Euclidean baselines. The tangent-space penalty is visible
on broad-radius data and softens on the structured datasets. Euclidean JL
generally has the largest distortion, highlighting the limitations of
flattening and compressing origin-log coordinates. Half-space is analogous
to fully hyperbolic projection in allowing radial drift: preserving
half-space height does not fix distance to a finite origin, while projecting
Lorentz spatial coordinates changes norms and hence radii. Their similar
performance is consistent with this shared source of radial error.
RAP instead fixes the radius and normalizes the projected direction,
removing radial perturbations while controlling angular distortion.
Appendix~\ref{app:learned-dr-comparison} details the geometric distinction.}

\setlength{\intextsep}{8pt plus 2pt minus 2pt}
\subsection{Analysis and Discussion}
\label{sec:analysis-discussion}

\begin{figure}[!htbp]
\centering
\setlength{\abovecaptionskip}{5pt}
\includegraphics[width=\textwidth]{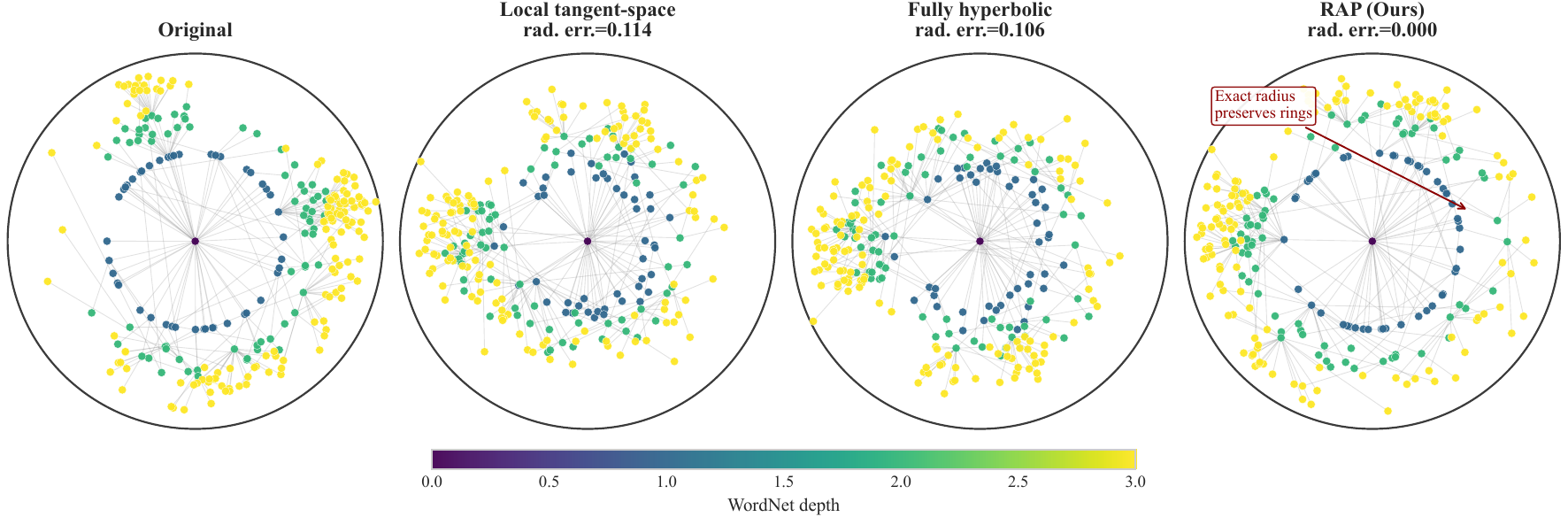}
\caption{Qualitative WordNet subtree visualization at $k=32$. RAP keeps the
depth rings closest to the original view because its radial coordinate is
preserved exactly, whereas the tangent-space and fully hyperbolic comparators
exhibit visible radial drift.}
\label{fig:wordnet-embedding-visualization}
\end{figure}

\textbf{RAP preserves the visible hierarchy structure.}
Figure~\ref{fig:wordnet-embedding-visualization} shows that RAP preserves
the depth-separated radial structure of the original WordNet embeddings,
whereas the two vanilla hyperbolic projections blur this structure through
radial drift. This reflects RAP's exact preservation of radial coordinates;
visualization details and per-node measurements are provided in
Appendix~\ref{app:radius-preservation-diagnostic}.

\begin{figure}[!htbp]
\centering
\setlength{\abovecaptionskip}{5pt}
\includegraphics[width=0.92\textwidth]{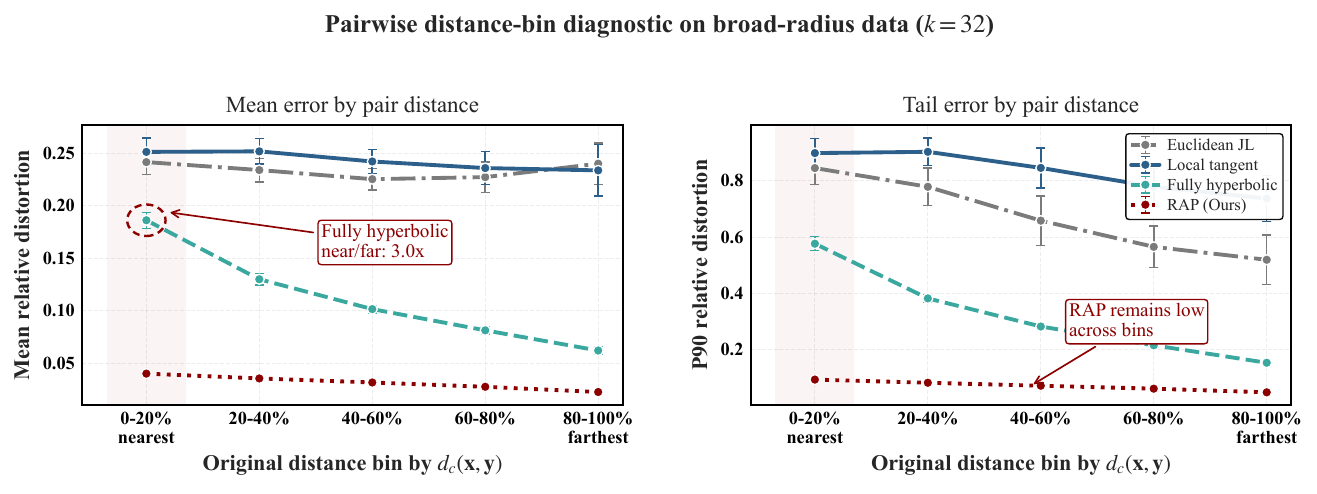}
\caption{Pairwise-distance-bin diagnostic on broad-radius data at $k=32$.
Pairs are binned by the original hyperbolic distance $d_c(x,y)$ before
distortion is measured. The fully hyperbolic comparator has pronounced
nearest-pair sensitivity, whereas RAP keeps mean and tail distortion low
and stable across all distance bins.}
\label{fig:pairwise-distance-diagnostic}
\end{figure}

\textbf{Distance-bin results support the predicted conditioning effect.}
Lemma~\ref{lem:arcosh-conditioning} predicts greater sensitivity for nearby
pairs, where $-c\langle x,y\rangle_L\approx1$.
Figure~\ref{fig:pairwise-distance-diagnostic} groups nonzero pairs by original
distance on broad-radius data at $k=32$: the fully hyperbolic comparator's
mean distortion falls from $0.186$ in the nearest $20\%$ to $0.062$ in the
farthest $20\%$. Euclidean JL and the tangent-space comparator remain
essentially flat, as neither uses $\operatorname{arcosh}$ to compute distances.
RAP has the lowest distortion in every bin; full results are in
Appendix~\ref{app:pairwise-distance-diagnostic}.

\FloatBarrier

\begin{wrapfigure}[11]{R}{0.40\textwidth}
\vspace{-1.35em}
\centering
\setlength{\abovecaptionskip}{0pt}
\includegraphics[width=\linewidth]{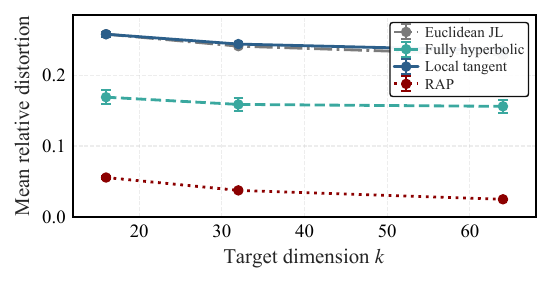}
\caption{Origin-centered synthetic benchmark; $95\%$ CIs over $20$ seeds.}
\label{fig:scheme-comparison}
\vspace{-0.6em}
\end{wrapfigure}
\textbf{The RAP gap over the strongest non-RAP comparator is structural
rather than scale-dependent.}
Figure~\ref{fig:scheme-comparison} plots
mean distortion on the origin-centered benchmark. RAP is best at every
$k$, improving over the fully hyperbolic baseline from roughly $3\times$
at $k=16$ to more than $6\times$ at $k=64$. This persistent gap is
diagnostic: increasing $k$ reduces angular JL noise, but the tangent and
fully hyperbolic baselines still perturb hierarchy-carrying radii. RAP removes
this radial error floor, leaving angular distortion as the controlled
residual error, exactly the mechanism isolated by the theory above.

\setlength{\intextsep}{12pt plus 2pt minus 2pt}
\textbf{The gain is not only a distance-distortion artifact.}
Appendix~\ref{app:property-analysis} further verifies complementary
signals beyond mean distortion: neighborhood order, rank-based
dimensionality-reduction quality, hierarchy/radius structure, and
distance-kernel spectra. Tables~\ref{tab:property-rank} and
\ref{tab:property-spectral} show that RAP gives the best mean rank
correlation, kNN/trustworthiness-style stability, and spectral-profile
errors on the tested hyperbolic representations; Table~\ref{tab:property-hierarchy}
is consistent with the exact-radius mechanism already discussed. These are
empirical diagnostics rather than new JL guarantees, but they indicate that
the same construction also preserves structural information used by
hyperbolic representation learning.

\textbf{Downstream performance on vision--language representations.}
On MERU and HyCoCLIP representations
\citep{desai2023meru,pal2024compositional}, RAP maintains competitive
classification accuracy and improves retrieval scores by
$5.5\%$--$20.1\%$ relative to the stronger of the two vanilla hyperbolic
projections. Experimental protocols and detailed results appear in
Appendix~\ref{app:vlm-evaluation} and Table~\ref{tab:vlm}.

\textbf{Worst-pair distortion and scalability.}
RAP's advantage over the vanilla hyperbolic projections extends to exact
worst-pair distortion and million-point compression.
Appendix~\ref{app:exact-and-scale} provides the all-pair and sampled-pair
protocols and detailed results.

\section{Conclusion}
\label{sec:conclusion}

RAP fills the gap in data-oblivious hyperbolic-to-hyperbolic compression,
with a simultaneous finite-set guarantee and explicit geometric conditioning.
Preserving hierarchy-carrying radii avoids tangent-space curvature bias and
radial-angular mixing using one Gaussian projection and normalization,
without data-dependent fitting. Experiments demonstrate improved mean and exact
worst-pair preservation, million-point scalability, and better vision--language
retrieval than hyperbolic projection baselines. Its simplicity and geometric
fidelity make RAP a promising building block for future hyperbolic foundation models.
In the future, we plan to extend RAP to multi-curvature and product-manifold
representations.

\FloatBarrier
\bigskip
\Needspace{8\baselineskip}
\subsection*{AI use statement}
Generative AI tools were used only for grammar, spelling, and typographical
checks, as well as reference verification. The authors reviewed all suggested edits and take full responsibility
for the final content, including all text, claims, proofs, experiments, and
artifacts.

\Needspace{8\baselineskip}
\subsection*{Ethics statement}
The method introduces no new data collection, human-subject interaction,
or generative capability; its societal impacts depend on the downstream
systems that use compressed hyperbolic representations. Biases and sensitive
information in source representations may persist after compression, and
our geometric and downstream evaluations do not establish fairness, privacy,
or safety guarantees. Since compression may affect tasks and data subgroups
differently, deployment benefits from application-specific evaluation and
safeguards, particularly when outputs inform consequential decisions.

\Needspace{8\baselineskip}
\subsection*{Reproducibility statement}
Section~\ref{sec:rap} specifies the RAP construction, and
Appendix~\ref{app:rap-advantage} provides its assumptions and proofs.
Section~\ref{sec:experiments} and Appendix~\ref{app:experimental-protocol}
document the datasets, projection settings, random-seed protocol, evaluation
metrics, and compute resources. Appendix~\ref{app:supplementary-robustness}
details the supplementary experiments, including the vision--language
evaluation in Appendix~\ref{app:vlm-evaluation} and the exact worst-pair
and scaling tests in Appendix~\ref{app:exact-and-scale}.

\Needspace{12\baselineskip}
\raggedbottom

\StartAppendixWithContents

\appsection{Hyperbolic Geometry Background}
\label{app:hyperbolic-background}

This appendix collects the model definitions, distance formulas, and
tangent-space maps used implicitly throughout the paper. The presentation is
self-contained and follows standard treatments of hyperbolic geometry
\citep{cannon1997hyperbolic,ratcliffe2019foundations} together with the
machine-learning conventions popularized by hyperbolic neural networks
\citep{ganea2018hyperbolic,nickel2018learning}. We write $\mathbb{H}^d_c$
for $d$-dimensional hyperbolic space of constant sectional curvature $-c$
with $c>0$, and $d_c$ for its geodesic distance.

\appsubsection{Models of Hyperbolic Geometry}
\label{app:hyperbolic-models}

Hyperbolic space admits several equivalent realizations as Riemannian
submanifolds of Euclidean spaces; any two such realizations are connected by
an isometry, so geometric quantities such as $d_c$ are model-independent.
Different models expose different aspects of the geometry, and the analysis
in this paper draws on two of them.

The \emph{Poincar\'e ball model} $\mathbb{D}^d_c$ is the open ball
$\{\mathbf{x}\in\mathbb{R}^d:c\|\mathbf{x}\|_2^2<1\}$ equipped with the
conformal metric
\begin{equation}
g^{\mathbb{D}}_{\mathbf{x}}=(\lambda^c_{\mathbf{x}})^2\,g^E,
\qquad
\lambda^c_{\mathbf{x}}=\frac{2}{1-c\|\mathbf{x}\|_2^2},
\end{equation}
where $g^E$ is the Euclidean inner product. The conformal factor
$\lambda^c_{\mathbf{x}}$ blows up at the boundary, so the boundary sphere
$\{c\|\mathbf{x}\|_2^2=1\}$ sits at infinite geodesic distance from any
interior point. Geodesic distance has the closed form
\begin{equation}
d_c(\mathbf{x},\mathbf{y})
=
\frac{1}{\sqrt{c}}\operatorname{arcosh}\!\left(
1+\frac{2c\|\mathbf{x}-\mathbf{y}\|_2^2}
{(1-c\|\mathbf{x}\|_2^2)(1-c\|\mathbf{y}\|_2^2)}
\right).
\label{eq:poincare-distance}
\end{equation}
The model is conformal: angles between tangent vectors agree with their
Euclidean angles, which makes it convenient for visualizing low-dimensional
embeddings. The price is that straight lines through the origin are the only
Euclidean-straight geodesics; all others are circular arcs orthogonal to the
boundary.

The \emph{Lorentz} or \emph{hyperboloid model} $\mathbb{L}^d_c$ realizes
hyperbolic space as the upper sheet of a two-sheeted hyperboloid in
$\mathbb{R}^{d+1}$. Writing a point as $x=(x_0,\bar{\mathbf{x}})$ with
$x_0\in\mathbb{R}$ and $\bar{\mathbf{x}}\in\mathbb{R}^d$, and equipping
$\mathbb{R}^{d+1}$ with the Lorentzian inner product
\begin{equation}
\langle x,y\rangle_L=-x_0 y_0+\langle\bar{\mathbf{x}},\bar{\mathbf{y}}\rangle,
\end{equation}
the model is
\begin{equation}
\mathbb{L}^d_c=\{x\in\mathbb{R}^{d+1}:\langle x,x\rangle_L=-1/c,\ x_0>0\}.
\end{equation}
The Riemannian metric is the restriction of $\langle\cdot,\cdot\rangle_L$ to
the tangent bundle, which is positive definite on each tangent space. The
geodesic distance reads
\begin{equation}
d_c(x,y)=\frac{1}{\sqrt{c}}\operatorname{arcosh}\!\bigl(-c\,\langle x,y\rangle_L\bigr).
\label{eq:lorentz-distance}
\end{equation}
Geodesics are intersections of $\mathbb{L}^d_c$ with two-dimensional linear
subspaces of $\mathbb{R}^{d+1}$, so the model is convenient for closed-form
manipulations and for the spatial-coordinate sketch
in Equation~\eqref{eq:lorentz-scheme}.

The Poincar\'e ball and Lorentz models are related by a stereographic
projection from the point $(-1/\sqrt{c},\mathbf{0})$:
\begin{equation}
\pi:\mathbb{L}^d_c\to\mathbb{D}^d_c,
\qquad
\pi(x_0,\bar{\mathbf{x}})=\frac{\sqrt{c}\,\bar{\mathbf{x}}}{1+\sqrt{c}\,x_0}.
\label{eq:stereographic}
\end{equation}
The map $\pi$ is a Riemannian isometry, and identities such as the law of
cosines (Equation~\eqref{eq:law-of-cosines}) are stable under Equation~\eqref{eq:stereographic}.
We therefore state intrinsic results in terms of the model-free
$\mathbb{H}^d_c$ and switch to the Lorentz model only when an explicit
spatial-coordinate computation is needed (Section~\ref{sec:lorentz} and
Appendix~\ref{app:lorentz-conditioning}).

\appsubsection{Tangent Spaces, Exponential and Logarithmic Maps}
\label{app:expmap-logmap}

For any base point $a\in\mathbb{H}^d_c$, the tangent space $T_a\mathbb{H}^d_c$
is a $d$-dimensional Euclidean inner-product space with the metric tensor
$g_a$. The geometry around $a$ is connected to its tangent space by the
Riemannian exponential map
\begin{equation}
\exp_a:T_a\mathbb{H}^d_c\to\mathbb{H}^d_c,
\qquad
\exp_a(\mathbf{v})=\gamma_{\mathbf{v}}(1),
\end{equation}
where $\gamma_{\mathbf{v}}$ is the unique unit-speed geodesic with
$\gamma_{\mathbf{v}}(0)=a$ and $\dot\gamma_{\mathbf{v}}(0)=\mathbf{v}/\|\mathbf{v}\|_{g_a}$
emanating from $a$. Because $\mathbb{H}^d_c$ is a Cartan--Hadamard manifold,
$\exp_a$ is a diffeomorphism onto $\mathbb{H}^d_c$ and admits a global
inverse, the logarithmic map
\begin{equation}
\log_a:\mathbb{H}^d_c\to T_a\mathbb{H}^d_c,
\qquad
\|\log_a(x)\|_{g_a}=d_c(a,x).
\end{equation}
The pair $(\exp_a,\log_a)$ furnishes the geodesic normal chart used
throughout the paper: it sends a neighborhood of $a$ to a Euclidean
neighborhood of the origin in $T_a\mathbb{H}^d_c$, with norms in the chart
matching geodesic distances to $a$ exactly along radial directions and
agreeing with Euclidean distances up to an $O(c\|\mathbf{u}\|^2)$ correction
for nearby pairs (Equation~\eqref{eq:metric-expansion}).

In the Lorentz model the maps admit closed forms. Tangent vectors at
$x\in\mathbb{L}^d_c$ are characterized by $\langle x,\mathbf{v}\rangle_L=0$,
their norm is $\|\mathbf{v}\|_L=\sqrt{\langle\mathbf{v},\mathbf{v}\rangle_L}$,
and
\begin{align}
\exp^L_x(\mathbf{v})
&=
\cosh\!\bigl(\sqrt{c}\,\|\mathbf{v}\|_L\bigr)\,x
+
\frac{\sinh\!\bigl(\sqrt{c}\,\|\mathbf{v}\|_L\bigr)}{\sqrt{c}\,\|\mathbf{v}\|_L}\,\mathbf{v},
\label{eq:lorentz-exp}\\
\log^L_x(y)
&=
\frac{\operatorname{arcosh}\!\bigl(-c\,\langle x,y\rangle_L\bigr)}
{\sinh\!\bigl(\operatorname{arcosh}(-c\,\langle x,y\rangle_L)\bigr)}
\,\bigl(y+c\,\langle x,y\rangle_L\,x\bigr).
\label{eq:lorentz-log}
\end{align}
Equation~\eqref{eq:lorentz-exp} reduces to a hyperbolic rotation of the pair
$(x,\mathbf{v}/\|\mathbf{v}\|_L)$ inside the two-dimensional plane that they
span; Equation~\eqref{eq:lorentz-log} expresses $\log^L_x(y)$ as the component
of $y$ orthogonal to $x$ in the Lorentz inner product, rescaled to have
$\|\log^L_x(y)\|_L=d_c(x,y)$.

In the Poincar\'e ball, the corresponding formulas are most compactly written
through M\"obius addition, the curved analogue of vector addition in the
ball~\citep{ganea2018hyperbolic}:
\begin{equation}
\mathbf{x}\oplus_c\mathbf{y}
=
\frac{(1+2c\langle\mathbf{x},\mathbf{y}\rangle+c\|\mathbf{y}\|_2^2)\mathbf{x}+(1-c\|\mathbf{x}\|_2^2)\mathbf{y}}
{1+2c\langle\mathbf{x},\mathbf{y}\rangle+c^2\|\mathbf{x}\|_2^2\|\mathbf{y}\|_2^2}.
\end{equation}
The exponential and logarithmic maps at $\mathbf{x}\in\mathbb{D}^d_c$ then
take the form
\begin{align}
\exp^{\mathbb{D}}_{\mathbf{x}}(\mathbf{v})
&=
\mathbf{x}\oplus_c
\tanh\!\left(\frac{\sqrt{c}\,\lambda^c_{\mathbf{x}}\|\mathbf{v}\|_2}{2}\right)
\frac{\mathbf{v}}{\sqrt{c}\,\|\mathbf{v}\|_2},
\\
\log^{\mathbb{D}}_{\mathbf{x}}(\mathbf{y})
&=
\frac{2}{\sqrt{c}\,\lambda^c_{\mathbf{x}}}
\operatorname{arctanh}\!\bigl(\sqrt{c}\,\|-\mathbf{x}\oplus_c\mathbf{y}\|_2\bigr)
\frac{-\mathbf{x}\oplus_c\mathbf{y}}{\|-\mathbf{x}\oplus_c\mathbf{y}\|_2},
\end{align}
specializing at the origin to
$\exp^{\mathbb{D}}_{\mathbf{0}}(\mathbf{v})=\tanh(\sqrt{c}\|\mathbf{v}\|_2)
\mathbf{v}/(\sqrt{c}\|\mathbf{v}\|_2)$ and
$\log^{\mathbb{D}}_{\mathbf{0}}(\mathbf{y})=
\operatorname{arctanh}(\sqrt{c}\|\mathbf{y}\|_2)\mathbf{y}/(\sqrt{c}\|\mathbf{y}\|_2)$.
The two model-specific exponentials are intertwined by
$\pi\circ\exp^L_x=\exp^{\mathbb{D}}_{\pi(x)}\circ d\pi_x$ on the appropriate
tangent vectors, which is why the analyses in the paper can be conducted in
either model without changing the underlying distortion quantities.

\appsubsection{Polar Coordinates and the Law of Cosines}
\label{app:polar-law-of-cosines}

Fixing an origin $o\in\mathbb{H}^d_c$ and identifying $T_o\mathbb{H}^d_c$ with
$\mathbb{R}^d$ through a Riemannian isometry, every point $x\ne o$ has a
unique decomposition
\begin{equation}
\log_o(x)=r_x\,\mathbf{u}_x,
\qquad
r_x=d_c(o,x)\in(0,\infty),
\qquad
\mathbf{u}_x\in\mathbb{S}^{d-1},
\end{equation}
which we abbreviate as $x=(r_x,\mathbf{u}_x)$ in polar form. In Lorentz
coordinates this corresponds to
$x=(\cosh(\sqrt{c}\,r_x)/\sqrt{c},\sinh(\sqrt{c}\,r_x)\mathbf{u}_x/\sqrt{c})$
with $o=(1/\sqrt{c},\mathbf{0})$, as used in the polar parameterization
$\iota_k$ of Equation~\eqref{eq:rap-def}. The hyperbolic law of
cosines (Equation~\eqref{eq:law-of-cosines}) follows directly by computing
$\langle x,y\rangle_L$ in these coordinates and applying the distance
formula (Equation~\eqref{eq:lorentz-distance}). Decoupling $r_x$ from $\mathbf{u}_x$
in this way is what makes the proposed radial-angular projection in
Section~\ref{sec:rap} a natural construction: the radius is preserved
verbatim, and the entire compression budget is spent on $\mathbf{u}_x$.

\appsection{Notation}
\label{app:notation}

Table~\ref{tab:notation} collects the symbols used in the main text and the
proofs that follow. Throughout, italic letters denote points on the hyperbolic
manifold, while bold letters denote Euclidean, tangent, angular, or spatial
coordinate vectors and matrices.

\begin{table}[h]
\centering
\caption{Core notation used throughout the paper.}
\label{tab:notation}
\small
\begin{tabular}{>{\raggedright\arraybackslash}p{2.4cm}p{9.8cm}}
\toprule
Symbol & Meaning \\
\midrule
$\mathbb{H}^d_c$ & $d$-dimensional hyperbolic space with constant curvature $-c$. \\
$c$ & Positive curvature magnitude, so the manifold curvature is $-c$. \\
$d_c(x,y)$ & Hyperbolic geodesic distance between manifold points $x$ and $y$. \\
$X=\{x_1,\ldots,x_n\}$ & Finite input point set used in the finite-set preservation problem. \\
$d,k,n$ & Ambient dimension, target dimension, and number of evaluated points. \\
$\varepsilon,\delta$ & Target distortion level and failure probability in the JL-style objective. \\
$x,y,a,\tilde a,o,\tilde{o}$ & Points on the hyperbolic manifold. \\
$o$ & Origin or anchor used to define polar coordinates. \\
$r_x,\mathbf{u}_x$ & Hyperbolic radius $d_c(o,x)$ and angular direction in $\mathbb{S}^{d-1}$. \\
$\widetilde{\mathbf{u}}_x$ & Compressed and renormalized angular direction produced by RAP. \\
$\bar{\mathbf{x}}$ & Spatial Lorentz coordinate vector of a hyperboloid point $x=(x_0,\bar{\mathbf{x}})$. \\
$U_a,U_{\tilde a}$ & Orthonormal coordinate identifications from tangent spaces to Euclidean coordinate spaces. \\
$\mathbf{P}$ & Euclidean JL matrix used to sketch tangent or angular coordinates. \\
$\mathbf{A}$ & Spatial-coordinate matrix used by the fully hyperbolic comparator. \\
$\Phi$ & Generic randomized compression map in the problem definition. \\
$\Delta_X(\Phi)$ & Worst-pair relative distortion of a realized map $\Phi$ on the finite set $X$. \\
$\Pi_{\mathrm{RAP}}$ & Proposed radial-angular projection $(r_x,\mathbf{u}_x)\mapsto (r_x,\mathbf{P}\mathbf{u}_x/\|\mathbf{P}\mathbf{u}_x\|)$. \\
$\Pi_{a,\mathbf{P}}$ & Tangent-space comparator $\exp_{\tilde a}(U_{\tilde a}^{-1}\mathbf{P}U_a\log_a(\cdot))$. \\
$\Pi_{\mathbf{A}}$ & Fully hyperbolic comparator that sketches spatial coordinates and renormalizes to the hyperboloid. \\
$\mathbf{z}_x$ & Origin-log Euclidean coordinate vector $U_o\log_o(x)$ used by the Euclidean JL baseline. \\
$\Phi_{\mathrm{Euc}}$ & Euclidean JL baseline map $\Phi_{\mathrm{Euc}}(x)=\mathbf{P}\mathbf{z}_x$, whose output lives in $\mathbb{R}^k$. \\
$d_{\mathrm{Euc},\mathbf{P}}$ & Euclidean output distance used by the Euclidean JL baseline. \\
$\eta$ & Angular inner-product perturbation used in the RAP sensitivity statement. \\
\bottomrule
\end{tabular}
\end{table}

\appsection{Euclidean JL Baseline}
\label{app:euclidean-baseline}

Fix the origin or anchor $o\in\mathbb{H}^d_c$ and let
$\mathbf{P}\in\mathbb{R}^{k\times d}$ be a Gaussian JL matrix. For each
$x\in\mathbb{H}^d_c$ write $\mathbf{z}_x=U_o\log_o(x)\in\mathbb{R}^d$. The
Euclidean JL diagnostic is
\begin{equation}
\Phi_{\mathrm{Euc}}(x)=\mathbf{P}\mathbf{z}_x\in\mathbb{R}^k,
\qquad
d_{\mathrm{Euc},\mathbf{P}}(x,y)=\|\Phi_{\mathrm{Euc}}(x)-\Phi_{\mathrm{Euc}}(y)\|_2,
\end{equation}
and its reported error is
$|d_{\mathrm{Euc},\mathbf{P}}(x,y)/d_c(x,y)-1|$. The output lives in
$\mathbb{R}^k$, not in $\mathbb{H}^k_c$, so the diagnostic is intentionally
cross-geometry: it asks whether a single logarithmic chart followed by an
ordinary Euclidean sketch already approximates hyperbolic distance. The
tangent-space comparator $\Pi_{o,\mathbf{P}}$ from
Section~\ref{sec:tangent} differs in that it composes the same Euclidean JL
matrix with $\exp_{\tilde o}$ and evaluates distortion in
$\mathbb{H}^k_c$.

The two are numerically close inside a small ball: for $x,y\in B(o,R)$ with
$R\sqrt{c}$ small, normal-coordinate comparison gives
$d_c(x,y)=\|\mathbf{z}_x-\mathbf{z}_y\|(1+O(cR^2))$, and the target-side
exponential adds another $O(c{R'}^2)$ term. As the radius grows, the
logarithmic chart ceases to be a faithful linearization, and the Euclidean
diagnostic increasingly mismeasures the underlying hyperbolic distance, while
the tangent-space comparator at least evaluates its output in the correct
geometry.

\appsection{Comparison with Learned Hyperbolic Dimensionality Reduction}
\label{app:learned-dr-comparison}

HoroPCA and nested hyperbolic spaces are the closest hyperbolic
dimensionality-reduction methods to our setting, but they solve a different
problem from an oblivious JL-style sketch. HoroPCA learns hyperbolic
principal directions using ideal points, Busemann coordinates, and
horospherical projections \citep{chami2021horopca}. Nested hyperbolic spaces
learn or select a low-dimensional embedded hyperboloid and use the resulting
projection as both a dimensionality-reduction primitive and a trainable
component of hyperbolic neural networks \citep{fan2022nested}. Both methods
can be strong reconstruction or visualization baselines when fitted to a
particular dataset, but their fitted map changes with the dataset. RAP, in
contrast, samples a single random map independently of the data and asks for
finite-set distance preservation.

\begin{table}[h]
\centering
\caption{Positioning relative to data-dependent hyperbolic dimensionality
reduction. ``JL-style'' means an oblivious finite-set distance-preservation
statement for a randomly sampled map.}
\label{tab:learned-dr-comparison}
\footnotesize
\setlength{\tabcolsep}{4pt}
\renewcommand{\arraystretch}{1.05}
\begin{tabularx}{\textwidth}{@{}p{2.45cm}p{3.25cm}p{2.45cm}X@{}}
\toprule
Method & Fitted objective / map & Oblivious JL-style? & Role relative to RAP \\
\midrule
Tangent PCA / PGA \citep{fletcher2004principal}
& Variance or reconstruction in a tangent/geodesic model fitted at a base point
& No: data-dependent principal directions
& Useful classical reference, but inherits locality and base-point dependence. \\
\addlinespace[2pt]
HoroPCA \citep{chami2021horopca}
& Hyperbolic PCA via fitted ideal points, Busemann coordinates, and horospherical projections
& No: fits directions to the dataset
& Strong learned dimensionality-reduction baseline for reconstruction or visualization, but not an oblivious random projection. \\
\addlinespace[2pt]
Nested hyperbolic spaces \citep{fan2022nested}
& Reconstruction through a learned lower-dimensional hyperboloid embedded in the original one
& No: learns projection parameters
& Closest learned hyperbolic-to-hyperbolic comparator, but evaluates a fitted submanifold objective. \\
\addlinespace[2pt]
\rowcolor{oursblue}
RAP (ours)
& Finite-set distance preservation by keeping radius and randomly sketching angular directions
& Yes: no fitting, one random map
& Targets the JL-style primitive for arbitrary sufficiently regular finite sets. \\
\bottomrule
\end{tabularx}
\end{table}

\begingroup
\paragraph{Geometric distinction.}
Half-space \citep{benjamini2009dimension,deng2026locality} fixes height but
does not generally preserve distance to a designated finite origin.
At curvature $-1$, the distance $r$ from
$(z,\mathbf{x})$ to $(1,\mathbf{0})$ satisfies
$\cosh r=(z^2+\|\mathbf{x}\|^2+1)/(2z)$, so projection-induced changes
in the horizontal norm can alter the radius. This is analogous to the
radial-drift mechanism of the fully hyperbolic baseline.
RAP instead fixes the radius and normalizes the projected direction,
preserving radii and distances along every ray exactly. When radius
encodes hierarchical depth, this retains depth rankings and radial
threshold decisions. RAP's intrinsic polar formulation admits native
Lorentz and Poincar\'e ball implementations; applying the published
half-space formulas to these models requires coordinate conversion or
an equivalent reformulation.
\par
\endgroup

For the learned methods in Table~\ref{tab:learned-dr-comparison}, the distinction
between fitted maps and oblivious sketches matters for empirical comparisons.
If HoroPCA or nested
hyperbolic spaces are fitted separately on each benchmark, they are allowed to
use the entire point cloud to choose a low-dimensional representation; RAP and
the other projection baselines in Section~\ref{sec:experiments} are not. Such
an experiment would answer a useful but different reconstruction question. On
the other hand, forcing HoroPCA or nested hyperbolic spaces to use random
unfitted directions would no longer evaluate the methods as proposed. We
therefore treat them as related data-dependent dimensionality-reduction
methods rather than main baselines for the JL-style problem.

\appsection{Experimental Protocol Details}
\label{app:experimental-protocol}

\paragraph{Datasets.}
The \emph{broad-radius benchmark} samples origin-centered hyperbolic points
in $\mathbb{H}^{128}$ with radial scales $\{0.5, 1.0, 2.0\}$; for each seed
we average over the three scales before computing the reported mean and
standard deviation. The \emph{balanced tree} embeds a $4$-ary depth-$5$
tree in $\mathbb{H}^{50}$ and subsamples $n\in\{128,256,512\}$ nodes
uniformly at random per seed. The \emph{clustered Gaussian} dataset places
eight cluster centers at different radii and angular directions in
$\mathbb{H}^{50}$, with $n/8$ points per cluster sampled from a wrapped
hyperbolic Gaussian around each center. The \emph{WordNet} dataset uses the
noun hierarchy rooted at \texttt{animal.n.01}, varying the retained hyponym
depth and capping each subtree at $512$ nodes via deterministic subsampling.
WordNet is distributed under the Princeton WordNet 3.0 license; we use only
synset identifiers and hyponym edges.

\paragraph{Comparison protocol.}
Each method uses an independent Gaussian projection matrix per seed; all
four methods on a given seed share the same source point set so that the
comparison is paired. Curvature is fixed to $c=1$ throughout the main
experiments; the curvature sensitivity of the local tangent-space
comparator is treated separately below in
Appendix~\ref{app:supplementary-robustness}.

\paragraph{Choice of mean over worst-pair distortion.}
The worst-pair distortion $\Delta_X(\Phi)$ defined in
Section~\ref{sec:problem-definition} is the natural target of the JL-style
guarantee, but it is sensitive to a single nearly-coincident pair under
finite sampling. We therefore report the mean relative distortion over
nonzero-distance pairs in Tables~\ref{tab:summary} and~\ref{tab:wordnet}.
Appendix~\ref{app:exact-and-scale} reports exact all-pair maxima for the three hyperbolic
methods, while Appendix~\ref{app:bound-validation-diagnostic} provides a
separate curvature-scaling calibration for the local tangent comparator.

\paragraph{Compute resources.}
The core synthetic and WordNet evaluations are lightweight CPU workloads and
involve no model fitting. The largest stress test projects one million points
from $d=4096$ to $k=128$ on one A40 GPU; one projection-and-evaluation run
takes about $11$ seconds and peaks at $15.8$ GiB. The vision--language study
uses frozen representations, so its reported cost is projection and metric
evaluation rather than training.

\appsection{Supplementary Robustness Experiments}
\label{app:supplementary-robustness}

Across all robustness checks reported in this appendix, the ranking of the
four projection rules established in Tables~\ref{tab:summary} and
\ref{tab:wordnet} is preserved, and RAP's advantage extends past the
regimes the theory directly characterizes. Specifically: (i) the
fully hyperbolic comparator has a pronounced nearest-pair sensitivity
consistent with the $\operatorname{arcosh}$ conditioning mechanism
(Appendix~\ref{app:pairwise-distance-diagnostic}); (ii) RAP preserves the
per-node radial coordinate exactly on the WordNet subtree, while the
comparators introduce systematic drift
(Appendix~\ref{app:radius-preservation-diagnostic}); (iii) the
tangent-space mean distortion decreases with $k$ while its tail grows
with local radius (Appendix~\ref{app:local-scaling-diagnostic}); (iv) a
curvature-scaling calibration separates the nominal $5cR^2$ trend from
small-radius near-pair sensitivity
(Appendix~\ref{app:bound-validation-diagnostic}); (v) the
tangent-space comparator is stable under anchor shifts up to a
geodesic distance of order $1/\sqrt{c}$
(Appendix~\ref{app:anchor-sensitivity-diagnostic}); (vi) the mean
relative distortion concentrates as $n$ grows, consistent with the
standard JL averaging behavior
(Appendix~\ref{app:sample-size-diagnostic}). Empirical
robustness therefore extends past the strict envelope of
Theorem~\ref{prop:local-jl} and the explicit conditioning factors in the
finite-set RAP guarantee.

\appsubsection{Pairwise-Distance Conditioning of the Fully Hyperbolic Comparator}
\label{app:pairwise-distance-diagnostic}

This appendix records the experimental setup and full numerical detail
behind Figure~\ref{fig:pairwise-distance-diagnostic}. We run the broad-radius synthetic
benchmark with $n=256$, ambient dimension $128$, target dimension $k=32$,
curvature $c=1$, and radial scale $2.0$. For each of $20$ seeds, we sort
all nonzero pairs by their original distance $d_c(x,y)$ and divide them
into five equal-frequency bins, then compute mean and P90 relative
distortion within each bin. P90 is the $90$th percentile of
$|d_c(\Pi x_i,\Pi x_j)/d_c(x_i,x_j)-1|$ within a bin, computed separately
for each seed before aggregation. The pattern in
Figure~\ref{fig:pairwise-distance-diagnostic} matches
Lemma~\ref{lem:arcosh-conditioning}: nearby pairs have
$-c\langle x,y\rangle_L=\cosh(\sqrt{c}\,d_c(x,y))$ close to $1$ and
therefore lie in the high-sensitivity region of $\operatorname{arcosh}$,
so small Lorentz-product perturbations get amplified into large distance
errors specifically on those pairs.
Table~\ref{tab:pairwise-distance-ratio} reports the nearest-bin and
farthest-bin mean distortion for every scheme.

\begin{table}[ht]
\centering
\caption{Nearest-pair versus farthest-pair mean distortion. Standard
deviations are computed over $20$ seeds.}
\label{tab:pairwise-distance-ratio}
\small
\begin{tabular}{lcc}
\toprule
Scheme & Nearest pairs & Farthest pairs \\
\midrule
Euclidean JL & $\ms{0.241}{0.026}$ & $\ms{0.240}{0.042}$ \\
Local tangent & $\ms{0.251}{0.028}$ & $\ms{0.234}{0.052}$ \\
Fully hyperbolic & $\ms{0.186}{0.016}$ & $\ms{0.062}{0.008}$ \\
 \ourscell{\textbf{RAP (Ours)}} & \ourscell{$\bms{0.040}{0.002}$} & \ourscell{$\bms{0.023}{0.001}$} \\
\bottomrule
\end{tabular}
\end{table}

\appsubsection{Per-Node Radial Preservation on WordNet}
\label{app:radius-preservation-diagnostic}

This appendix provides visualization details and per-node measurements
for the WordNet comparison in Section~\ref{sec:analysis-discussion}.
We use the same WordNet subtree at $k=32$.

In the two-dimensional Poincare view in
Figure~\ref{fig:wordnet-embedding-visualization}, each node's distance from
the origin is set to its full hyperbolic radius, and its angular position
is given by the leading uncentered principal-plane direction of the
projected embedding. In the two comparator panels, nodes from distinct
hyponym depths overlap in radius, blurring the depth coloring across rings.
The RAP panel reproduces the original concentric layering and isolates
depth to a single shell. The black circle marks the display zoom window
rather than the unit Poincare boundary. The Euclidean diagnostic is omitted
because its output does not lie in $\mathbb{H}^k_c$.

\begin{figure}[ht]
\centering
\includegraphics[width=\textwidth]{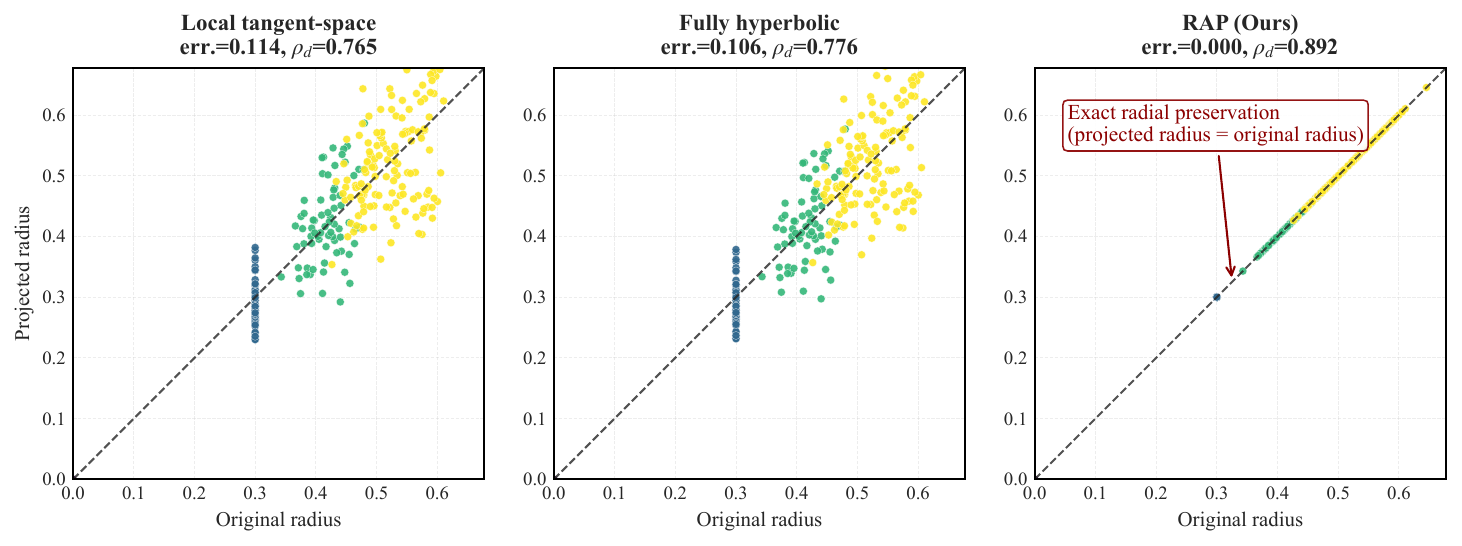}
\caption{Radius preservation on the WordNet subtree at $k=32$. Points are
colored by WordNet depth. The diagonal indicates exact radial preservation.}
\label{fig:wordnet-radius-preservation}
\end{figure}

Figure~\ref{fig:wordnet-radius-preservation} plots each node's original
hyperbolic radius against its projected radius, with points colored by
WordNet depth. RAP lies on the diagonal by construction, whereas the
tangent-space and fully hyperbolic comparators introduce systematic radial
drift across depths. Their mean relative radius errors are $0.114$ and
$0.106$, respectively, compared with $0.000$ for RAP.

\appsubsection{Local-Radius Behavior of the Tangent-Space Comparator}
\label{app:local-scaling-diagnostic}

\begin{figure}[ht]
\centering
\includegraphics[width=0.86\textwidth]{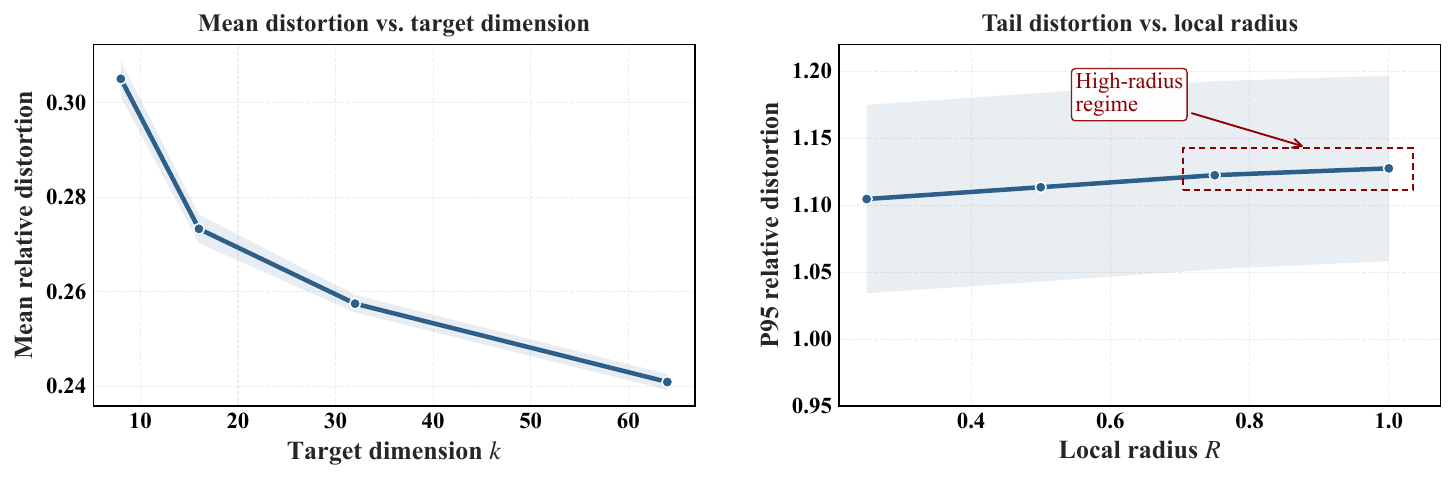}
\caption{Tangent-space tail distortion as a function of the local radius $R$.
Shaded bands show $95\%$ confidence intervals ($t$-distribution) over $20$
seeds. Mean distortion decreases with target dimension; tail distortion rises
with local radius, matching the $\varepsilon+O(cR^2)$ envelope of
Theorem~\ref{prop:local-jl}.}
\label{fig:local-scaling}
\end{figure}

This diagnostic isolates the tangent-space tail behavior. We sample
synthetic balls of increasing local radius $R$ and evaluate the local
tangent-space projection across target dimensions. Figure~\ref{fig:local-scaling}
shows that mean distortion decreases with the target dimension $k$, while
tail distortion grows as $R$ increases. Both effects are consistent with
Theorem~\ref{prop:local-jl}: the JL term shrinks with $k$, but the curvature
term $cR^2$ is independent of $k$ and tracks the local radius. The figure is
therefore a calibration of the local-JL guarantee rather than a cross-method
comparison.

\FloatBarrier
\newpage
\appsubsection{Tangent-Space Curvature-Scaling Calibration}
\label{app:bound-validation-diagnostic}

\begin{figure}[ht]
\centering
\includegraphics[width=0.62\textwidth]{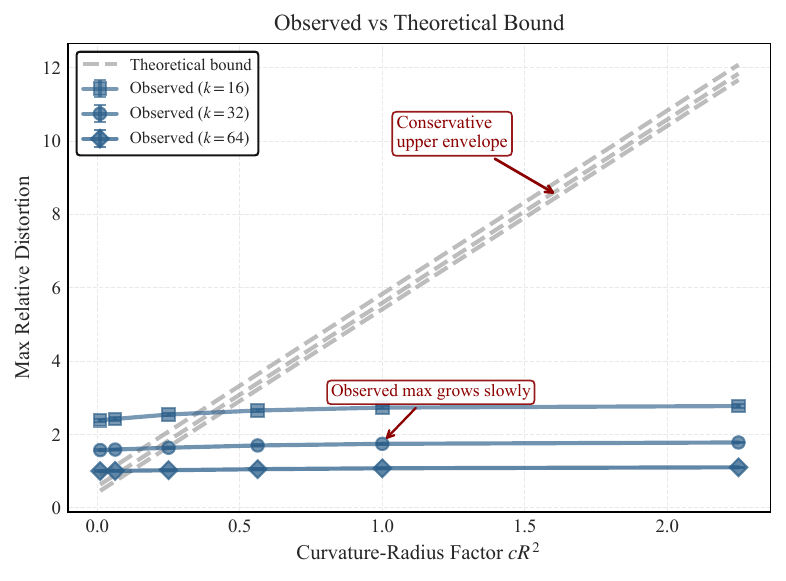}
\caption{Observed maximum distortion and the nominal curvature-scaling curve
$\varepsilon+5cR^2$ (dashed gray) for $k\in\{16,32,64\}$ over $20$ seeds.
The curve is a calibration reference, not a certified per-run bound; the
small-radius maximum is sensitive to near-pair effects.}
\label{fig:bound-validation}
\end{figure}

We evaluate the tangent-space scheme on synthetic balls of varying local
radius $R$ and compare the observed worst-pair distortion with the nominal
curve $\varepsilon+5cR^2$ derived in Appendix~\ref{app:full-proof}. This plot
does not certify the theorem seed by seed: the theorem also conditions on the
realized tangent-space JL event and on source and target locality. We therefore
use Figure~\ref{fig:bound-validation} only to calibrate the $cR^2$ trend. The
observed maximum grows more slowly than the nominal curve at moderate radii;
at small radii it is dominated by near-coincident pairs and separation
regularity. The constant $C_1=5C_0$ is deliberately conservative.

\FloatBarrier
\appsubsection{Anchor Sensitivity of the Tangent-Space Comparator}
\label{app:anchor-sensitivity-diagnostic}

\begin{figure}[ht]
\centering
\includegraphics[width=0.76\textwidth]{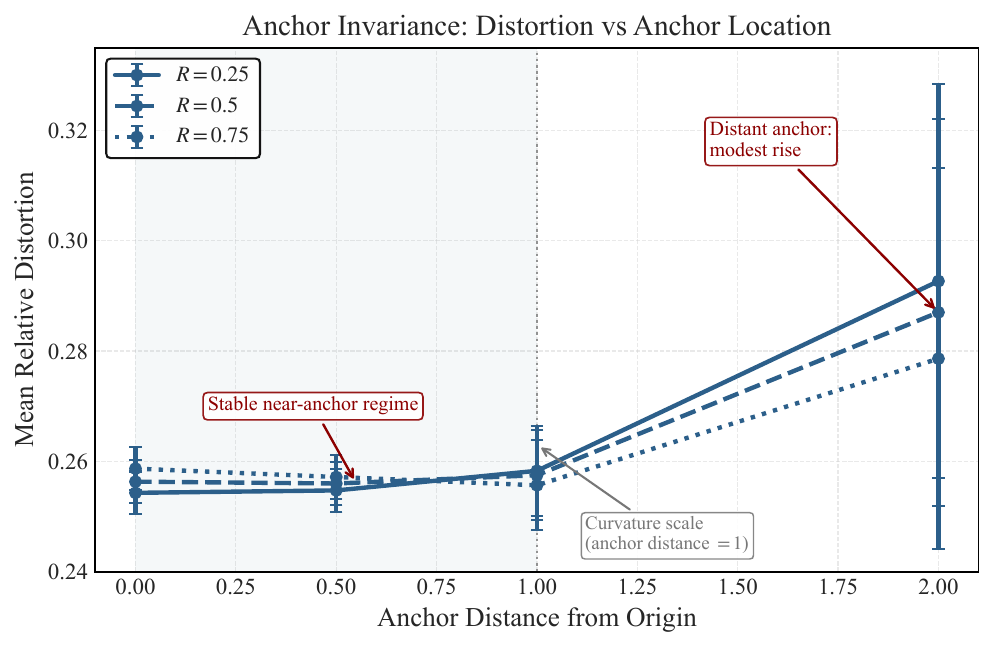}
\caption{Anchor invariance: mean distortion as a function of the anchor's
geodesic distance from the origin, for local radii $R\in\{0.25, 0.5, 0.75\}$
and $k=32$ over $20$ seeds. Distortion remains stable for anchor distances up
to $1.0$ and increases modestly at distance $2.0$.}
\label{fig:anchor-invariance}
\end{figure}

This experiment probes whether the tangent-space comparator is sensitive
to the choice of anchor, which is the only externally specified geometric
parameter besides the JL matrix. We sweep the anchor's geodesic distance
from the origin from $0.0$ to $2.0$, evaluate at three local radii
$R\in\{0.25, 0.5, 0.75\}$, and fix $k=32$. Figure~\ref{fig:anchor-invariance}
shows that mean distortion is essentially flat up to anchor distance
$\sim 1.0$ (the curvature length scale $1/\sqrt{c}=1$ at $c=1$) and grows
only modestly at distance $2.0$. The pattern matches the local-JL
prediction that the bound depends on the local-radius parameter $R$
rather than on the anchor's absolute position. This experiment does not
replace a RAP-specific comparison among the canonical origin, a Fr\'echet
mean, and a semantic root; we treat that comparison as future work.

\FloatBarrier
\appsubsection{Sample-Size Behavior of Mean Distortion}
\label{app:sample-size-diagnostic}

\begin{figure}[!ht]
\centering
\includegraphics[width=0.72\textwidth]{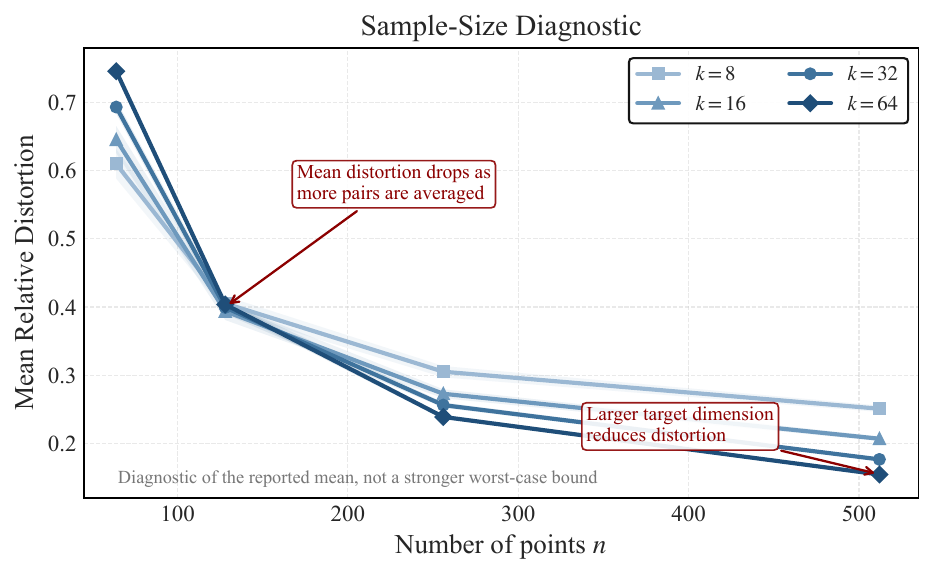}
\caption{Supplementary sample-size diagnostic for the tangent-space comparator.
Mean distortion is shown for $k\in\{8,16,32,64\}$ with $R=0.5$ over $20$ seeds.
The downward trend reflects concentration of the reported sample average over
more pairs, rather than a stronger worst-case preservation guarantee.}
\label{fig:sample-scaling}
\end{figure}

This experiment isolates the tangent-space comparator on local balls of
radius $R=0.5$ and varies the number of evaluated points $n$ and the
target dimension $k\in\{8,16,32,64\}$. Figure~\ref{fig:sample-scaling}
shows that the mean relative distortion reported in the main tables
decreases as $n$ grows. The effect is a concentration effect of the
reported sample average rather than a strengthening of the underlying
guarantee: the worst-pair distortion, which the theorem actually controls,
is by definition non-decreasing in $n$, while the mean over $\binom{n}{2}$
pair-wise relative errors concentrates around its expectation as more
pairs are averaged. The diagnostic clarifies the metric used in the main
tables and shows that the $n$-trend is the standard JL averaging behavior,
not an artifact specific to hyperbolic projection.

\FloatBarrier
\newpage
\appsubsection{Modern Vision--Language Evaluation Details}
\label{app:vlm-evaluation}

The frozen-representation evaluation uses released embeddings from MERU
ViT-S, ViT-B, and ViT-L \citep{desai2023meru} and HyCoCLIP ViT-S and ViT-B
\citep{pal2024compositional}. Each representation contains $512$ Lorentz
spatial coordinates at curvature $c=0.1$; the time coordinate is reconstructed
from the hyperboloid constraint, giving a $513$-coordinate point. The input
audit covers all $150$ feature files and gives a maximum hyperboloid-constraint
error of $2.9\times10^{-6}$.

We evaluate $13$ classification datasets: FGVC-Aircraft
\citep{maji2013fine}, Caltech-101 \citep{feifei2004learning},
CIFAR-10 and CIFAR-100 \citep{krizhevsky2009learning},
CLEVR-Counts \citep{johnson2017clevr}, Country211 \citep{radford2021learning},
Describable Textures Dataset (DTD) \citep{cimpoi2014describing},
EuroSAT \citep{helber2019eurosat}, Oxford Flowers-102
\citep{nilsback2008automated}, Food-101 \citep{bossard2014food},
MNIST \citep{lecun1998gradient}, PatchCamelyon (PCam)
\citep{veeling2018rotation}, and Rendered SST-2
\citep{socher2013recursive,radford2021learning}.
We also evaluate bidirectional image--text retrieval on COCO
\citep{lin2014microsoft} and Flickr30K \citep{young2014image}.
Across the five backbones, these datasets produce $65$ classification and
$10$ retrieval backbone--dataset cells.

RAP and the two vanilla hyperbolic projections use target spatial
dimensions $k\in\{64,128,256\}$ and five paired projection seeds. All methods
receive the same frozen features in each cell. Classification uses the
original zero-shot accuracy interface. Retrieval reports the mean of
text-to-image and image-to-text Recall@$1$, Recall@$5$, and Recall@$10$.
Table~\ref{tab:vlm} compares RAP with the tangent-space vanilla method
(Section~\ref{sec:tangent}) and the fully hyperbolic vanilla method
(Section~\ref{sec:lorentz}). For each task, method,
target dimension, and seed, we take an equally weighted mean over the
backbone--dataset cells. We report the mean and sample standard deviation
of these five seed-level macro scores, rather than the variation across
datasets. Relative gains use the ratio of unrounded macro means, with the
larger mean of the two hyperbolic baselines as the denominator; they are
not averages of per-cell relative gains. The fully hyperbolic baseline has
the higher macro mean in all six task--dimension settings. RAP's relative
classification gains are $1.74\%$, $0.43\%$, and $0.49\%$ at
$k=64,128,256$, respectively; the corresponding retrieval gains are
$20.14\%$, $11.00\%$, and $5.50\%$.

\begin{table}[ht]
\centering
\caption{Frozen-representation evaluation against vanilla hyperbolic projections.
Classification reports accuracy (\%); retrieval averages Recall@$\{1,5,10\}$
(\%) in both directions. Entries are mean $\pm$ sample standard deviation
over five paired projection seeds, each macro-averaging $65$ classification
or $10$ retrieval backbone--dataset cells. Higher is better; bold marks the
best compressed mean. Relative gain is $100(S_{\mathrm{RAP}}/S_{\mathrm{best}}-1)$,
where $S_{\mathrm{best}}$ is the larger unrounded macro mean of the two
hyperbolic baselines. Uncompressed reference scores are $26.21$ for
classification and $31.44$ for retrieval.}
\label{tab:vlm}
\small
\setlength{\tabcolsep}{3pt}
\renewcommand{\arraystretch}{1.05}
\begin{tabular*}{\linewidth}{@{\extracolsep{\fill}}lrrrrr@{}}
\toprule
Task & $k$ & \shortstack{Tangent-space\\vanilla} & \shortstack{Fully hyperbolic\\vanilla} & \ourscell{RAP} & Rel. gain \\
\midrule
Classification & 64  & \ms{20.12}{0.34} & \ms{20.14}{0.35} & \ourscell{\bms{20.49}{0.28}} & $+1.74\%$ \\
Classification & 128 & \ms{22.64}{0.31} & \ms{22.65}{0.31} & \ourscell{\bms{22.74}{0.35}} & $+0.43\%$ \\
Classification & 256 & \ms{24.23}{0.19} & \ms{24.24}{0.20} & \ourscell{\bms{24.36}{0.18}} & $+0.49\%$ \\
\midrule
Retrieval & 64  & \ms{15.17}{0.63} & \ms{15.24}{0.63} & \ourscell{\bms{18.31}{0.14}} & $+20.14\%$ \\
Retrieval & 128 & \ms{21.03}{0.41} & \ms{21.10}{0.41} & \ourscell{\bms{23.42}{0.19}} & $+11.00\%$ \\
Retrieval & 256 & \ms{25.36}{0.43} & \ms{25.41}{0.43} & \ourscell{\bms{26.81}{0.27}} & $+5.50\%$ \\
\bottomrule
\end{tabular*}
\end{table}

The $75$ backbone--dataset cells, three hyperbolic projection methods,
three target dimensions, and five seeds yield $3{,}375$ compressed
evaluations, all of which are present in the evaluation and geometry audits. The largest output hyperboloid-constraint error is
$5.8\times10^{-6}$. RAP's radial-distance drift is at most
$1.8\times10^{-7}$ across the drift audit, consistent with exact radius
preservation up to floating-point precision. These are frozen-embedding
compression experiments rather than end-to-end retraining results.

\FloatBarrier
\appsubsection{Exact Worst-Pair and Scaling Audits}
\label{app:exact-and-scale}

\paragraph{Exact worst-pair audit.}
We repeat the broad-radius setting at $n=256$, $d=128$, $k=32$, radial scale
$2.0$, and $c=1$ with independent data and projection random streams. Every
one of the $32{,}640$ unordered pairs is evaluated for each of $20$ paired
seeds. The maximum column in Table~\ref{tab:exact-worst-pair} is the mean and
standard deviation of the per-seed global maxima, rather than a maximum over
sampled pairs.

\begin{table}[ht]
\centering
\caption{Exact all-pair distortion audit. Lower is better.}
\label{tab:exact-worst-pair}
\small
\setlength{\tabcolsep}{7pt}
\begin{tabular}{lrrrr}
\toprule
Method & Mean & P95 & P99 & Exact maximum \\
\midrule
Local tangent & 0.098 & 0.241 & 0.316 & \ms{0.483}{0.062} \\
Fully hyperbolic & 0.059 & 0.161 & 0.232 & \ms{0.428}{0.055} \\
 \ourscell{\textbf{RAP (Ours)}} & \ourscell{\textbf{0.034}} & \ourscell{\textbf{0.099}} & \ourscell{\textbf{0.149}} & \ourscell{\bms{0.335}{0.038}} \\
\bottomrule
\end{tabular}
\end{table}

RAP lowers the exact maximum by $21.8\%$ relative to the fully hyperbolic
comparator. The gain therefore extends beyond average distortion: RAP also
reduces the largest pairwise error in this all-pair evaluation, improving
both typical and worst-pair distance preservation.
The realized minimum input separation is
$0.0197\pm0.0133$ across seeds, with a minimum of $0.00228$, making explicit
the near-pair conditioning behind the tail metric.

The worst-pair advantage also holds in both distance regimes
(Table~\ref{tab:near-far-worst-pair}). For each seed, we sort all pairs by
their original hyperbolic distance and take the nearest and farthest
$20\%$, giving $6{,}528$ pairs per bin. RAP attains the lowest exact
maximum in both bins, showing that its advantage is not confined to
either nearby or distant pairs.

\begin{table}[ht]
\centering
\caption{Exact worst-pair relative distortion within the nearest and farthest
$20\%$ of input pairs, using the all-pair setup in Table~\ref{tab:exact-worst-pair}.
Entries are mean $\pm$ sample standard deviation of the per-seed bin maxima
over $20$ paired seeds. Lower is better; bold marks the best mean.}
\label{tab:near-far-worst-pair}
\small
\setlength{\tabcolsep}{7pt}
\begin{tabular}{lcc}
\toprule
Method & Near exact max. $\downarrow$ & Far exact max. $\downarrow$ \\
\midrule
Local tangent & \ms{0.459}{0.070} & \ms{0.389}{0.041} \\
Fully hyperbolic & \ms{0.426}{0.056} & \ms{0.225}{0.029} \\
\ourscell{RAP (Ours)} & \ourscell{\bms{0.328}{0.038}} & \ourscell{\bms{0.193}{0.031}} \\
\bottomrule
\end{tabular}
\end{table}

\paragraph{Ambient-dimension and point-count scaling.}
Table~\ref{tab:scale-audit} fixes $k=128$. At $n=512$, all $130{,}816$
unordered pairs are evaluated over $20$ paired seeds. For $n\ge10^5$, every
point is projected and one million distinct pairs are sampled per seed over
five paired seeds. The sampled rows therefore support large-scale mean and
tail estimates, not claims about the global worst pair among all
$O(n^2)$ pairs.

\begin{table}[ht]
\centering
\caption{High-dimensional and large-scale mean relative distortion (lower is better). ``All'' denotes exact
evaluation of every unordered pair; ``1M sampled'' denotes one million
distinct pairs per seed.}
\label{tab:scale-audit}
\small
\setlength{\tabcolsep}{6pt}
\begin{tabular}{rrrrrl}
\toprule
$d$ & $n$ & \ourscell{RAP} & Fully hyp. & Reduction & Pair evaluation \\
\midrule
768   & 512       & \ourscell{\textbf{0.0194}} & 0.0379 & 48.7\% & All \\
1,024 & 512       & \ourscell{\textbf{0.0195}} & 0.0380 & 48.7\% & All \\
4,096 & 512       & \ourscell{\textbf{0.0197}} & 0.0378 & 48.0\% & All \\
1,024 & 100,000   & \ourscell{\textbf{0.0196}} & 0.0381 & 48.5\% & 1M sampled \\
1,024 & 1,000,000 & \ourscell{\textbf{0.0196}} & 0.0381 & 48.5\% & 1M sampled \\
4,096 & 1,000,000 & \ourscell{\textbf{0.0196}} & 0.0381 & 48.5\% & 1M sampled \\
\bottomrule
\end{tabular}
\end{table}

Table~\ref{tab:highd-worst-pair} reports exact worst-pair distortion
for the full-pair $n=512$ runs. For each seed and method, we take the
maximum over all pairs and all three radial scales $\{0.5,1.0,2.0\}$
before aggregating across seeds. RAP attains the lowest exact maximum
at every tested ambient dimension, extending its worst-pair advantage
to high-dimensional representations.

\begin{table}[ht]
\centering
\caption{Exact worst-pair relative distortion at increasing ambient dimension
$d$, with $n=512$ and $k=128$. Each radial scale evaluates all $130{,}816$
unordered pairs. Entries are mean $\pm$ sample standard deviation of the
per-seed maxima over all pairs and radial scales across $20$ paired seeds.
Lower is better; bold marks the best mean.}
\label{tab:highd-worst-pair}
\small
\setlength{\tabcolsep}{7pt}
\begin{tabular}{lccc}
\toprule
Method & $d=768$ & $d=1{,}024$ & $d=4{,}096$ \\
\midrule
Local tangent & \ms{0.268}{0.017} & \ms{0.274}{0.016} & \ms{0.277}{0.028} \\
Fully hyperbolic & \ms{0.257}{0.013} & \ms{0.265}{0.018} & \ms{0.272}{0.028} \\
\ourscell{RAP (Ours)} & \ourscell{\bms{0.190}{0.012}} & \ourscell{\bms{0.195}{0.019}} & \ourscell{\bms{0.189}{0.009}} \\
\bottomrule
\end{tabular}
\end{table}

\FloatBarrier
\newpage
\begingroup
\appsubsection{\texorpdfstring{{Sensitivity to Radius and Minimum Separation}}{Sensitivity to Radius and Minimum Separation}}
\label{app:radius-separation-sensitivity}

RAP achieves the lowest mean relative distortion among all four methods
across all $50$ tested settings, maintaining its advantage as the radius
$R$ increases or the minimum separation $\tau$ decreases.

\paragraph{Experimental setup.}
We use $n=256$, $d=128$, $k=32$, $c=1$, and $20$ paired runs, with
$R\in\{1,2,4,6,8\}$ and
$\tau\in\{0.5,0.1,0.01,0.001,0.0001\}$.
Each dataset contains $128$ randomly oriented point pairs at distance
$\tau$, with equal radii $(R,R)$ or different radii $(R,R-\tau/2)$.
All methods share the same inputs and Gaussian projection matrix within
each run. We report mean relative distortion over all $32{,}640$ unordered
pairs.

\begin{figure}[!ht]
\centering
\includegraphics[width=\textwidth]{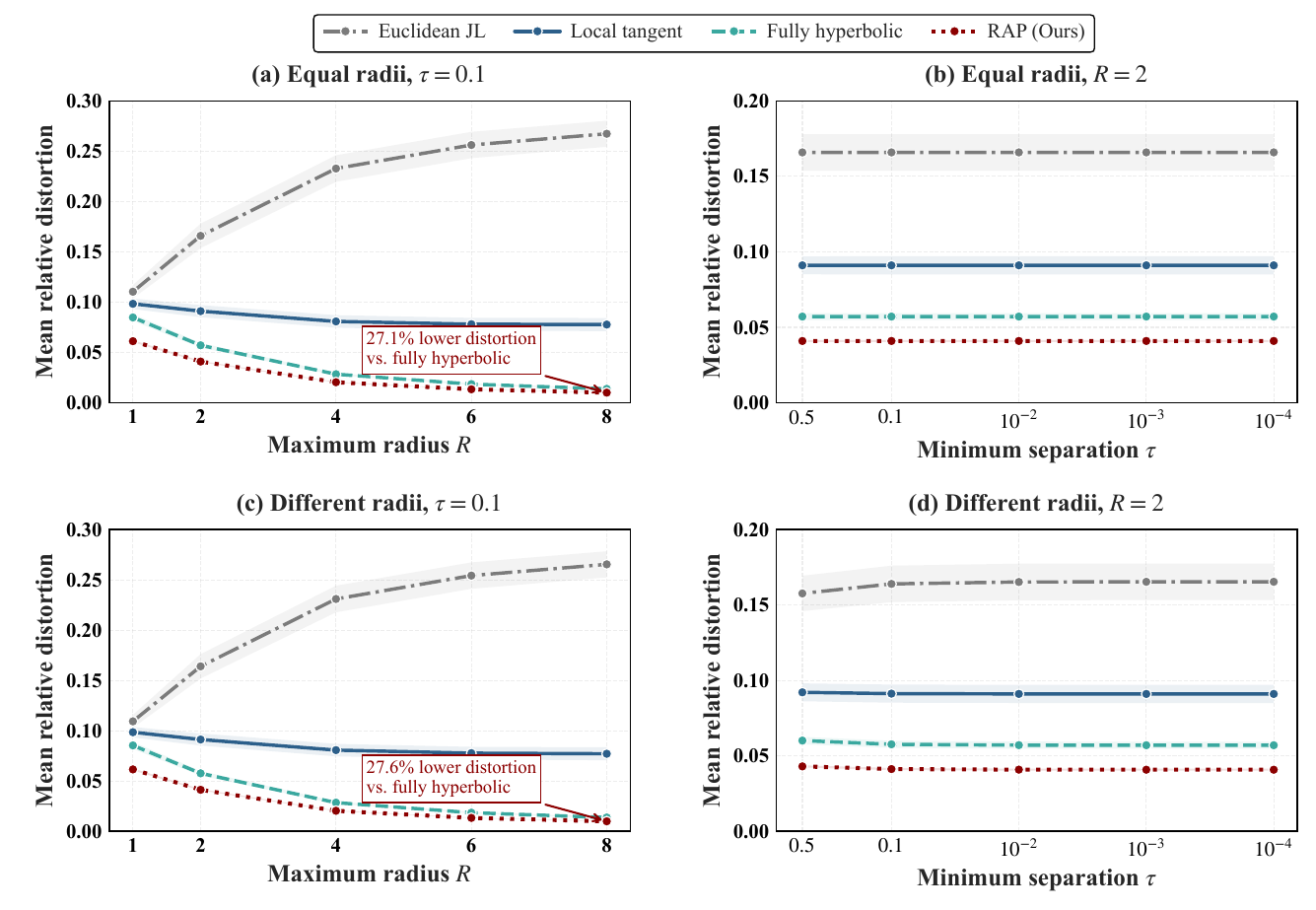}
\caption{{Sensitivity to radius and minimum separation.
Panels (a,c) increase $R$ at fixed $\tau=0.1$; panels (b,d) decrease $\tau$
at fixed $R=2$ on a logarithmic axis. The top and bottom rows use equal and
different pair radii, respectively. Lines show mean relative distortion
over all point pairs, averaged over $20$ paired runs; bands show one sample
standard deviation. At $R=8$, RAP reduces distortion relative to fully
hyperbolic by $27.1\%$ (equal radii) and $27.6\%$ (different radii).}}
\label{fig:radius-separation-sensitivity}
\end{figure}

\paragraph{Results and analysis.}
Figure~\ref{fig:radius-separation-sensitivity} shows that RAP's mean
distortion decreases from approximately $0.061$ to $0.010$ in both
constructions as $R$ increases from $1$ to $8$ at $\tau=0.1$.
At $R=2$, decreasing $\tau$ from $0.5$ to $10^{-4}$ leaves the equal-radius
mean near $0.0409$ and slightly reduces the different-radius mean from
$0.0430$ to $0.0408$.

The narrowing curve gap at large $R$ follows from the relative-error
normalization: most original pair distances grow, while absolute errors
stabilize. For equal radii over $R=4$--$8$, mean absolute errors remain
near $0.145$ for RAP and $0.204$ for fully hyperbolic. Larger distances
therefore reduce both relative errors and their vertical gap, while RAP's
relative reduction over fully hyperbolic remains $27.1\%$--$28.4\%$ across
the radius sweeps.
\par
\endgroup
\FloatBarrier

\begingroup
\appsubsection{\texorpdfstring{{Additional Tree Families and Branching Factors}}{Additional Tree Families and Branching Factors}}
\label{app:tree-family-extension}

RAP achieves the lowest mean relative distortion among all four methods in
all $30$ tested compression settings across binomial, Barab\'asi--Albert,
and $r$-ary trees, with $20.5\%$--$39.5\%$ reductions relative to
Euclidean JL.

\paragraph{Experimental setup.}
We construct five tree families with $2{,}048$ nodes each: binomial $B_{11}$,
Barab\'asi--Albert with one edge per new node ($m=1$), and breadth-first
$r$-ary trees with $r\in\{2,4,8\}$.
We use the same tree embedding procedure in $\mathbb H^{50}$ with $c=1$
and radial-step parameter $0.3$, sampling $n\in\{128,256,512\}$ nodes
and projecting to $k\in\{16,32\}$.
Each setting uses $20$ paired runs; all methods share the sampled nodes
and Gaussian projection matrix within each run. Barab\'asi--Albert topology
is regenerated per run. Mean relative distortion is evaluated over every
unordered pair against the original hyperbolic distances.

\begin{table}[!ht]
\centering
\caption{{Mean relative distance distortion on additional
tree families at $n=512$, $k=32$. Entries are mean $\pm$ sample standard
deviation over $20$ paired runs, evaluating all $130{,}816$ unordered pairs
per run. Reductions are relative to Euclidean JL; bold marks the lowest mean.}}
\label{tab:tree-family-extension}
\small
\setlength{\tabcolsep}{2.8pt}
\begin{tabular}{lccccc}
\toprule
Tree & Euclidean JL & \shortstack{Local\\tangent} & \shortstack{Fully\\hyperbolic} & \ourscell{RAP} & \shortstack{Reduction vs.\\Euclidean JL} \\
\midrule
Binomial & \ms{0.1038}{0.0077} & \ms{0.1000}{0.0065} & \ms{0.0914}{0.0051} & \ourscell{\bms{0.0704}{0.0038}} & $32.2\%$ \\
Barab\'asi--Albert & \ms{0.1022}{0.0055} & \ms{0.0994}{0.0049} & \ms{0.0924}{0.0040} & \ourscell{\bms{0.0767}{0.0100}} & $25.0\%$ \\
$r=2$ & \ms{0.1155}{0.0160} & \ms{0.0970}{0.0086} & \ms{0.0854}{0.0065} & \ourscell{\bms{0.0699}{0.0041}} & $39.4\%$ \\
$r=4$ & \ms{0.1001}{0.0072} & \ms{0.0962}{0.0060} & \ms{0.0893}{0.0053} & \ourscell{\bms{0.0710}{0.0040}} & $29.1\%$ \\
$r=8$ & \ms{0.1002}{0.0050} & \ms{0.0986}{0.0060} & \ms{0.0933}{0.0050} & \ourscell{\bms{0.0716}{0.0032}} & $28.5\%$ \\
\bottomrule
\end{tabular}
\end{table}

\paragraph{Results and analysis.}
Table~\ref{tab:tree-family-extension} shows that RAP's advantage extends
to the two imbalanced tree families, reducing mean distortion by $32.2\%$
on binomial trees and $25.0\%$ on Barab\'asi--Albert trees relative to
Euclidean JL. Across $r=2,4,8$, RAP's mean distortion stays between
$0.0699$ and $0.0716$, indicating stable average distance preservation
as the branching factor varies.
\par
\endgroup
\FloatBarrier
\appsection{Property Preservation Beyond Distance Distortion}
\label{app:property-preservation}

\appsubsection{Quantitative Analysis of Preserved Properties}
\label{app:property-analysis}

The main experiments report mean relative distance distortion. We also ask
whether the same projections preserve geometric properties that make
Euclidean random projections useful beyond distance compression:
neighborhood rankings, hierarchy/radius structure, and low-dimensional
spectral signatures. These evaluations are motivated by analyses of random
projections for manifold-modeled data, clustering, and margin-based learning
\citep{baraniuk2009random,boutsidis2010random,paul2013random}, together with
rank-based quality criteria for dimensionality reduction
\citep{lee2009quality}. All results below use $k=32$ and report
mean$\pm$std over $20$ random projection seeds.

For neighborhood structure, we compare pairwise distance rankings by
per-point Spearman correlation, kNN@10 recall, trustworthiness, and
continuity. For hierarchy structure, we measure radial error, the Spearman
correlation between tree depth and hyperbolic radius, whether the parent of a
node remains among its ten nearest neighbors, and the mean rank of the parent.
For spectral structure, we form the distance kernel $\exp(-d/\tau)$, normalize
its eigenvalues, and compare effective rank, participation ratio, and the
top-$20$ spectral profile; effective rank follows the entropy-based notion of
\citet{roy2007effective}. Euclidean JL is included as a cross-geometry
baseline, but it has no hyperbolic radius after projection, so radius-based
entries are omitted.

\begin{table}[ht]
\centering
\caption{Neighborhood and rank preservation at target dimension $k=32$. Higher values are better; bold marks the best mean in each dataset--metric block.}
\label{tab:property-rank}
\small
\setlength{\tabcolsep}{2pt}
\renewcommand{\arraystretch}{0.94}
\begin{tabularx}{\textwidth}{@{}p{2.4cm}p{2.7cm}>{\centering\arraybackslash}X>{\centering\arraybackslash}X>{\centering\arraybackslash}X>{\centering\arraybackslash}X@{}}
\toprule
Dataset & Scheme & Rank $\rho$ $\uparrow$ & kNN@10 $\uparrow$ & Trust. $\uparrow$ & Cont. $\uparrow$ \\
\midrule
Balanced tree & Euclidean JL & \ms{0.770}{0.028} & \ms{0.762}{0.013} & \ms{0.988}{0.003} & \ms{0.991}{0.002} \\
 & Local tangent-space & \ms{0.772}{0.028} & \ms{0.767}{0.014} & \ms{0.989}{0.003} & \ms{0.992}{0.002} \\
 & Fully hyperbolic & \ms{0.782}{0.028} & \ms{0.770}{0.014} & \ms{0.989}{0.003} & \bms{0.992}{0.002} \\
 & \ourscell{RAP (Ours)} & \ourscell{\bms{0.825}{0.026}} & \ourscell{\bms{0.775}{0.015}} & \ourscell{\bms{0.990}{0.003}} & \ourscell{\bms{0.992}{0.002}} \\
\addlinespace[2pt]
Clustered Gaussian & Euclidean JL & \ms{0.882}{0.037} & \ms{0.595}{0.012} & \ms{0.987}{0.001} & \bms{0.988}{0.001} \\
 & Local tangent-space & \ms{0.911}{0.031} & \ms{0.598}{0.011} & \bms{0.987}{0.001} & \bms{0.988}{0.001} \\
 & Fully hyperbolic & \ms{0.925}{0.023} & \ms{0.600}{0.011} & \bms{0.987}{0.001} & \bms{0.988}{0.001} \\
 & \ourscell{RAP (Ours)} & \ourscell{\bms{0.951}{0.014}} & \ourscell{\bms{0.602}{0.012}} & \ourscell{\bms{0.988}{0.001}} & \ourscell{\bms{0.988}{0.001}} \\
\addlinespace[2pt]
WordNet subtree & Euclidean JL & \ms{0.798}{0.023} & \ms{0.605}{0.015} & \ms{0.968}{0.003} & \ms{0.975}{0.003} \\
 & Local tangent-space & \ms{0.802}{0.023} & \ms{0.604}{0.015} & \ms{0.969}{0.003} & \ms{0.975}{0.003} \\
 & Fully hyperbolic & \ms{0.810}{0.022} & \ms{0.607}{0.015} & \ms{0.970}{0.003} & \ms{0.975}{0.003} \\
 & \ourscell{RAP (Ours)} & \ourscell{\bms{0.864}{0.013}} & \ourscell{\bms{0.623}{0.014}} & \ourscell{\bms{0.976}{0.002}} & \ourscell{\bms{0.979}{0.002}} \\
\bottomrule
\end{tabularx}
\end{table}

\begin{table}[ht]
\centering
\caption{Hierarchy and radius preservation on tree-structured data at target dimension $k=32$. Euclidean JL has no hyperbolic radius, so radius-based entries are omitted.}
\label{tab:property-hierarchy}
\small
\setlength{\tabcolsep}{2pt}
\renewcommand{\arraystretch}{0.94}
\begin{tabularx}{\textwidth}{@{}p{2.4cm}p{2.7cm}>{\centering\arraybackslash}X>{\centering\arraybackslash}X>{\centering\arraybackslash}X>{\centering\arraybackslash}X@{}}
\toprule
Dataset & Scheme & Radius err. $\downarrow$ & Depth--radius $\rho$ $\uparrow$ & Parent@10 $\uparrow$ & Parent rank $\downarrow$ \\
\midrule
Balanced tree & Euclidean JL & -- & -- & \bms{1.000}{0.001} & \ms{1.609}{0.045} \\
 & Local tangent-space & \ms{0.101}{0.011} & \ms{0.559}{0.048} & \bms{1.000}{0.001} & \ms{1.565}{0.044} \\
 & Fully hyperbolic & \ms{0.091}{0.010} & \ms{0.578}{0.045} & \bms{1.000}{0.001} & \ms{1.562}{0.047} \\
 & \ourscell{RAP (Ours)} & \ourscell{\bms{0.000}{0.000}} & \ourscell{\bms{0.720}{0.000}} & \ourscell{\bms{1.000}{0.000}} & \ourscell{\bms{1.555}{0.039}} \\
\addlinespace[2pt]
WordNet subtree & Euclidean JL & -- & -- & \ms{0.987}{0.005} & \ms{1.740}{0.067} \\
 & Local tangent-space & \ms{0.101}{0.010} & \ms{0.764}{0.033} & \bms{0.989}{0.005} & \ms{1.673}{0.071} \\
 & Fully hyperbolic & \ms{0.094}{0.009} & \ms{0.775}{0.030} & \bms{0.988}{0.005} & \ms{1.664}{0.068} \\
 & \ourscell{RAP (Ours)} & \ourscell{\bms{0.000}{0.000}} & \ourscell{\bms{0.892}{0.000}} & \ourscell{\ms{0.987}{0.004}} & \ourscell{\bms{1.626}{0.056}} \\
\bottomrule
\end{tabularx}
\end{table}

\begin{table}[ht]
\centering
\caption{Spectral-signature preservation for the distance kernel $\exp(-d/\tau)$ at target dimension $k=32$. Lower values are better.}
\label{tab:property-spectral}
\small
\setlength{\tabcolsep}{2pt}
\renewcommand{\arraystretch}{0.94}
\begin{tabularx}{\textwidth}{@{}p{2.4cm}p{2.7cm}>{\centering\arraybackslash}X>{\centering\arraybackslash}X>{\centering\arraybackslash}X@{}}
\toprule
Dataset & Scheme & Eff. rank err. $\downarrow$ & PR err. $\downarrow$ & Top-20 $L_1$ $\downarrow$ \\
\midrule
Balanced tree & Euclidean JL & \ms{0.159}{0.035} & \ms{0.105}{0.041} & \ms{0.049}{0.008} \\
 & Local tangent-space & \ms{0.104}{0.040} & \ms{0.067}{0.042} & \ms{0.038}{0.009} \\
 & Fully hyperbolic & \ms{0.103}{0.036} & \ms{0.064}{0.039} & \ms{0.037}{0.008} \\
 & \ourscell{RAP (Ours)} & \ourscell{\bms{0.055}{0.032}} & \ourscell{\bms{0.026}{0.023}} & \ourscell{\bms{0.030}{0.006}} \\
\addlinespace[2pt]
Clustered Gaussian & Euclidean JL & \ms{0.199}{0.025} & \ms{0.107}{0.035} & \ms{0.059}{0.010} \\
 & Local tangent-space & \ms{0.067}{0.035} & \ms{0.044}{0.032} & \ms{0.040}{0.011} \\
 & Fully hyperbolic & \ms{0.065}{0.026} & \ms{0.037}{0.025} & \ms{0.038}{0.009} \\
 & \ourscell{RAP (Ours)} & \ourscell{\bms{0.049}{0.025}} & \ourscell{\bms{0.015}{0.011}} & \ourscell{\bms{0.031}{0.005}} \\
\addlinespace[2pt]
WordNet subtree & Euclidean JL & \ms{0.142}{0.036} & \ms{0.082}{0.037} & \ms{0.048}{0.005} \\
 & Local tangent-space & \ms{0.108}{0.039} & \ms{0.064}{0.031} & \ms{0.041}{0.005} \\
 & Fully hyperbolic & \ms{0.107}{0.036} & \ms{0.061}{0.030} & \ms{0.041}{0.005} \\
 & \ourscell{RAP (Ours)} & \ourscell{\bms{0.072}{0.017}} & \ourscell{\bms{0.018}{0.011}} & \ourscell{\bms{0.032}{0.003}} \\
\bottomrule
\end{tabularx}
\end{table}

Overall, RAP is strongest on hierarchy-aware analyses: exact radial
preservation gives near-zero radius error and consistently high
depth--radius correlation.  The neighborhood and spectral analyses are
not formal JL guarantees, but they give a useful sanity check that RAP's
lower distance distortion also translates into more stable local order and
kernel-spectrum behavior on the tested hyperbolic representations.

\FloatBarrier
\Needspace{12\baselineskip}
\appsubsection{Concrete WordNet Hierarchy Case Study}
\label{app:wordnet-hierarchy-case-study}

\begin{figure}[ht]
\centering
\includegraphics[width=0.98\textwidth]{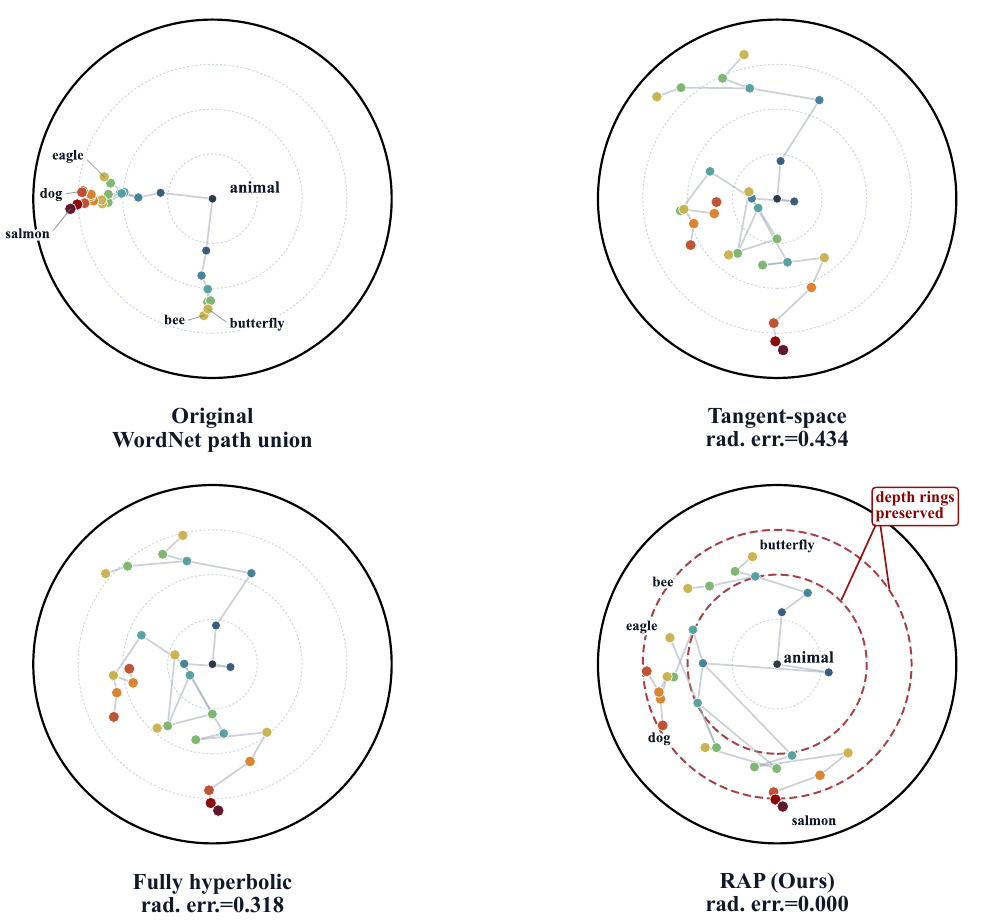}
\caption{Concrete WordNet projection case study in Poincare disks.  Nodes
follow selected hypernym paths under \texttt{animal.n.01}; node color encodes
WordNet depth, so same-level nodes share the same color.  The original panel
is a radial two-dimensional view of the input tree embedding, while the
remaining panels are actual $k=2$ hyperbolic projection outputs.  The
tangent-space and fully hyperbolic projections introduce visible radial drift;
RAP preserves the depth-carrying radial coordinate exactly, with red dashed
rings highlighting the preserved depth rings.}
\label{fig:wordnet-hierarchy-case-study}
\end{figure}

Figure~\ref{fig:wordnet-hierarchy-case-study} gives a concrete visual
case study for the hierarchy/radius interpretation.  We take a small union of
real WordNet hypernym paths rooted at \texttt{animal.n.01}, embed this path
union with the same hyperbolic tree routine used in the WordNet benchmark, and
apply the three projection rules to a two-dimensional target for visualization.
This is not a replacement for the full WordNet sweep in
Table~\ref{tab:wordnet}; rather, it makes the qualitative meaning of radial
depth explicit through familiar words such as dog, eagle, salmon, and bee.

The qualitative WordNet visualization in
Figure~\ref{fig:wordnet-embedding-visualization} complements these property
tables by showing the hierarchy rings directly. Figure~\ref{fig:wordnet-radius-preservation}
in the main text isolates the radial component by plotting each node's
original radius against its projected radius. The RAP points lie on the
diagonal by construction, while the two comparators introduce small but
systematic radial displacement.
\FloatBarrier

\appsection{Proofs for the Tangent-Space Comparator}
\label{app:full-proof}

We state and prove each step as a self-contained lemma.

\appsubsection{Source-Side Metric Comparison}
\label{app:metric-compare}

\begin{lemma}[Local metric comparison in constant curvature]
\label{lem:metric-compare}
Let $\mathbf{u},\mathbf{v}\in T_a\mathbb{H}^d_c$ with
$\|\mathbf{u}\|,\|\mathbf{v}\|\le R$ and
$R\sqrt{c}\le R_0$ for a sufficiently small universal constant $R_0\le 1$.
Then:
\begin{enumerate}[leftmargin=16pt]
\item (Global lower bound) $\|\mathbf{u}-\mathbf{v}\| \le d_c(\exp_a \mathbf{u},\exp_a \mathbf{v})$.
\item (Local upper bound) $d_c(\exp_a \mathbf{u},\exp_a \mathbf{v}) \le (1+C_0 cR^2)\|\mathbf{u}-\mathbf{v}\|$
where $C_0$ depends only on $R_0$.
\end{enumerate}
\end{lemma}

\begin{proof}
\emph{Lower bound.}
By the Cartan--Hadamard theorem, in a simply connected manifold of
nonpositive sectional curvature, the exponential map at any point is a
distance-expanding map:
$d_c(\exp_a \mathbf{u},\exp_a \mathbf{v})\ge \|\mathbf{u}-\mathbf{v}\|$. This holds
globally, not just for small $R$.

\emph{Upper bound.}
Write $\mathbf{u}=r\boldsymbol{\xi}$ and
$\mathbf{v}=s\boldsymbol{\eta}$ with
$\|\boldsymbol{\xi}\|=\|\boldsymbol{\eta}\|=1$, and let
$\theta\in[0,\pi]$ be the angle between them
($\cos\theta=\langle\boldsymbol{\xi},\boldsymbol{\eta}\rangle$).
The hyperbolic law of cosines in constant curvature $-c$ gives
\begin{equation}\label{eq:hyp-cosines}
\cosh(\sqrt{c}\, d_c(\exp_a \mathbf{u},\exp_a \mathbf{v}))
=
\cosh(\sqrt{c}\, r)\cosh(\sqrt{c}\, s)
- \sinh(\sqrt{c}\, r)\sinh(\sqrt{c}\, s)\cos\theta.
\end{equation}
Taylor-expand around $\sqrt{c}\,r$ and $\sqrt{c}\,s$ small.
Using $\cosh(t)=1+\frac{t^2}{2}+\frac{t^4}{24}+\cdots$ and
$\sinh(t)=t+\frac{t^3}{6}+\cdots$, the right-hand side becomes
\begin{align*}
&\left(1+\frac{cr^2}{2}+O(c^2r^4)\right)
\left(1+\frac{cs^2}{2}+O(c^2s^4)\right) \\
&\quad - \left(\sqrt{c}\,r+O(c^{3/2}r^3)\right)
\left(\sqrt{c}\,s+O(c^{3/2}s^3)\right)\cos\theta \\
&= 1 + \frac{c}{2}(r^2+s^2) - cs\,r\cos\theta + O(c^2R^4) \\
&= 1 + \frac{c}{2}(r^2+s^2-2rs\cos\theta) + O(c^2R^4) \\
&= 1 + \frac{c}{2}\|\mathbf{u}-\mathbf{v}\|^2 + O(c^2R^4),
\end{align*}
where we used
$\|\mathbf{u}-\mathbf{v}\|^2 = r^2+s^2-2rs\cos\theta$.

Now set $D=d_c(\exp_a \mathbf{u},\exp_a \mathbf{v})$ and
$L=\|\mathbf{u}-\mathbf{v}\|$. We have
\begin{equation}
\cosh(\sqrt{c}\,D) = 1 + \frac{c}{2}L^2 + E,
\qquad |E|\le C' c^2 R^4,
\end{equation}
for a universal $C'$. Using
$\operatorname{arcosh}(1+z)=\sqrt{2z}\,(1+O(z))$ for small $z>0$, and
noting that $z=\frac{c}{2}L^2+E$ with $L\le 2R$ gives $z\le 2cR^2+C'c^2R^4
\le 3cR^2$ (for $R_0\le 1$), we obtain
\begin{equation}
D = L\sqrt{1+\frac{2E}{cL^2}}\,(1+O(cR^2)).
\end{equation}
If $L=0$ (i.e., $\mathbf{u}=\mathbf{v}$), then $D=0$ as well, so the bound $D\le (1+C_0 cR^2)L$ holds trivially.
For $L>0$ we handle two cases.
If $L\ge c^{1/2}R^2$, then $|2E/(cL^2)|\le 2C'c^2R^4/(c\cdot cR^4)=2C'$,
which is $O(1)$ and the square root contributes only an $O(cR^2)$ factor.

If $L < c^{1/2}R^2$, a tighter estimate of the error $E$ is needed.
Returning to the Taylor expansion of the right-hand side
of Equation~\eqref{eq:hyp-cosines}, we collect all terms beyond the leading
$1+\frac{c}{2}L^2$ more carefully. The next-order contribution is
\begin{equation}
E = \frac{c^2}{4}r^2 s^2\sin^2\theta + O(c^3 R^6),
\end{equation}
where $\sin^2\theta = 1-\cos^2\theta$. We claim $r^2 s^2\sin^2\theta
\le R^2 L^2$. Indeed, $rs|\sin\theta|$ is twice the area of the triangle
with sides $r,s$ and included angle $\theta$, while $L=\|\mathbf{u}-\mathbf{v}\|$ is the
third side, so $rs|\sin\theta| = rL\sin\alpha$ for some angle $\alpha$,
giving $rs|\sin\theta|\le rL\le RL$. Hence
\begin{equation}
|E| \le \tfrac{c^2 R^2}{4} L^2 + O(c^3 R^6).
\end{equation}
Substituting back, $\frac{2|E|}{cL^2} \le \frac{cR^2}{2}+O(c^2 R^4)$,
which is $O(cR^2)$ uniformly for $R\sqrt{c}\le R_0\le 1$, regardless
of the size of $L$. With this refined bound, the identity
$\cosh(\sqrt{c}\,D) = 1+\frac{c}{2}L^2(1+O(cR^2))$ holds in both regimes.
Applying $\operatorname{arcosh}(1+z)=\sqrt{2z}\,(1+O(z))$ with
$z=\frac{c}{2}L^2(1+O(cR^2))=O(cR^2)$ gives $D=L(1+O(cR^2))$,
hence $D\le (1+C_0 cR^2)L$ with a universal $C_0$
(which absorbs both the refined error $E$ from the Taylor expansion
and the remainder from the arcosh inversion).
\end{proof}

\appsubsection{JL with the Origin Included}

\begin{lemma}[Euclidean JL with norm control]
\label{lem:jl-origin}
Let $\{\mathbf{u}_1,\dots,\mathbf{u}_n\}\subset\mathbb{R}^d$ and let
$\mathbf{P}\in\mathbb{R}^{k\times d}$ be a standard JL matrix with
$k\ge C\varepsilon^{-2}\log((n+1)/\delta)$. Then with probability at least
$1-\delta$, the JL event holds simultaneously for the augmented set
$\{0,\mathbf{u}_1,\dots,\mathbf{u}_n\}$. In particular, for all $i$:
\begin{equation}
\|\mathbf{P}\mathbf{u}_i\| = \|\mathbf{P}\mathbf{u}_i - \mathbf{P}\cdot 0\|
\le (1+\varepsilon)\|\mathbf{u}_i\|.
\end{equation}
\end{lemma}

\begin{proof}
The augmented set has $n+1$ points, hence $\binom{n+1}{2}\le n^2$ pairs.
A standard JL matrix with $k\ge C\varepsilon^{-2}\log(n^2/\delta)
= C\varepsilon^{-2}(2\log n + \log(1/\delta))$ preserves all pairwise
distances with probability $\ge 1-\delta$, by a union bound over all pairs.
The norm bound follows from the pair $(0,\mathbf{u}_i)$.
\end{proof}

\appsubsection{Target-Side Comparison}

\begin{lemma}[Target-side metric comparison]
\label{lem:target-compare}
Under the assumption $R'=(1+\varepsilon)R\le R_1/\sqrt{c}$, the same bounds
as Lemma~\ref{lem:metric-compare} apply in $\mathbb{H}^k_c$ around
$\tilde a$ for vectors of norm at most $R'$:
\begin{equation}
\|\mathbf{w}-\mathbf{w}'\| \le d_c(\exp_{\tilde a}\mathbf{w},\exp_{\tilde a}\mathbf{w}')
\le (1+C_0 c{R'}^2)\|\mathbf{w}-\mathbf{w}'\|.
\end{equation}
\end{lemma}

\begin{proof}
Lemma~\ref{lem:metric-compare} applies to any hyperbolic space of constant
curvature $-c$, regardless of dimension. The target space $\mathbb{H}^k_c$
has the same curvature, so the same derivation holds with $R'$ in place of $R$.
\end{proof}

\begin{remark}[Explicit values of $C_0$ and $R_0$]
\label{rem:constants}
Tracing through the Taylor expansion in
Lemma~\ref{lem:metric-compare}, the universal constant $C'$ in the error
bound $|E|\le \frac{c^2R^2}{4}L^2 + O(c^3R^6)$ satisfies $C'\le 1$ for
$R_0\le 1$. The arcosh inversion step introduces a further $(1+O(cR^2))$
factor. Taking $R_0 = 1/(2\sqrt{c})$ (i.e., $R\sqrt{c}\le 1/2$) keeps
all higher-order remainders below their leading terms, and one may verify
that $C_0=1$ suffices in this regime. Concretely, the condition
$R\sqrt{c}\le 1/2$ means the geodesic ball radius is at most half the
curvature length scale $1/\sqrt{c}$. For the experiments in this paper
($c=1$, $R\le 1.5$), the value $R\sqrt{c}=1.5$ exceeds this strict
threshold; nonetheless the bound remains empirically conservative, which
is consistent with the constants being pessimistic. The factor $C_1=5C_0=5$
used in the bound-validation diagnostic arises from the
error composition lemma: the source side contributes $C_0 cR^2$, the target
side contributes $C_0 c{R'}^2 \le 4C_0 cR^2$ (since $R'\le 2R$ for
$\varepsilon\le 1$), and the cross-term adds at most $C_0 cR^2$, yielding
$C_1 = 1+4+\text{cross} \le 5C_0$ as a safe upper envelope. In the code
this appears as the hardcoded multiplier $5.0$.
\end{remark}

\appsubsection{Error Composition}

\begin{lemma}[Abstract composition of distortion factors]
\label{lem:compose}
Let $\alpha,\beta\ge 0$ with $\alpha\le 1$ and $\beta\le 1$. If
\begin{equation}
\frac{1-\alpha}{1+\beta}\le \frac{d_c(\Pi x_i,\Pi x_j)}{d_c(x_i,x_j)}
\le (1+\alpha)(1+\beta),
\end{equation}
then
\begin{equation}
1-\alpha-\beta \le \frac{d_c(\Pi x_i,\Pi x_j)}{d_c(x_i,x_j)}
\le 1+\alpha+\beta+\alpha\beta.
\end{equation}
In particular, if $\alpha\le 1$, then $1+\alpha+\beta+\alpha\beta
\le 1+\alpha+2\beta$.
\end{lemma}

\begin{proof}
\emph{Lower bound.}
For $\alpha,\beta\in[0,1]$,
$\frac{1-\alpha}{1+\beta}\ge 1-\alpha-\beta$
because $(1-\alpha-\beta)(1+\beta)=1-\alpha-\beta^2-\alpha\beta
\le 1-\alpha$.

\emph{Upper bound.}
$(1+\alpha)(1+\beta)=1+\alpha+\beta+\alpha\beta$, which gives the
first form. For the simplified bound, $\alpha\beta\le\beta$ when
$\alpha\le 1$, so $1+\alpha+\beta+\alpha\beta\le 1+\alpha+2\beta$.
\end{proof}

\begin{corollary}[$C_1 = 5C_0$]
\label{cor:C1}
In the proof of Theorem~\ref{prop:local-jl}, the source-side
curvature error is $\beta_{\mathrm{src}} = C_0 cR^2$
(Lemma~\ref{lem:metric-compare}) and the target-side curvature error
is $\beta_{\mathrm{tgt}} = C_0 c{R'}^2$ with $R'=(1+\varepsilon)R\le 2R$
(Lemma~\ref{lem:target-compare}), so
$\beta_{\mathrm{tgt}}\le 4C_0 cR^2$. Applying
Lemma~\ref{lem:compose} with $\alpha=\varepsilon$ and
$\beta = \beta_{\mathrm{src}} + \beta_{\mathrm{tgt}} \le 5C_0 cR^2$
(upper-bounding the composition of the source and target distortion
by their sum plus cross-term, as in the proof of Theorem~\ref{prop:local-jl})
yields $C_1 = 5C_0$.
\end{corollary}

\noindent
Combining Lemmas~\ref{lem:metric-compare}--\ref{lem:compose} completes the
proof of Theorem~\ref{prop:local-jl}. The constant $C_1$ in the theorem equals $5C_0$,
where $C_0$ is the metric comparison constant from Lemma~\ref{lem:metric-compare}.

\appsection{Fully Hyperbolic Comparator Conditioning}
\label{app:lorentz-conditioning}

The fully hyperbolic comparator stays on the hyperboloid by renormalization, so its
weakness is not loss of the manifold constraint but that hyperbolic distance
is read off through $\operatorname{arcosh}$ of a Lorentz inner product. Two
lemmas isolate the resulting conditioning issue.

\begin{lemma}[Arcosh conditioning]
\label{lem:arcosh-conditioning}
Let $z\ge 1+\rho$ and let $|\Delta|\le \rho/2$ for some $\rho>0$. Then
\begin{equation}
\left|\operatorname{arcosh}(z+\Delta)-\operatorname{arcosh}(z)\right|
\le \frac{|\Delta|}{\sqrt{\rho(2+\rho)}}\left(1+O\!\left(\frac{|\Delta|}{\rho}\right)\right).
\end{equation}
In particular, when $z=1+\rho$ with $\rho\ll 1$, the first-order sensitivity is
of order $|\Delta|/\sqrt{\rho}$.
\end{lemma}

\begin{proof}
The derivative of $\operatorname{arcosh}$ is
\begin{equation}
\frac{d}{dz}\operatorname{arcosh}(z)=\frac{1}{\sqrt{z-1}\sqrt{z+1}}.
\end{equation}
For $t$ between $z$ and $z+\Delta$, the assumption $|\Delta|\le \rho/2$ gives
$t-1\ge \rho/2$ and $t+1\ge 2+\rho/2$. The mean value theorem therefore gives
a Lipschitz bound of order $|\Delta|/\sqrt{\rho(2+\rho)}$. Expanding the
denominator around $z=1+\rho$ yields the displayed first-order expression. For
nearby pairs in hyperbolic space, $z=-c\langle x,y\rangle_L
=\cosh(\sqrt{c}\,d_c(x,y))=1+O(c\,d_c(x,y)^2)$, so small Lorentz-product
errors are amplified by roughly the inverse distance scale.
\end{proof}

\begin{lemma}[Renormalized fully hyperbolic sketches mix radial and angular errors]
\label{lem:lorentz-mixing}
For the fully hyperbolic comparator
\begin{equation}
\Pi_{\mathbf{A}}(x)=
\left(\sqrt{\frac{1}{c}+\|\mathbf{A}\bar{\mathbf{x}}\|_2^2},\,
\mathbf{A}\bar{\mathbf{x}}\right),
\end{equation}
the perturbed Lorentz product between two projected points is
\begin{equation}
-c\langle \Pi_{\mathbf{A}}(x),\Pi_{\mathbf{A}}(y)\rangle_L
=
c\sqrt{\frac{1}{c}+\|\mathbf{A}\bar{\mathbf{x}}\|_2^2}\,
\sqrt{\frac{1}{c}+\|\mathbf{A}\bar{\mathbf{y}}\|_2^2}
-c\langle \mathbf{A}\bar{\mathbf{x}},\mathbf{A}\bar{\mathbf{y}}\rangle.
\end{equation}
Thus the same spatial sketch controls both the radial time coordinate and the
spatial angular term.
\end{lemma}

\begin{proof}
The identity follows by substituting the renormalized coordinates into the
Lorentz product
$\langle x,y\rangle_L=-x_0y_0+\langle \bar{\mathbf{x}},\bar{\mathbf{y}}\rangle$.
The square-root terms depend on the sketched norms
$\|\mathbf{A}\bar{\mathbf{x}}\|_2$ and $\|\mathbf{A}\bar{\mathbf{y}}\|_2$,
while the final term depends on the sketched Euclidean inner product. Hence a
single Euclidean sketch perturbs radial magnitude and angular correlation
simultaneously. Lemma~\ref{lem:arcosh-conditioning} then explains why these
perturbations are difficult to turn into a uniform multiplicative distance
guarantee across all distance scales.
\end{proof}

Composing the two lemmas yields a multiplicative envelope for the fully hyperbolic
comparator on a finite point set. The statement parallels the finite-set
RAP guarantee in Appendix~\ref{app:rap-advantage} and isolates the same
$1/[\sqrt{c}\,\tau\sinh(\sqrt{c}\,\tau)]$ conditioning factor.

\begin{corollary}[Fully hyperbolic comparator multiplicative envelope]
\label{cor:lorentz-envelope}
Let $X=\{x_1,\dots,x_n\}\subset\mathbb{H}^d_c$ have polar coordinates
$x_i=(r_i,\mathbf{u}_i)$ with $r_i\le R$ for all $i$ and
$d_c(x_i,x_j)\ge\tau>0$ for all $i\ne j$. Fix $\varepsilon\in(0,1]$ and
let $\mathbf{A}\in\mathbb{R}^{k\times d}$ be a Gaussian sketch with
$k\ge C\varepsilon^{-2}\log(n/\delta)$. With probability at least
$1-\delta$, the squared-norm JL event holds simultaneously for the spatial
parts and their pairwise differences:
\begin{equation}
\bigl|\|\mathbf{A}v\|_2^2-\|v\|_2^2\bigr|\le\varepsilon\|v\|_2^2
\qquad\text{for every}\quad
v\in\{\bar{\mathbf{x}}_i\}_{i=1}^n
\cup\{\bar{\mathbf{x}}_i-\bar{\mathbf{x}}_j\}_{i<j}.
\label{eq:lorentz-jl-event}
\end{equation}
If in addition
$6\,\varepsilon\,\sinh^2(\sqrt{c}\,R)
\le\tfrac{1}{2}\bigl(\cosh(\sqrt{c}\,\tau)-1\bigr)$,
then on the event in Equation~\eqref{eq:lorentz-jl-event},
\begin{equation}
\max_{i<j}
\left|
\frac{d_c(\Pi_{\mathbf{A}}x_i,\Pi_{\mathbf{A}}x_j)}{d_c(x_i,x_j)}-1
\right|
\le
\frac{12\,\varepsilon\,\sinh^2(\sqrt{c}\,R)}
{\sqrt{c}\,\tau\,\sinh(\sqrt{c}\,\tau)}.
\label{eq:lorentz-envelope}
\end{equation}
The numerator splits into two contributions of comparable order: a radial
term of order
$\varepsilon\bigl(\sinh^2(\sqrt{c}\,r_i)+\sinh^2(\sqrt{c}\,r_j)\bigr)$
from the renormalization (Lemma~\ref{lem:lorentz-mixing}) and a matching
angular term of the same form from the spatial inner-product sketch.
\end{corollary}

\begin{proof}
Let $T_i=\sqrt{1/c+\|\bar{\mathbf{x}}_i\|_2^2}=\cosh(\sqrt{c}\,r_i)/\sqrt{c}$
and $\widehat{T}_i=\sqrt{1/c+\|\mathbf{A}\bar{\mathbf{x}}_i\|_2^2}$, and
abbreviate $c_\bullet=\cosh(\sqrt{c}\,r_\bullet)$,
$s_\bullet=\sinh(\sqrt{c}\,r_\bullet)$. By Lemma~\ref{lem:lorentz-mixing},
\begin{equation}
-c\langle\Pi_{\mathbf{A}}x_i,\Pi_{\mathbf{A}}x_j\rangle_L
=
c\,\widehat{T}_i\widehat{T}_j
-c\langle\mathbf{A}\bar{\mathbf{x}}_i,\mathbf{A}\bar{\mathbf{x}}_j\rangle.
\end{equation}
Equation~\eqref{eq:lorentz-jl-event} gives
$c\,\widehat{T}_i^2=1+c\|\mathbf{A}\bar{\mathbf{x}}_i\|_2^2
=c_i^2+\varepsilon_i s_i^2$ with $|\varepsilon_i|\le\varepsilon$. Squaring,
\begin{equation}
\bigl(c\,\widehat{T}_i\widehat{T}_j\bigr)^2
=(c_i^2+\varepsilon_i s_i^2)(c_j^2+\varepsilon_j s_j^2)
=(c_i c_j)^2+\varepsilon_i c_j^2 s_i^2+\varepsilon_j c_i^2 s_j^2
+\varepsilon_i\varepsilon_j s_i^2 s_j^2.
\end{equation}
Since $c\,\widehat{T}_i\widehat{T}_j+c_i c_j\ge c_i c_j$ and
$|\varepsilon_\bullet|\le\varepsilon\le 1$, dividing the squared identity
by $c\widehat{T}_i\widehat{T}_j+c_i c_j$ yields
\begin{equation}
|\Delta^{\mathrm{rad}}_{ij}|
:=\bigl|c\,\widehat{T}_i\widehat{T}_j-c_i c_j\bigr|
\le \varepsilon
\Bigl(\tfrac{c_j s_i^2}{c_i}+\tfrac{c_i s_j^2}{c_j}\Bigr)
+\varepsilon^2\,\tfrac{s_i^2 s_j^2}{c_i c_j}.
\end{equation}
The function $r\mapsto \sinh(r)\tanh(r)$ is non-decreasing on $r\ge 0$, so
$\frac{c_j s_i^2}{c_i}+\frac{c_i s_j^2}{c_j}\le s_i^2+s_j^2$ for all
$r_i,r_j\ge 0$. Combining with
$\frac{s_i^2 s_j^2}{c_i c_j}\le s_i s_j\le\tfrac{1}{2}(s_i^2+s_j^2)$,
\begin{align}
|\Delta^{\mathrm{rad}}_{ij}|
&\le \varepsilon\bigl(\sinh^2(\sqrt{c}\,r_i)+\sinh^2(\sqrt{c}\,r_j)\bigr) \notag\\
&\quad+\tfrac{\varepsilon^2}{2}
\bigl(\sinh^2(\sqrt{c}\,r_i)+\sinh^2(\sqrt{c}\,r_j)\bigr) \notag\\
&\le \tfrac{3\varepsilon}{2}
\bigl(\sinh^2(\sqrt{c}\,r_i)+\sinh^2(\sqrt{c}\,r_j)\bigr),
\end{align}
where the last inequality uses $\varepsilon\le 1$.

For the angular term, polarization
$\langle\mathbf{A}\bar{\mathbf{x}}_i,\mathbf{A}\bar{\mathbf{x}}_j\rangle
=\bigl(\|\mathbf{A}\bar{\mathbf{x}}_i\|_2^2
+\|\mathbf{A}\bar{\mathbf{x}}_j\|_2^2
-\|\mathbf{A}(\bar{\mathbf{x}}_i-\bar{\mathbf{x}}_j)\|_2^2\bigr)/2$
combined with Equation~\eqref{eq:lorentz-jl-event} and
$\|\bar{\mathbf{x}}_i-\bar{\mathbf{x}}_j\|_2^2
\le 2(\|\bar{\mathbf{x}}_i\|_2^2+\|\bar{\mathbf{x}}_j\|_2^2)$ gives
\begin{equation}
\bigl|\langle\mathbf{A}\bar{\mathbf{x}}_i,\mathbf{A}\bar{\mathbf{x}}_j\rangle
-\langle\bar{\mathbf{x}}_i,\bar{\mathbf{x}}_j\rangle\bigr|
\le \tfrac{\varepsilon}{2}\bigl(
\|\bar{\mathbf{x}}_i\|_2^2+\|\bar{\mathbf{x}}_j\|_2^2
+\|\bar{\mathbf{x}}_i-\bar{\mathbf{x}}_j\|_2^2\bigr)
\le \tfrac{3\varepsilon}{2}
\bigl(\|\bar{\mathbf{x}}_i\|_2^2+\|\bar{\mathbf{x}}_j\|_2^2\bigr).
\end{equation}
Multiplying by $c$ and using $\|\bar{\mathbf{x}}_i\|_2^2=s_i^2/c$,
\begin{equation}
|\Delta^{\mathrm{ang}}_{ij}|
:=
c\bigl|\langle\mathbf{A}\bar{\mathbf{x}}_i,\mathbf{A}\bar{\mathbf{x}}_j\rangle
-\langle\bar{\mathbf{x}}_i,\bar{\mathbf{x}}_j\rangle\bigr|
\le \tfrac{3\varepsilon}{2}
\bigl(\sinh^2(\sqrt{c}\,r_i)+\sinh^2(\sqrt{c}\,r_j)\bigr).
\end{equation}
Subtracting the unperturbed identity
$c_i c_j-s_i s_j\cos\theta_{ij}=\cosh(\sqrt{c}\,d_c(x_i,x_j))$, the
total perturbation
$\Delta_{ij}=(c\,\widehat{T}_i\widehat{T}_j-c_i c_j)
-c\bigl(\langle\mathbf{A}\bar{\mathbf{x}}_i,\mathbf{A}\bar{\mathbf{x}}_j\rangle
-\langle\bar{\mathbf{x}}_i,\bar{\mathbf{x}}_j\rangle\bigr)$
of the $\operatorname{arcosh}$ argument satisfies (non-asymptotically)
\begin{equation}
|\Delta_{ij}|\le|\Delta^{\mathrm{rad}}_{ij}|+|\Delta^{\mathrm{ang}}_{ij}|
\le 3\varepsilon
\bigl(\sinh^2(\sqrt{c}\,r_i)+\sinh^2(\sqrt{c}\,r_j)\bigr)
\le 6\varepsilon\sinh^2(\sqrt{c}\,R).
\end{equation}
The hypothesis ensures
$|\Delta_{ij}|\le\tfrac{1}{2}(\cosh(\sqrt{c}\,\tau)-1)
\le\tfrac{1}{2}(\cosh(\sqrt{c}\,d_c(x_i,x_j))-1)$, so the mean-value bound
for $\operatorname{arcosh}$ along the segment between
$-c\langle x_i,x_j\rangle_L$ and
$-c\langle\Pi_{\mathbf{A}}x_i,\Pi_{\mathbf{A}}x_j\rangle_L$ (the same
bound used in the proof of Lemma~\ref{lem:rap-distance-transfer}) yields
$|d_c(\Pi_{\mathbf{A}}x_i,\Pi_{\mathbf{A}}x_j)-d_c(x_i,x_j)|
\le 2|\Delta_{ij}|/[\sqrt{c}\sinh(\sqrt{c}\,\tau)]$. Dividing by
$d_c(x_i,x_j)\ge\tau$ gives Equation~\eqref{eq:lorentz-envelope}.
\end{proof}

\appsection{RAP Radial Preservation and Angular Sensitivity}
\label{app:rap-advantage}

We prove Proposition~\ref{prop:rap-advantage} and the finite-set RAP
guarantee. The argument has three steps: well-definedness of the projected
point on the target hyperboloid, normalized JL preservation of angular inner
products, and transfer of the angular error into hyperbolic distance error
through the law of cosines.

\appsubsection{Well-Definedness and the Hyperboloid Constraint}

We work in the hyperboloid model
\begin{equation}
\mathbb{H}^k_c
=
\left\{
z=(z_0,\bar{\mathbf{z}})\in\mathbb{R}^{k+1}:
-z_0^2+\|\bar{\mathbf{z}}\|_2^2=-\frac{1}{c},\ z_0>0
\right\},
\end{equation}
with curvature $-c$. The origin is
$\tilde{o}=(1/\sqrt{c},\mathbf{0})$.

\begin{lemma}[Polar coordinates give valid hyperboloid points]
\label{lem:rap-hyperboloid}
For any $r\ge 0$ and any $\mathbf{v}\in\mathbb{S}^{k-1}$, define
\begin{equation}
\iota_k(r,\mathbf{v})
=
\frac{1}{\sqrt{c}}\bigl(\cosh(\sqrt{c}\,r),
\sinh(\sqrt{c}\,r)\mathbf{v}\bigr).
\end{equation}
Then $\iota_k(r,\mathbf{v})\in\mathbb{H}^k_c$ and
$d_c(\tilde{o},\iota_k(r,\mathbf{v}))=r$.
\end{lemma}

\begin{proof}
Let $z=\iota_k(r,\mathbf{v})$. Since $\|\mathbf{v}\|_2=1$,
\begin{equation}
-z_0^2+\|\bar{\mathbf{z}}\|_2^2
=
-\frac{1}{c}\cosh^2(\sqrt{c}\,r)
+\frac{1}{c}\sinh^2(\sqrt{c}\,r)
=-\frac{1}{c},
\end{equation}
and $z_0>0$. Thus $z\in\mathbb{H}^k_c$. Moreover,
$-c\langle \tilde{o},z\rangle_L=\cosh(\sqrt{c}\,r)$, so
\begin{equation}
d_c(\tilde{o},z)
=
\frac{1}{\sqrt{c}}\operatorname{arcosh}
\bigl(\cosh(\sqrt{c}\,r)\bigr)
=r.
\end{equation}
\end{proof}

\begin{corollary}[RAP is a valid hyperbolic random map on finite sets]
\label{cor:rap-valid-map}
Let $X=\{x_1,\ldots,x_n\}\subset\mathbb{H}^d_c$ be fixed, and write
$x_i=(r_i,\mathbf{u}_i)$ around an origin $o$, with an arbitrary direction
chosen when $r_i=0$. If $\|\mathbf{P}\mathbf{u}_i\|_2>0$ for every
non-anchor point, then RAP maps every $x_i$ into $\mathbb{H}^k_c$ and preserves
the anchor distance $d_c(o,x_i)=r_i$ exactly. If $\mathbf{P}$ is Gaussian with
entries $\mathcal{N}(0,1/k)$, this nonzero condition holds for the fixed finite
set $X$ with probability one.
\end{corollary}

\begin{proof}
For each non-anchor point, RAP sets
$\widetilde{\mathbf{u}}_i=\mathbf{P}\mathbf{u}_i/
\|\mathbf{P}\mathbf{u}_i\|_2$, hence
$\widetilde{\mathbf{u}}_i\in\mathbb{S}^{k-1}$. Lemma~\ref{lem:rap-hyperboloid}
then implies that
$\Pi_{\mathrm{RAP}}(x_i)=\iota_k(r_i,\widetilde{\mathbf{u}}_i)$ lies in
$\mathbb{H}^k_c$ and has radius $r_i$. If $r_i=0$, RAP maps $x_i=o$ to
$\tilde{o}$, so the conclusion is immediate.

For a fixed nonzero $\mathbf{u}_i$, the random vector
$\mathbf{P}\mathbf{u}_i$ is a nondegenerate Gaussian vector in
$\mathbb{R}^k$, so
$\Pr[\mathbf{P}\mathbf{u}_i=\mathbf{0}]=0$. A union bound over the finite set
of non-anchor points gives the probability-one statement.
\end{proof}

\appsubsection{Normalized Angular JL}

RAP does not use the projected vector $\mathbf{P}\mathbf{u}$ directly; it
renormalizes it back to the target sphere. The next lemma records that the
usual JL norm-preservation event is sufficient to preserve angular inner
products after this normalization.

\begin{lemma}[Normalized JL preserves angular inner products]
\label{lem:normalized-angular-jl}
Let $\mathbf{u}_1,\ldots,\mathbf{u}_n\in\mathbb{S}^{d-1}$ and let
$0<\varepsilon<1$. Suppose a matrix $\mathbf{P}\in\mathbb{R}^{k\times d}$
satisfies, for all $i,j$,
\begin{equation}
\left|\|\mathbf{P}\mathbf{u}_i\|_2^2-1\right|\le \varepsilon,
\qquad
\left|
\|\mathbf{P}(\mathbf{u}_i-\mathbf{u}_j)\|_2^2
-\|\mathbf{u}_i-\mathbf{u}_j\|_2^2
\right|
\le
\varepsilon\|\mathbf{u}_i-\mathbf{u}_j\|_2^2.
\end{equation}
Define
$\widetilde{\mathbf{u}}_i=\mathbf{P}\mathbf{u}_i/
\|\mathbf{P}\mathbf{u}_i\|_2$. Then, for all $i,j$,
\begin{equation}
\left|
\langle \widetilde{\mathbf{u}}_i,\widetilde{\mathbf{u}}_j\rangle
-\langle \mathbf{u}_i,\mathbf{u}_j\rangle
\right|
\le 4\varepsilon.
\end{equation}
Moreover, if $\mathbf{P}$ is a Gaussian JL matrix, then the displayed
hypotheses hold with probability at least $1-\delta$ whenever
$k\ge C\varepsilon^{-2}\log(n/\delta)$ for a universal constant $C$.
\end{lemma}

\begin{proof}
Write
$a_i=\|\mathbf{P}\mathbf{u}_i\|_2^2$,
$a_j=\|\mathbf{P}\mathbf{u}_j\|_2^2$,
$D_{ij}=\|\mathbf{u}_i-\mathbf{u}_j\|_2^2$, and
$\widehat{D}_{ij}=\|\mathbf{P}(\mathbf{u}_i-\mathbf{u}_j)\|_2^2$. By
polarization,
\begin{equation}
\langle \mathbf{P}\mathbf{u}_i,\mathbf{P}\mathbf{u}_j\rangle
=
\frac{a_i+a_j-\widehat{D}_{ij}}{2},
\qquad
\langle \mathbf{u}_i,\mathbf{u}_j\rangle
=
\frac{2-D_{ij}}{2}.
\end{equation}
Since $D_{ij}\le 4$, the assumptions imply
\begin{equation}
\left|
\langle \mathbf{P}\mathbf{u}_i,\mathbf{P}\mathbf{u}_j\rangle
-\langle \mathbf{u}_i,\mathbf{u}_j\rangle
\right|
\le
\frac{\varepsilon+\varepsilon+\varepsilon D_{ij}}{2}
\le 3\varepsilon.
\end{equation}
Also $a_i,a_j\in[1-\varepsilon,1+\varepsilon]$, so
$\sqrt{a_i a_j}\in[1-\varepsilon,1+\varepsilon]$. Because
$|\langle \mathbf{P}\mathbf{u}_i,\mathbf{P}\mathbf{u}_j\rangle|
\le \sqrt{a_i a_j}$,
\begin{equation}
\left|
\frac{\langle \mathbf{P}\mathbf{u}_i,\mathbf{P}\mathbf{u}_j\rangle}
{\sqrt{a_i a_j}}
-\langle \mathbf{P}\mathbf{u}_i,\mathbf{P}\mathbf{u}_j\rangle
\right|
\le
|1-\sqrt{a_i a_j}|
\le \varepsilon.
\end{equation}
Combining the two estimates gives the $4\varepsilon$ bound.

For the probability statement, apply the standard JL norm guarantee to the
finite collection
\begin{equation}
\mathcal{V}
=
\{\mathbf{u}_i:1\le i\le n\}
\cup
\{\mathbf{u}_i-\mathbf{u}_j:1\le i<j\le n,\ \mathbf{u}_i\ne\mathbf{u}_j\}.
\end{equation}
This collection has at most $n+n^2$ vectors, so
$k\ge C\varepsilon^{-2}\log(n/\delta)$ suffices after a union bound.
\end{proof}

\appsubsection{From Angular Error to Hyperbolic Distance Error}

For polar points $x=(r_x,\mathbf{u}_x)$ and $y=(r_y,\mathbf{u}_y)$, define
\begin{equation}
Z(x,y)
=
\cosh(\sqrt{c}\,r_x)\cosh(\sqrt{c}\,r_y)
-\sinh(\sqrt{c}\,r_x)\sinh(\sqrt{c}\,r_y)
\langle \mathbf{u}_x,\mathbf{u}_y\rangle.
\end{equation}
Then $Z(x,y)=\cosh(\sqrt{c}\,d_c(x,y))$.

\begin{lemma}[Angular error transfers through the hyperbolic law of cosines]
\label{lem:rap-distance-transfer}
Let $x,y\in\mathbb{H}^d_c$ have distance $d_{xy}=d_c(x,y)>0$, and suppose RAP
is defined on both points. If
\begin{equation}
\left|
\langle \widetilde{\mathbf{u}}_x,\widetilde{\mathbf{u}}_y\rangle
-\langle \mathbf{u}_x,\mathbf{u}_y\rangle
\right|
\le \eta,
\end{equation}
then
\begin{equation}
\left|
Z(\Pi_{\mathrm{RAP}}x,\Pi_{\mathrm{RAP}}y)-Z(x,y)
\right|
\le
\eta\,\sinh(\sqrt{c}\,r_x)\sinh(\sqrt{c}\,r_y).
\end{equation}
If the right-hand side is at most
$\frac{1}{2}\bigl(\cosh(\sqrt{c}\,d_{xy})-1\bigr)$, then
\begin{equation}
\left|
d_c(\Pi_{\mathrm{RAP}}x,\Pi_{\mathrm{RAP}}y)-d_{xy}
\right|
\le
\frac{
2\eta\,\sinh(\sqrt{c}\,r_x)\sinh(\sqrt{c}\,r_y)
}{
\sqrt{c}\,\sinh(\sqrt{c}\,d_{xy})
}.
\end{equation}
Consequently,
\begin{equation}
\left|
\frac{d_c(\Pi_{\mathrm{RAP}}x,\Pi_{\mathrm{RAP}}y)}{d_c(x,y)}-1
\right|
\le
\frac{
2\eta\,\sinh(\sqrt{c}\,r_x)\sinh(\sqrt{c}\,r_y)
}{
\sqrt{c}\,d_{xy}\sinh(\sqrt{c}\,d_{xy})
}.
\end{equation}
\end{lemma}

\begin{proof}
RAP leaves $r_x$ and $r_y$ unchanged, so the only changed term in
$Z(x,y)$ is the angular inner product. This gives the first bound.

Let $Z=Z(x,y)$ and $\widetilde{Z}=Z(\Pi_{\mathrm{RAP}}x,\Pi_{\mathrm{RAP}}y)$.
By the mean value theorem,
\begin{equation}
\left|
\operatorname{arcosh}(\widetilde{Z})-\operatorname{arcosh}(Z)
\right|
\le
|\widetilde{Z}-Z|
\sup_{t\text{ between }Z\text{ and }\widetilde{Z}}
\frac{1}{\sqrt{t-1}\sqrt{t+1}}.
\end{equation}
The assumed small-perturbation condition gives
$t-1\ge (Z-1)/2$ and $t+1\ge (Z+1)/2$ along the segment. Hence the derivative
is at most
$2/\sqrt{(Z-1)(Z+1)}=2/\sinh(\sqrt{c}\,d_{xy})$. Multiplying by
$1/\sqrt{c}$ converts arcosh error to hyperbolic distance error and proves the
displayed inequalities.
\end{proof}

\begin{proposition}[Pairs on the same radial ray are exact]
\label{prop:rap-same-ray}
If RAP is defined on $x=(r_x,\mathbf{u})$ and $y=(r_y,\mathbf{u})$, then
\begin{equation}
d_c(\Pi_{\mathrm{RAP}}x,\Pi_{\mathrm{RAP}}y)=d_c(x,y)=|r_x-r_y|.
\end{equation}
More generally, if
$\langle \widetilde{\mathbf{u}}_x,\widetilde{\mathbf{u}}_y\rangle
=\langle \mathbf{u}_x,\mathbf{u}_y\rangle$, then RAP preserves the distance
between $x$ and $y$ exactly.
\end{proposition}

\begin{proof}
If $\mathbf{u}_x=\mathbf{u}_y=\mathbf{u}$ and
$\mathbf{P}\mathbf{u}\ne \mathbf{0}$, then the normalized projected directions
are equal:
$\widetilde{\mathbf{u}}_x=\widetilde{\mathbf{u}}_y
=\mathbf{P}\mathbf{u}/\|\mathbf{P}\mathbf{u}\|_2$. Therefore the angular inner
product remains $1$. The hyperbolic law of cosines gives
\begin{equation}
\cosh(\sqrt{c}\,d_c(x,y))
=
\cosh(\sqrt{c}\,r_x)\cosh(\sqrt{c}\,r_y)
-\sinh(\sqrt{c}\,r_x)\sinh(\sqrt{c}\,r_y)
=
\cosh(\sqrt{c}\,|r_x-r_y|),
\end{equation}
and the same identity holds after RAP. The more general statement follows
because the law-of-cosines expression is unchanged whenever the angular inner
product is unchanged.
\end{proof}

\appsubsection{A Finite-Set RAP Guarantee}

The following theorem is not a universal hyperbolic JL theorem. Instead, it is
the finite-set guarantee naturally suggested by the RAP mechanism: if radii are
bounded, pairs are not arbitrarily close, and the angular directions satisfy a
JL event, then RAP preserves all distances with an explicit distortion bound.

\begin{theorem}[Finite-set RAP guarantee under bounded radius and separation]
\label{thm:rap-finite-set}
Let $X=\{x_1,\ldots,x_n\}\subset\mathbb{H}^d_c$ have polar coordinates
$x_i=(r_i,\mathbf{u}_i)$ around $o$. Assume
$r_i\le R$ for all $i$ and
$d_c(x_i,x_j)\ge \tau>0$ for all $i\ne j$. Let $\mathbf{P}$ be a Gaussian JL
matrix and let $0<\varepsilon<1$. If
\begin{equation}
4\varepsilon\sinh^2(\sqrt{c}\,R)
\le
\frac{1}{2}\bigl(\cosh(\sqrt{c}\,\tau)-1\bigr),
\end{equation}
and
$k\ge C\varepsilon^{-2}\log(n/\delta)$, then with probability at least
$1-\delta$, RAP is defined on $X$, maps $X$ into $\mathbb{H}^k_c$, preserves
all anchor radii exactly, and satisfies
\begin{equation}
\max_{i<j}
\left|
\frac{d_c(\Pi_{\mathrm{RAP}}x_i,\Pi_{\mathrm{RAP}}x_j)}
{d_c(x_i,x_j)}
-1
\right|
\le
\frac{
8\varepsilon\sinh^2(\sqrt{c}\,R)
}{
\sqrt{c}\,\tau\sinh(\sqrt{c}\,\tau)
}.
\end{equation}
More precisely, each pair obeys the sharper bound
\begin{equation}
\left|
\frac{d_c(\Pi_{\mathrm{RAP}}x_i,\Pi_{\mathrm{RAP}}x_j)}
{d_c(x_i,x_j)}
-1
\right|
\le
\frac{
8\varepsilon
\sinh(\sqrt{c}\,r_i)\sinh(\sqrt{c}\,r_j)
}{
\sqrt{c}\,d_c(x_i,x_j)\sinh(\sqrt{c}\,d_c(x_i,x_j))
}.
\end{equation}
\end{theorem}

\begin{proof}
By Lemma~\ref{lem:normalized-angular-jl}, with probability at least
$1-\delta$,
\begin{equation}
\left|
\langle \widetilde{\mathbf{u}}_i,\widetilde{\mathbf{u}}_j\rangle
-\langle \mathbf{u}_i,\mathbf{u}_j\rangle
\right|
\le 4\varepsilon
\end{equation}
for all pairs. The same event implies
$\|\mathbf{P}\mathbf{u}_i\|_2^2\ge 1-\varepsilon>0$, so RAP is defined on all
non-anchor points; Corollary~\ref{cor:rap-valid-map} gives the hyperboloid
constraint and exact anchor-radius preservation.

For a fixed pair, Lemma~\ref{lem:rap-distance-transfer} applies with
$\eta=4\varepsilon$. Since $r_i,r_j\le R$ and
$d_c(x_i,x_j)\ge\tau$, the theorem's small-perturbation condition ensures
\begin{equation}
4\varepsilon
\sinh(\sqrt{c}\,r_i)\sinh(\sqrt{c}\,r_j)
\le
4\varepsilon\sinh^2(\sqrt{c}\,R)
\le
\frac{1}{2}\bigl(\cosh(\sqrt{c}\,d_c(x_i,x_j))-1\bigr).
\end{equation}
Therefore Lemma~\ref{lem:rap-distance-transfer} yields the sharper pairwise
bound. Replacing $r_i,r_j$ by $R$ and $d_c(x_i,x_j)$ by $\tau$ gives the
uniform maximum bound.
\end{proof}

\begin{proof}[Proof of Proposition~\ref{prop:rap-advantage}]
The statement that RAP lands in $\mathbb{H}^k_c$ and preserves
$d_c(o,x)=r_x$ follows from Corollary~\ref{cor:rap-valid-map}. The exactness
for pairs with the same angular direction follows from
Proposition~\ref{prop:rap-same-ray}. The law-of-cosines perturbation bound is
the first conclusion of Lemma~\ref{lem:rap-distance-transfer}.
\end{proof}

\clearpage
\appsection{Theoretical Comparison of the Three Schemes}
\label{app:theoretical-comparison}

This appendix collects the multiplicative distortion envelopes proved for
the three constructions and arranges them under a common comparison. The
underlying statements are Theorem~\ref{prop:local-jl} for the tangent
scheme, Corollary~\ref{cor:lorentz-envelope}
(Appendix~\ref{app:lorentz-conditioning}) for the fully hyperbolic scheme, and
Theorem~\ref{thm:rap-finite-set} (Appendix~\ref{app:rap-advantage}) for
RAP. All three use the same JL parameter $\varepsilon$ and target
dimension $k\ge C\varepsilon^{-2}\log(n/\delta)$.

\begin{table}[!ht]
\centering
\caption{Multiplicative distortion envelope of the three projection schemes
under their respective regularity conditions: tangent uses anchor
$a\in\mathbb{H}^d_c$ with radius bound $R\le R_0/\sqrt{c}$; the fully
hyperbolic scheme and RAP use bounded radius $r_i\le R$ and pairwise separation
$d_c(x_i,x_j)\ge\tau$.}
\label{tab:bounds}
\small
\begin{tabularx}{\textwidth}{@{}>{\raggedright\arraybackslash}p{2.6cm}>{\raggedright\arraybackslash}X>{\raggedright\arraybackslash}X@{}}
\toprule
Scheme & Distortion envelope & Error decomposition \\
\midrule
Tangent\newline (Theorem~\ref{prop:local-jl}) &
$\varepsilon + C_1\,cR^2$ &
JL term $\varepsilon$ shrinks with $k$; curvature term $cR^2$ persists
and forces the local ball $R\le R_0/\sqrt{c}$. \\
\addlinespace[2pt]
Fully hyperbolic\newline (Corollary~\ref{cor:lorentz-envelope}) &
$\dfrac{12\,\varepsilon\,\sinh^2(\sqrt{c}\,R)}{\sqrt{c}\,\tau\,\sinh(\sqrt{c}\,\tau)}$ &
A radial contribution from the renormalization
(Lemma~\ref{lem:lorentz-mixing}) and a comparable angular contribution from
the spatial inner-product sketch; both pass through the
$\operatorname{arcosh}$ conditioning factor at small distances. \\
\addlinespace[2pt]
RAP\newline (Theorem~\ref{thm:rap-finite-set}) &
$\dfrac{8\,\varepsilon\,\sinh(\sqrt{c}\,r_x)\sinh(\sqrt{c}\,r_y)}
{\sqrt{c}\,d_c(x,y)\sinh(\sqrt{c}\,d_c(x,y))}$ &
A single angular contribution; radii are preserved exactly and the
radial source of error is removed by construction. \\
\bottomrule
\end{tabularx}
\end{table}

The envelopes for the fully hyperbolic scheme and RAP share the same denominator: both
constructions read distance through $\operatorname{arcosh}$ and pay the
conditioning factor $1/[\sqrt{c}\,d\sinh(\sqrt{c}\,d)]$ for nearby pairs.
The numerators differ in structure. The fully hyperbolic numerator carries two
contributions of comparable order, one from the renormalized time
coordinate and one from the spatial inner-product sketch, and both are
bounded only by the worst-case factor $\sinh^2(\sqrt{c}\,R)$. RAP
eliminates the radial source by construction: the radius is preserved
exactly, only the angular inner product is perturbed, and the per-pair
numerator collapses to $\sinh(\sqrt{c}\,r_x)\sinh(\sqrt{c}\,r_y)$ with no
cross-term. For a pair with one near-anchor endpoint, $\sinh(\sqrt{c}\,r)$
is small at that endpoint and the RAP numerator is much smaller than
$\sinh^2(\sqrt{c}\,R)$, while the fully hyperbolic numerator is not.

The tangent envelope is multiplicatively cleanest inside its small ball
but contains a non-vanishing $cR^2$ term, so it does not extend to large
radii without an additional global argument. The envelopes for the fully
hyperbolic scheme and RAP are stated for the finite-set regime with bounded radius and
pairwise separation, mirroring the finite-set form of classical Euclidean
JL. The empirical hierarchy in Section~\ref{sec:experiments} tracks this
theoretical ordering.

\clearpage
\appsection{Extended Related Work}
\label{app:extended-related-work}

\paragraph{Compression-based graph pattern mining.}
The CSPM line of work mines representative, star-shaped attribute patterns
in attributed graphs using conditional entropy and minimum description
length \citep{liu2022cspm,liu2022attributestars}. Here compression selects
descriptive patterns and codes for graph attributes, whereas RAP reduces
the dimension of continuous hyperbolic embeddings with a pairwise-distance
guarantee. These approaches share an interest in compact representations
but optimize different objects and criteria.

\paragraph{Graph learning and recommendation.}
Hyperbolic graph learning encompasses a broad range of architectures,
objectives, and applications \citep{yang2025graphreview}. Beyond the
representation methods discussed in the main text, hyperbolic geometry has
been used for graph contrastive learning and graph-level anomaly detection
\citep{liu2022contrastive,fu2026hcglad}, as well as collaborative filtering
\citep{yang2022hicf,yang2022hrcf}. These methods learn task-specific
representations and objectives that exploit hierarchical structure. They
provide application settings in which compact hyperbolic embeddings can be
useful; the compression question studied here concerns the geometry of an
already learned representation.

\paragraph{Personalized graph federated learning.}
FlatLand embeds different clients' graphs in tailored Lorentz spaces and
decouples client-specific geometric information from shared parameters
for federated aggregation \citep{liu2026flatland}. Its focus is learning
under heterogeneous graph structure, including low-dimensional settings.
RAP addresses a complementary stage: projecting a given finite set in a
fixed-curvature hyperbolic space while preserving radial coordinates and
controlling pairwise distance distortion. Extending this analysis across
client-specific curvatures would require additional assumptions.

\paragraph{Foundation-model systems and personalization.}
Tutorials on structured data, knowledge, and memory
\citep{liu2026tutorial} and modular hyperbolic libraries
\citep{he2025hypercore} organize the broader foundation-model ecosystem;
hyperbolic indexes also support retrieval-augmented generation
\citep{madhu2026hyprag}. Personalized language-model research spans
prompting, model adaptation, and preference alignment
\citep{liu2025personalizedsurvey}, including representation-space
personalization through learned low-rank interventions in PerFit
\citep{liu2026perfit}. Riemannian language modeling introduces learned
token-adaptive metrics into the output head \citep{xu2026riemannian}.
These developments motivate efficient geometric representations, while
the distance guarantees in this paper apply specifically to the
hyperbolic metric and the RAP map defined in Section~\ref{sec:rap}.

\begin{thebibliography}{80}
\providecommand{\natexlab}[1]{#1}
\providecommand{\url}[1]{\texttt{#1}}
\expandafter\ifx\csname urlstyle\endcsname\relax
  \providecommand{\doi}[1]{doi: #1}\else
  \providecommand{\doi}{doi: \begingroup \urlstyle{rm}\Url}\fi

\bibitem[Achlioptas(2003)]{achlioptas2003database}
Dimitris Achlioptas.
\newblock Database-friendly random projections: {Johnson--Lindenstrauss} with
  binary coins.
\newblock \emph{Journal of Computer and System Sciences}, 66\penalty0
  (4):\penalty0 671--687, 2003.

\bibitem[Ailon \& Chazelle(2009)Ailon and Chazelle]{ailon2009fast}
Nir Ailon and Bernard Chazelle.
\newblock The fast {Johnson--Lindenstrauss} transform and approximate nearest
  neighbors.
\newblock \emph{SIAM Journal on Computing}, 39\penalty0 (1):\penalty0 302--322,
  2009.

\bibitem[Alyoussef et~al.(2026)Alyoussef, Bdeir, Coello~de Portugal~Mecke,
  Hanika, Landwehr, and Schmidt-Thieme]{alyoussef2026hexformer}
Haya Alyoussef, Ahmad Bdeir, Diego Coello~de Portugal~Mecke, Tom Hanika, Niels
  Landwehr, and Lars Schmidt-Thieme.
\newblock {HexFormer}: Hyperbolic vision transformer with exponential map
  aggregation.
\newblock \emph{arXiv preprint arXiv:2601.19849}, 2026.

\bibitem[Baraniuk \& Wakin(2009)Baraniuk and Wakin]{baraniuk2009random}
Richard~G. Baraniuk and Michael~B. Wakin.
\newblock Random projections of smooth manifolds.
\newblock \emph{Foundations of Computational Mathematics}, 9\penalty0
  (1):\penalty0 51--77, 2009.
\newblock \doi{10.1007/s10208-007-9011-z}.

\bibitem[Bdeir et~al.(2024)Bdeir, Schwethelm, and Landwehr]{bdeir2023fully}
Ahmad Bdeir, Kristian Schwethelm, and Niels Landwehr.
\newblock Fully hyperbolic convolutional neural networks for computer vision.
\newblock In \emph{International Conference on Learning Representations}, 2024.
\newblock URL \url{https://openreview.net/forum?id=ekz1hN5QNh}.

\bibitem[Benjamini \& Makarychev(2009)Benjamini and
  Makarychev]{benjamini2009dimension}
Itai Benjamini and Yury Makarychev.
\newblock Dimension reduction for hyperbolic space.
\newblock \emph{Proceedings of the American Mathematical Society}, 137\penalty0
  (2):\penalty0 695--698, 2009.
\newblock \doi{10.1090/s0002-9939-08-09714-1}.

\bibitem[Bingham \& Mannila(2001)Bingham and Mannila]{bingham2001random}
Ella Bingham and Heikki Mannila.
\newblock Random projection in dimensionality reduction: Applications to image
  and text data.
\newblock In \emph{Proceedings of the Seventh ACM SIGKDD International
  Conference on Knowledge Discovery and Data Mining}, pp.\  245--250, 2001.

\bibitem[Bossard et~al.(2014)Bossard, Guillaumin, and
  Van~Gool]{bossard2014food}
Lukas Bossard, Matthieu Guillaumin, and Luc Van~Gool.
\newblock {Food-101} -- mining discriminative components with random forests.
\newblock In \emph{European Conference on Computer Vision}, 2014.

\bibitem[Bourgain(1985)]{bourgain1985lipschitz}
Jean Bourgain.
\newblock On {Lipschitz} embedding of finite metric spaces in {Hilbert} space.
\newblock \emph{Israel Journal of Mathematics}, 52\penalty0 (1--2):\penalty0
  46--52, 1985.

\bibitem[Boutsidis et~al.(2010)Boutsidis, Zouzias, and
  Drineas]{boutsidis2010random}
Christos Boutsidis, Anastasios Zouzias, and Petros Drineas.
\newblock Random projections for {$k$}-means clustering.
\newblock In \emph{Advances in Neural Information Processing Systems},
  volume~23, 2010.

\bibitem[Cannon et~al.(1997)Cannon, Floyd, Kenyon, and
  Parry]{cannon1997hyperbolic}
James~W. Cannon, William~J. Floyd, Richard Kenyon, and Walter~R. Parry.
\newblock Hyperbolic geometry.
\newblock In Silvio Levy (ed.), \emph{Flavors of Geometry}, volume~31 of
  \emph{MSRI Publications}, pp.\  59--115. Cambridge University Press, 1997.

\bibitem[Chami et~al.(2019)Chami, Ying, R{\'e}, and
  Leskovec]{chami2019hyperbolic}
Ines Chami, Zhitao Ying, Christopher R{\'e}, and Jure Leskovec.
\newblock Hyperbolic graph convolutional neural networks.
\newblock In \emph{Advances in Neural Information Processing Systems},
  volume~32, pp.\  4869--4880, 2019.

\bibitem[Chami et~al.(2021)Chami, Gu, Nguyen, and R{\'e}]{chami2021horopca}
Ines Chami, Albert Gu, Dat~P. Nguyen, and Christopher R{\'e}.
\newblock {HoroPCA}: Hyperbolic dimensionality reduction via horospherical
  projections.
\newblock In \emph{Proceedings of the 38th International Conference on Machine
  Learning}, volume 139 of \emph{Proceedings of Machine Learning Research},
  pp.\  1419--1429. PMLR, 2021.

\bibitem[Chang et~al.(2025)Chang, Lin, Lin, Chen, Hu, Wang, Huang, Ceze,
  Abdelfattah, and Wu]{chang2025palu}
Chi-Chih Chang, Wei-Cheng Lin, Chien-Yu Lin, Chong-Yan Chen, Yu-Fang Hu,
  Pei-Shuo Wang, Ning-Chi Huang, Luis Ceze, Mohamed Abdelfattah, and Kai-Chiang
  Wu.
\newblock {Palu}: {KV-Cache} compression with low-rank projection.
\newblock In \emph{International Conference on Learning Representations}, 2025.

\bibitem[Chen et~al.(2022)Chen, Han, Lin, Zhao, Liu, Li, Sun, and
  Zhou]{chen2021fully}
Weize Chen, Xu~Han, Yankai Lin, Hexu Zhao, Zhiyuan Liu, Peng Li, Maosong Sun,
  and Jie Zhou.
\newblock Fully hyperbolic neural networks.
\newblock In \emph{Proceedings of the 60th Annual Meeting of the Association
  for Computational Linguistics (Volume 1: Long Papers)}, pp.\  5672--5686,
  Dublin, Ireland, 2022. Association for Computational Linguistics.
\newblock \doi{10.18653/v1/2022.acl-long.389}.
\newblock URL \url{https://aclanthology.org/2022.acl-long.389/}.

\bibitem[Chen et~al.(2026)Chen, Sch{\"o}lkopf, and Sebe]{chen2026busemann}
Ziheng Chen, Bernhard Sch{\"o}lkopf, and Nicu Sebe.
\newblock Hyperbolic busemann neural networks.
\newblock In \emph{Proceedings of the IEEE/CVF Conference on Computer Vision
  and Pattern Recognition}, pp.\  42029--42038, 2026.
\newblock URL
  \url{https://openaccess.thecvf.com/content/CVPR2026/papers/Chen_Hyperbolic_Busemann_Neural_Networks_CVPR_2026_paper.pdf}.

\bibitem[Choromanski et~al.(2021)Choromanski, Likhosherstov, Dohan, Song, Gane,
  Sarlos, Hawkins, Davis, Mohiuddin, Kaiser, Belanger, Colwell, and
  Weller]{choromanski2021rethinking}
Krzysztof Choromanski, Valerii Likhosherstov, David Dohan, Xingyou Song,
  Georgiana Gane, Tamas Sarlos, Peter Hawkins, Jared Davis, Afroz Mohiuddin,
  Lukasz Kaiser, David Belanger, Lucy Colwell, and Adrian Weller.
\newblock Rethinking attention with performers.
\newblock In \emph{International Conference on Learning Representations}, 2021.

\bibitem[Cimpoi et~al.(2014)Cimpoi, Maji, Kokkinos, Mohamed, and
  Vedaldi]{cimpoi2014describing}
Mircea Cimpoi, Subhransu Maji, Iasonas Kokkinos, Sammy Mohamed, and Andrea
  Vedaldi.
\newblock Describing textures in the wild.
\newblock In \emph{Proceedings of the IEEE Conference on Computer Vision and
  Pattern Recognition}, 2014.

\bibitem[Dasgupta \& Gupta(2003)Dasgupta and Gupta]{dasgupta2003elementary}
Sanjoy Dasgupta and Anupam Gupta.
\newblock An elementary proof of the {Johnson--Lindenstrauss} lemma.
\newblock \emph{Random Structures \& Algorithms}, 22\penalty0 (1):\penalty0
  60--65, 2003.

\bibitem[Deng et~al.(2026)Deng, Gao, Lu, Luo, and Xin]{deng2026locality}
Chengyuan Deng, Jie Gao, Kevin Lu, Feng Luo, and Cheng Xin.
\newblock {Locality Sensitive Hashing in Hyperbolic Space}.
\newblock In Hee-Kap Ahn, Michael Hoffmann, and Amir Nayyeri (eds.), \emph{42nd
  International Symposium on Computational Geometry (SoCG 2026)}, volume 367 of
  \emph{Leibniz International Proceedings in Informatics (LIPIcs)}, pp.\
  39:1--39:19, Dagstuhl, Germany, 2026. Schloss Dagstuhl -- Leibniz-Zentrum
  f{\"u}r Informatik.
\newblock \doi{10.4230/LIPIcs.SoCG.2026.39}.
\newblock URL
  \url{https://drops.dagstuhl.de/entities/document/10.4230/LIPIcs.SoCG.2026.39}.

\bibitem[Desai et~al.(2023)Desai, Nickel, Rajpurohit, Johnson, and
  Vedantam]{desai2023meru}
Karan Desai, Maximilian Nickel, Tanmay Rajpurohit, Justin Johnson, and
  Shanmukha~Ramakrishna Vedantam.
\newblock Hyperbolic image-text representations.
\newblock In \emph{Proceedings of the 40th International Conference on Machine
  Learning}, volume 202 of \emph{Proceedings of Machine Learning Research},
  pp.\  7694--7731. PMLR, 2023.
\newblock URL \url{https://proceedings.mlr.press/v202/desai23a.html}.

\bibitem[Fan et~al.(2022)Fan, Yang, and Vemuri]{fan2022nested}
Xiran Fan, Chun-Hao Yang, and Baba~C. Vemuri.
\newblock Nested hyperbolic spaces for dimensionality reduction and hyperbolic
  {NN} design.
\newblock In \emph{Proceedings of the IEEE/CVF Conference on Computer Vision
  and Pattern Recognition}, pp.\  356--365, 2022.

\bibitem[Fei-Fei et~al.(2004)Fei-Fei, Fergus, and Perona]{feifei2004learning}
Li~Fei-Fei, Rob Fergus, and Pietro Perona.
\newblock Learning generative visual models from few training examples.
\newblock In \emph{Workshop on Generative-Model Based Vision, IEEE Conference
  on Computer Vision and Pattern Recognition}, 2004.

\bibitem[Fletcher et~al.(2004)Fletcher, Lu, Pizer, and
  Joshi]{fletcher2004principal}
P.~Thomas Fletcher, Conglin Lu, Stephen~M. Pizer, and Sarang Joshi.
\newblock Principal geodesic analysis for the study of nonlinear statistics of
  shape.
\newblock \emph{IEEE Transactions on Medical Imaging}, 23\penalty0
  (8):\penalty0 995--1005, 2004.

\bibitem[Fu et~al.(2026)Fu, Li, Liu, Xing, Wang, and King]{fu2026hcglad}
Yali Fu, Jindong Li, Jiahong Liu, Qianli Xing, Qi~Wang, and Irwin King.
\newblock {HC-GLAD: Dual hyperbolic contrastive learning for unsupervised
  graph-level anomaly detection}.
\newblock \emph{Neural Networks}, 202:\penalty0 109009, 2026.
\newblock \doi{10.1016/j.neunet.2026.109009}.
\newblock URL \url{https://doi.org/10.1016/j.neunet.2026.109009}.

\bibitem[Ganea et~al.(2018)Ganea, B{\'e}cigneul, and
  Hofmann]{ganea2018hyperbolic}
Octavian-Eugen Ganea, Gary B{\'e}cigneul, and Thomas Hofmann.
\newblock Hyperbolic neural networks.
\newblock In \emph{Advances in Neural Information Processing Systems}, 2018.

\bibitem[Halko et~al.(2011)Halko, Martinsson, and Tropp]{halko2011finding}
Nathan Halko, Per-Gunnar Martinsson, and Joel~A. Tropp.
\newblock Finding structure with randomness: Probabilistic algorithms for
  constructing approximate matrix decompositions.
\newblock \emph{SIAM Review}, 53\penalty0 (2):\penalty0 217--288, 2011.

\bibitem[He et~al.(2025{\natexlab{a}})He, Anand, Madhu, Maatouk, Krishnaswamy,
  Tassiulas, Yang, and Ying]{he2025helm}
Neil He, Rishabh Anand, Hiren Madhu, Ali Maatouk, Smita Krishnaswamy, Leandros
  Tassiulas, Menglin Yang, and Rex Ying.
\newblock {HELM}: Hyperbolic large language models via mixture-of-curvature
  experts.
\newblock In \emph{Advances in Neural Information Processing Systems},
  volume~38, pp.\  142604--142635. Curran Associates, Inc., 2025{\natexlab{a}}.
\newblock \doi{10.52202/085713-4774}.
\newblock URL
  \url{https://proceedings.neurips.cc/paper_files/paper/2025/hash/d1e2f808a51842eedaf6ef0099d716c6-Abstract-Conference.html}.

\bibitem[He et~al.(2025{\natexlab{b}})He, Yang, and Ying]{he2025hypercore}
Neil He, Menglin Yang, and Rex Ying.
\newblock {HyperCore}: The core framework for building hyperbolic foundation
  models with comprehensive modules.
\newblock \emph{arXiv preprint arXiv:2504.08912}, 2025{\natexlab{b}}.

\bibitem[Helber et~al.(2019)Helber, Bischke, Dengel, and
  Borth]{helber2019eurosat}
Patrick Helber, Benjamin Bischke, Andreas Dengel, and Damian Borth.
\newblock {EuroSAT}: A novel dataset and deep learning benchmark for land use
  and land cover classification.
\newblock \emph{IEEE Journal of Selected Topics in Applied Earth Observations
  and Remote Sensing}, 2019.

\bibitem[Johnson et~al.(2017)Johnson, Hariharan, van~der Maaten, Fei-Fei,
  Zitnick, and Girshick]{johnson2017clevr}
Justin Johnson, Bharath Hariharan, Laurens van~der Maaten, Li~Fei-Fei,
  C.~Lawrence Zitnick, and Ross Girshick.
\newblock {CLEVR}: A diagnostic dataset for compositional language and
  elementary visual reasoning.
\newblock In \emph{Proceedings of the IEEE Conference on Computer Vision and
  Pattern Recognition}, 2017.

\bibitem[Johnson \& Lindenstrauss(1984)Johnson and
  Lindenstrauss]{johnson1984extensions}
William~B. Johnson and Joram Lindenstrauss.
\newblock Extensions of {Lipschitz} mappings into a {Hilbert} space.
\newblock In \emph{Contemporary Mathematics}, volume~26, pp.\  189--206.
  American Mathematical Society, 1984.

\bibitem[Khrulkov et~al.(2020)Khrulkov, Mirvakhabova, Ustinova, Oseledets, and
  Lempitsky]{khrulkov2020hyperbolic}
Valentin Khrulkov, Leyla Mirvakhabova, Evgeniya Ustinova, Ivan Oseledets, and
  Victor Lempitsky.
\newblock Hyperbolic image embeddings.
\newblock In \emph{Proceedings of the IEEE/CVF Conference on Computer Vision
  and Pattern Recognition}, 2020.

\bibitem[Krizhevsky(2009)]{krizhevsky2009learning}
Alex Krizhevsky.
\newblock Learning multiple layers of features from tiny images.
\newblock Technical report, University of Toronto, 2009.
\newblock URL
  \url{https://www.cs.toronto.edu/~kriz/learning-features-2009-TR.pdf}.

\bibitem[Larsen \& Nelson(2017)Larsen and Nelson]{larsen2017optimal}
Kasper~Green Larsen and Jelani Nelson.
\newblock Optimality of the {Johnson--Lindenstrauss} lemma.
\newblock In \emph{Proceedings of the 58th Annual IEEE Symposium on Foundations
  of Computer Science (FOCS)}, pp.\  633--638. IEEE, 2017.

\bibitem[Law et~al.(2019)Law, Liao, Snell, and Zemel]{law2019lorentzian}
Marc Law, Renjie Liao, Jake Snell, and Richard Zemel.
\newblock Lorentzian distance learning for hyperbolic representations.
\newblock In \emph{Proceedings of the 36th International Conference on Machine
  Learning}, volume~97 of \emph{Proceedings of Machine Learning Research}, pp.\
   3672--3681. PMLR, 2019.

\bibitem[LeCun et~al.(1998)LeCun, Bottou, Bengio, and
  Haffner]{lecun1998gradient}
Yann LeCun, L{\'e}on Bottou, Yoshua Bengio, and Patrick Haffner.
\newblock Gradient-based learning applied to document recognition.
\newblock \emph{Proceedings of the IEEE}, 86\penalty0 (11):\penalty0
  2278--2324, 1998.

\bibitem[Lee \& Verleysen(2009)Lee and Verleysen]{lee2009quality}
John~A. Lee and Michel Verleysen.
\newblock Quality assessment of dimensionality reduction: Rank-based criteria.
\newblock \emph{Neurocomputing}, 72\penalty0 (7--9):\penalty0 1431--1443, 2009.
\newblock \doi{10.1016/j.neucom.2008.12.017}.

\bibitem[Lin et~al.(2014)Lin, Maire, Belongie, Hays, Perona, Ramanan,
  Doll{\'a}r, and Zitnick]{lin2014microsoft}
Tsung-Yi Lin, Michael Maire, Serge Belongie, James Hays, Pietro Perona, Deva
  Ramanan, Piotr Doll{\'a}r, and C.~Lawrence Zitnick.
\newblock Microsoft {COCO}: Common objects in context.
\newblock In \emph{European Conference on Computer Vision}, 2014.

\bibitem[Liu et~al.(2022{\natexlab{a}})Liu, Fournier-Viger, Zhou, He, and
  Nouioua]{liu2022cspm}
Jiahong Liu, Philippe Fournier-Viger, Min Zhou, Ganghuan He, and Mourad
  Nouioua.
\newblock {CSPM: Discovering compressing stars in attributed graphs}.
\newblock \emph{Information Sciences}, 611:\penalty0 126--158,
  2022{\natexlab{a}}.
\newblock \doi{10.1016/j.ins.2022.08.008}.
\newblock URL \url{https://doi.org/10.1016/j.ins.2022.08.008}.

\bibitem[Liu et~al.(2022{\natexlab{b}})Liu, Yang, Zhou, Feng, and
  Fournier-Viger]{liu2022contrastive}
Jiahong Liu, Menglin Yang, Min Zhou, Shanshan Feng, and Philippe
  Fournier-Viger.
\newblock {Enhancing Hyperbolic Graph Embeddings via Contrastive Learning}.
\newblock \emph{arXiv preprint arXiv:2201.08554}, 2022{\natexlab{b}}.
\newblock URL \url{https://arxiv.org/abs/2201.08554}.

\bibitem[Liu et~al.(2022{\natexlab{c}})Liu, Zhou, Fournier-Viger, Yang, Pan,
  and Nouioua]{liu2022attributestars}
Jiahong Liu, Min Zhou, Philippe Fournier-Viger, Menglin Yang, Lujia Pan, and
  Mourad Nouioua.
\newblock {Discovering Representative Attribute-stars via Minimum Description
  Length}.
\newblock In \emph{2022 IEEE 38th International Conference on Data Engineering
  (ICDE)}, pp.\  68--80. IEEE, 2022{\natexlab{c}}.
\newblock \doi{10.1109/icde53745.2022.00010}.
\newblock URL \url{https://doi.org/10.1109/icde53745.2022.00010}.

\bibitem[Liu et~al.(2025)Liu, Qiu, Li, Dai, Yu, Zhu, Hu, Yang, Chua, and
  King]{liu2025personalizedsurvey}
Jiahong Liu, Zexuan Qiu, Zhongyang Li, Quanyu Dai, Wenhao Yu, Jieming Zhu,
  Minda Hu, Menglin Yang, Tat-Seng Chua, and Irwin King.
\newblock {A Survey of Personalized Large Language Models: Progress and Future
  Directions}.
\newblock \emph{arXiv preprint arXiv:2502.11528}, 2025.
\newblock URL \url{https://arxiv.org/abs/2502.11528}.

\bibitem[Liu et~al.(2026{\natexlab{a}})Liu, B, Fu, Yang, Zhang, Ying, and
  King]{liu2026flatland}
Jiahong Liu, Ram Samarth~B B, Xinyu Fu, Menglin Yang, Weixi Zhang, Zhitao Ying,
  and Irwin King.
\newblock {FlatLand}: Personalized graph federated learning via tailored
  {Lorentz} space.
\newblock In \emph{Proceedings of the 43rd International Conference on Machine
  Learning}, volume 306 of \emph{Proceedings of Machine Learning Research},
  pp.\  75360--75393. PMLR, 2026{\natexlab{a}}.
\newblock URL \url{https://proceedings.mlr.press/v306/liu26b.html}.

\bibitem[Liu et~al.(2026{\natexlab{b}})Liu, Shen, Liu, Ying, Yang, Chua, and
  King]{liu2026hmcl}
Jiahong Liu, Ming Shen, Xiaohao Liu, Rex Ying, Menglin Yang, Tat-Seng Chua, and
  Irwin King.
\newblock {Hyperbolic Multimodal Continual Learning: A Closest-Admissible
  Solution}.
\newblock \emph{arXiv preprint arXiv:2609.29329}, 2026{\natexlab{b}}.
\newblock URL \url{https://arxiv.org/abs/2609.29329}.

\bibitem[Liu et~al.(2026{\natexlab{c}})Liu, Yang, and King]{liu2026tutorial}
Jiahong Liu, Menglin Yang, and Irwin King.
\newblock {Hyperbolic Learning for Structured Data, Knowledge, and Memory: A
  Tutorial}.
\newblock In \emph{Proceedings of the 32nd ACM SIGKDD Conference on Knowledge
  Discovery and Data Mining V.2}, pp.\  13351--13355. ACM, 2026{\natexlab{c}}.
\newblock \doi{10.1145/3770855.3816467}.
\newblock URL \url{https://doi.org/10.1145/3770855.3816467}.

\bibitem[Liu et~al.(2026{\natexlab{d}})Liu, Yu, Dai, Li, Zhu, Yang, Chua, and
  King]{liu2026perfit}
Jiahong Liu, Wenhao Yu, Quanyu Dai, Zhongyang Li, Jieming Zhu, Menglin Yang,
  Tat-Seng Chua, and Irwin King.
\newblock {PerFit: Exploring Personalization Shifts in Representation Space of
  LLMs}.
\newblock In \emph{International Conference on Learning Representations},
  volume 2026, pp.\  75361--75396, 2026{\natexlab{d}}.
\newblock URL
  \url{https://proceedings.iclr.cc/paper_files/paper/2026/file/7a1ac10ea9d98178a389c7fbb3575567-Paper-Conference.pdf}.

\bibitem[Lou et~al.(2020)Lou, Katsman, Jiang, Belongie, Lim, and
  De~Sa]{lou2020differentiating}
Aaron Lou, Isay Katsman, Qingxuan Jiang, Serge Belongie, Ser-Nam Lim, and
  Christopher De~Sa.
\newblock Differentiating through the {Fr{\'e}chet} mean.
\newblock In \emph{International Conference on Machine Learning}, 2020.

\bibitem[Madhu et~al.(2026)Madhu, Bui, Maatouk, Tassiulas, Krishnaswamy, Yang,
  Ganguly, Srinivasan, and Ying]{madhu2026hyprag}
Hiren Madhu, Ngoc Bui, Ali Maatouk, Leandros Tassiulas, Smita Krishnaswamy,
  Menglin Yang, Sukanta Ganguly, Kiran Srinivasan, and Rex Ying.
\newblock {HypRAG}: Hyperbolic dense retrieval for retrieval augmented
  generation.
\newblock \emph{arXiv preprint arXiv:2602.07739}, 2026.

\bibitem[Maji et~al.(2013)Maji, Rahtu, Kannala, Blaschko, and
  Vedaldi]{maji2013fine}
Subhransu Maji, Esa Rahtu, Juho Kannala, Matthew Blaschko, and Andrea Vedaldi.
\newblock Fine-grained visual classification of aircraft, 2013.
\newblock URL \url{https://arxiv.org/abs/1306.5151}.

\bibitem[Matou{\v{s}}ek(2008)]{matousek2008variants}
Ji{\v{r}}{\'\i} Matou{\v{s}}ek.
\newblock On variants of the {Johnson--Lindenstrauss} lemma.
\newblock \emph{Random Structures \& Algorithms}, 33\penalty0 (2):\penalty0
  142--156, 2008.

\bibitem[Miller(1995)]{miller1995wordnet}
George~A. Miller.
\newblock {WordNet}: A lexical database for {English}.
\newblock \emph{Communications of the ACM}, 38\penalty0 (11):\penalty0 39--41,
  1995.
\newblock \doi{10.1145/219717.219748}.

\bibitem[Nagano et~al.(2019)Nagano, Yamaguchi, Fujita, and
  Koyama]{nagano2019wrapped}
Yoshihiro Nagano, Shoichiro Yamaguchi, Yasuhiro Fujita, and Masanori Koyama.
\newblock A wrapped normal distribution on hyperbolic space for gradient-based
  learning.
\newblock In \emph{Proceedings of the 36th International Conference on Machine
  Learning}, volume~97 of \emph{Proceedings of Machine Learning Research}, pp.\
   4693--4702. PMLR, 2019.

\bibitem[Nickel \& Kiela(2017)Nickel and Kiela]{nickel2017poincare}
Maximilian Nickel and Douwe Kiela.
\newblock Poincar{\'e} embeddings for learning hierarchical representations.
\newblock In \emph{Advances in Neural Information Processing Systems}, 2017.

\bibitem[Nickel \& Kiela(2018)Nickel and Kiela]{nickel2018learning}
Maximilian Nickel and Douwe Kiela.
\newblock Learning continuous hierarchies in the {L}orentz model of hyperbolic
  geometry.
\newblock In \emph{Proceedings of the 35th International Conference on Machine
  Learning}, 2018.

\bibitem[Nilsback \& Zisserman(2008)Nilsback and
  Zisserman]{nilsback2008automated}
Maria-Elena Nilsback and Andrew Zisserman.
\newblock Automated flower classification over a large number of classes.
\newblock In \emph{Indian Conference on Computer Vision, Graphics and Image
  Processing}, 2008.

\bibitem[Pal et~al.(2025)Pal, van Spengler, D'Amely~di Melendugno, Flaborea,
  Galasso, and Mettes]{pal2024compositional}
Avik Pal, Max van Spengler, Guido~Maria D'Amely~di Melendugno, Alessandro
  Flaborea, Fabio Galasso, and Pascal Mettes.
\newblock Compositional entailment learning for hyperbolic vision-language
  models.
\newblock In \emph{International Conference on Learning Representations}, 2025.

\bibitem[Paul et~al.(2013)Paul, Boutsidis, Magdon-Ismail, and
  Drineas]{paul2013random}
Saurabh Paul, Christos Boutsidis, Malik Magdon-Ismail, and Petros Drineas.
\newblock Random projections for support vector machines.
\newblock In \emph{Proceedings of the Sixteenth International Conference on
  Artificial Intelligence and Statistics}, volume~31 of \emph{Proceedings of
  Machine Learning Research}, pp.\  498--506. PMLR, 2013.

\bibitem[Peng et~al.(2025)Peng, Xu, Liu, Yang, and Shen]{peng2025hyperet}
Zelin Peng, Zhengqin Xu, Qingyang Liu, Xiaokang Yang, and Wei Shen.
\newblock {HyperET}: Efficient training in hyperbolic space for multi-modal
  large language models.
\newblock In \emph{Advances in Neural Information Processing Systems}, 2025.

\bibitem[Qiu et~al.(2024)Qiu, Liu, Chen, and King]{qiu2024hihpq}
Zexuan Qiu, Jiahong Liu, Yankai Chen, and Irwin King.
\newblock {HiHPQ: Hierarchical Hyperbolic Product Quantization for Unsupervised
  Image Retrieval}.
\newblock \emph{Proceedings of the AAAI Conference on Artificial Intelligence},
  38\penalty0 (5):\penalty0 4614--4622, 2024.
\newblock \doi{10.1609/aaai.v38i5.28261}.
\newblock URL \url{https://doi.org/10.1609/aaai.v38i5.28261}.

\bibitem[Radford et~al.(2021)Radford, Kim, Hallacy, Ramesh, Goh, Agarwal,
  Sastry, Askell, Mishkin, Clark, Krueger, and Sutskever]{radford2021learning}
Alec Radford, Jong~Wook Kim, Chris Hallacy, Aditya Ramesh, Gabriel Goh,
  Sandhini Agarwal, Girish Sastry, Amanda Askell, Pamela Mishkin, Jack Clark,
  Gretchen Krueger, and Ilya Sutskever.
\newblock Learning transferable visual models from natural language
  supervision.
\newblock In \emph{Proceedings of the 38th International Conference on Machine
  Learning}, volume 139 of \emph{Proceedings of Machine Learning Research},
  pp.\  8748--8763. PMLR, 2021.
\newblock URL \url{https://proceedings.mlr.press/v139/radford21a.html}.

\bibitem[Ratcliffe(2019)]{ratcliffe2019foundations}
John~G. Ratcliffe.
\newblock \emph{Foundations of Hyperbolic Manifolds}, volume 149 of
  \emph{Graduate Texts in Mathematics}.
\newblock Springer, 3 edition, 2019.

\bibitem[Roy \& Vetterli(2007)Roy and Vetterli]{roy2007effective}
Olivier Roy and Martin Vetterli.
\newblock The effective rank: A measure of effective dimensionality.
\newblock In \emph{Proceedings of the 15th European Signal Processing
  Conference}, pp.\  606--610, 2007.

\bibitem[Saha et~al.(2024)Saha, Sagan, Srivastava, Goldsmith, and
  Pilanci]{saha2024compressing}
Rajarshi Saha, Naomi Sagan, Varun Srivastava, Andrea~J. Goldsmith, and Mert
  Pilanci.
\newblock Compressing large language models using low rank and low precision
  decomposition.
\newblock In \emph{Advances in Neural Information Processing Systems}, 2024.

\bibitem[Sala et~al.(2018)Sala, De~Sa, Gu, and R{\'e}]{sala2018representation}
Frederic Sala, Christopher De~Sa, Albert Gu, and Christopher R{\'e}.
\newblock Representation tradeoffs for hyperbolic embeddings.
\newblock In \emph{Proceedings of the 35th International Conference on Machine
  Learning}, volume~80 of \emph{Proceedings of Machine Learning Research}, pp.\
   4460--4469. PMLR, 2018.

\bibitem[Shimizu et~al.(2021)Shimizu, Mukuta, and Harada]{shimizu2021hnnpp}
Ryohei Shimizu, Yusuke Mukuta, and Tatsuya Harada.
\newblock Hyperbolic neural networks++.
\newblock In \emph{International Conference on Learning Representations}, 2021.

\bibitem[Socher et~al.(2013)Socher, Perelygin, Wu, Chuang, Manning, Ng, and
  Potts]{socher2013recursive}
Richard Socher, Alex Perelygin, Jean Wu, Jason Chuang, Christopher~D. Manning,
  Andrew Ng, and Christopher Potts.
\newblock Recursive deep models for semantic compositionality over a sentiment
  treebank.
\newblock In \emph{Proceedings of the 2013 Conference on Empirical Methods in
  Natural Language Processing}, pp.\  1631--1642. Association for Computational
  Linguistics, 2013.
\newblock URL \url{https://aclanthology.org/D13-1170/}.

\bibitem[Tifrea et~al.(2019)Tifrea, B{\'e}cigneul, and
  Ganea]{tifrea2019poincare}
Alexandru Tifrea, Gary B{\'e}cigneul, and Octavian-Eugen Ganea.
\newblock {Poincar{\'e}} {GloVe}: Hyperbolic word embeddings.
\newblock In \emph{International Conference on Learning Representations}, 2019.

\bibitem[van~der Klis et~al.(2026)van~der Klis, Ch{\'a}vez~Torres, van
  Spengler, Ding, Hofmann, and Mettes]{vanderklis2026fast}
Robert van~der Klis, Ricardo Ch{\'a}vez~Torres, Max van Spengler, Yuhui Ding,
  Thomas Hofmann, and Pascal Mettes.
\newblock Fast and geometrically grounded lorentz neural networks.
\newblock \emph{arXiv preprint arXiv:2601.21529}, 2026.

\bibitem[Veeling et~al.(2018)Veeling, Linmans, Winkens, Cohen, and
  Welling]{veeling2018rotation}
Bastiaan~S. Veeling, Jasper Linmans, Jim Winkens, Taco Cohen, and Max Welling.
\newblock Rotation equivariant {CNNs} for digital pathology.
\newblock In \emph{Medical Image Computing and Computer Assisted Intervention
  -- MICCAI 2018}, volume 11071 of \emph{Lecture Notes in Computer Science},
  pp.\  210--218. Springer International Publishing, 2018.
\newblock \doi{10.1007/978-3-030-00934-2_24}.
\newblock URL \url{https://doi.org/10.1007/978-3-030-00934-2_24}.

\bibitem[Weinberger et~al.(2009)Weinberger, Dasgupta, Langford, Smola, and
  Attenberg]{weinberger2009feature}
Kilian~Q. Weinberger, Anirban Dasgupta, John Langford, Alexander~J. Smola, and
  Josh Attenberg.
\newblock Feature hashing for large scale multitask learning.
\newblock In \emph{International Conference on Machine Learning}, 2009.

\bibitem[Xu et~al.(2026)Xu, Liu, Zhuang, Li, Wang, Sun, and
  Yang]{xu2026riemannian}
Shu Xu, Jiahong Liu, Wenhao Zhuang, Jiabao Li, Yefan Wang, Ming Sun, and
  Menglin Yang.
\newblock {Riemannian Large Language Modeling}.
\newblock In \emph{Findings of the {A}ssociation for {C}omputational
  {L}inguistics: {EMNLP} 2026}, pp.\  44427--44440, Budapest, Hungary, 2026.
  Association for Computational Linguistics.
\newblock \doi{10.18653/2026.findings-emnlp.2087}.
\newblock URL \url{https://aclanthology.org/2026.findings-emnlp.2087/}.

\bibitem[Yang et~al.(2022{\natexlab{a}})Yang, Li, Zhou, Liu, and
  King]{yang2022hicf}
Menglin Yang, Zhihao Li, Min Zhou, Jiahong Liu, and Irwin King.
\newblock {HICF: Hyperbolic Informative Collaborative Filtering}.
\newblock In \emph{Proceedings of the 28th ACM SIGKDD Conference on Knowledge
  Discovery and Data Mining}, pp.\  2212--2221. ACM, 2022{\natexlab{a}}.
\newblock \doi{10.1145/3534678.3539475}.
\newblock URL \url{https://doi.org/10.1145/3534678.3539475}.

\bibitem[Yang et~al.(2022{\natexlab{b}})Yang, Zhou, Liu, Lian, and
  King]{yang2022hrcf}
Menglin Yang, Min Zhou, Jiahong Liu, Defu Lian, and Irwin King.
\newblock {HRCF: Enhancing Collaborative Filtering via Hyperbolic Geometric
  Regularization}.
\newblock In \emph{Proceedings of the ACM Web Conference 2022}, pp.\
  2462--2471. ACM, 2022{\natexlab{b}}.
\newblock \doi{10.1145/3485447.3512118}.
\newblock URL \url{https://doi.org/10.1145/3485447.3512118}.

\bibitem[Yang et~al.(2024)Yang, Verma, Zhang, Liu, King, and
  Ying]{yang2024hypformer}
Menglin Yang, Harshit Verma, Delvin~Ce Zhang, Jiahong Liu, Irwin King, and Rex
  Ying.
\newblock Hypformer: Exploring efficient transformer fully in hyperbolic space.
\newblock In \emph{Proceedings of the 30th ACM SIGKDD Conference on Knowledge
  Discovery and Data Mining}, pp.\  3770--3781. ACM, 2024.
\newblock \doi{10.1145/3637528.3672039}.

\bibitem[Yang et~al.(2025{\natexlab{a}})Yang, B~B, Feng, Xiong, Liu, King, and
  Ying]{yang2025hyplora}
Menglin Yang, Ram~Samarth B~B, Aosong Feng, Bo~Xiong, Jiahong Liu, Irwin King,
  and Rex Ying.
\newblock Hyperbolic fine-tuning for large language models.
\newblock In \emph{Advances in Neural Information Processing Systems},
  2025{\natexlab{a}}.

\bibitem[Yang et~al.(2025{\natexlab{b}})Yang, Zhou, Zhang, Liu, Li, Pan, Xiong,
  and King]{yang2025graphreview}
Menglin Yang, Min Zhou, Tong Zhang, Jiahong Liu, Zhihao Li, Lujia Pan, Hui
  Xiong, and Irwin King.
\newblock {Hyperbolic Graph Neural Networks: A Review of Methods and
  Applications}.
\newblock \emph{arXiv preprint arXiv:2202.13852}, 2025{\natexlab{b}}.
\newblock URL \url{https://arxiv.org/abs/2202.13852v4}.

\bibitem[Yang et~al.(2026)Yang, Liu, Liu, Li, Cao, Zhang, and
  King]{yang2026survey}
Menglin Yang, Zihao Liu, Jiahong Liu, Jindong Li, Linxiao Cao, Buze Zhang, and
  Irwin King.
\newblock {A Survey of Non-Euclidean Geometric Learning from Hyperbolic to
  Mixed-Curvature Spaces}, 2026.
\newblock URL \url{https://hyperboliclearning.github.io/survey/}.

\bibitem[Young et~al.(2014)Young, Lai, Hodosh, and Hockenmaier]{young2014image}
Peter Young, Alice Lai, Micah Hodosh, and Julia Hockenmaier.
\newblock From image descriptions to visual denotations: New similarity metrics
  for semantic inference over event descriptions.
\newblock \emph{Transactions of the Association for Computational Linguistics},
  2:\penalty0 67--78, 2014.
\newblock \doi{10.1162/tacl_a_00166}.
\newblock URL \url{https://aclanthology.org/Q14-1006/}.

\bibitem[Zhang et~al.(2025)Zhang, Zhu, Yang, Liu, Ying, King, and
  Koniusz]{zhang2025collapse}
Yifei Zhang, Hao Zhu, Menglin Yang, Jiahong Liu, Rex Ying, Irwin King, and
  Piotr Koniusz.
\newblock {Understanding and Mitigating Hyperbolic Dimensional Collapse in
  Graph Contrastive Learning}.
\newblock In \emph{Proceedings of the 31st ACM SIGKDD Conference on Knowledge
  Discovery and Data Mining V.1}, pp.\  1984--1995. ACM, 2025.
\newblock \doi{10.1145/3690624.3709249}.
\newblock URL \url{https://doi.org/10.1145/3690624.3709249}.

\end{thebibliography}
\end{document}